\documentclass{elsarticle}
\usepackage{amsmath,amssymb,multirow,mathrsfs}
\usepackage{subfigure,graphicx}
\usepackage{lineno}
\usepackage{epstopdf}
\usepackage{bigstrut, multirow, rotating}
\usepackage{diagbox}
\usepackage{rotating}
\usepackage{tabularx}
\usepackage{makecell}
\usepackage[thmmarks, amsmath]{ntheorem}
{
 \theoremstyle{nonumberplain}
 \theoremheaderfont{\bfseries}
 \theorembodyfont{\normalfont}
 \theoremsymbol{\mbox{$\Box$}}
 
}
\biboptions{numbers,sort&compress}
\DeclareFixedFont{\myfont}{OT1}{ptm}{m}{n}{6pt}
\DeclareFixedFont{\myfontb}{OT1}{ptm}{bx}{n}{6pt}
\newtheorem{theorem}{Theorem}

\newtheorem{definition}{Definition}
\newtheorem{example}{Example}
\usepackage[left=1.25in,right=1.25in,top=1.25in,bottom=1.25in,letterpaper]{geometry}
\begin{document}

\title{A Fast Algorithm for Cosine Transform Based Tensor Singular Value Decomposition}
\author[uestc]{Wen-Hao Xu}
\ead{seanxwh@gmail.com}
\author[uestc]{Xi-Le Zhao\corref{cor1}}
\ead{xlzhao122003@163.com}
\author[hkbu]{Michael Ng}
\ead{mng@math.hkbu.edu.hk}

\cortext[cor1]{Corresponding author}

\address[uestc]{School of Mathematical Sciences, University of Electronic Science and Technology of China, Chengdu, Sichuan, 611731, P. R. China}
\address[hkbu]{Department of Mathematics, Hong Kong Baptist University, Kowloon Tong, Hong Kong}

\begin{frontmatter}
\begin{abstract}
Recently, there has been a lot of research into tensor singular value decomposition (t-SVD) by using discrete Fourier transform (DFT) matrix. The main aims of this paper are to propose and study tensor singular value decomposition based on the discrete cosine transform (DCT) matrix. The advantages of using DCT are that (i) the complex arithmetic is not involved in the cosine transform based tensor singular value decomposition, so the computational cost required can be saved; (ii) the intrinsic reflexive boundary condition along the tubes in the third dimension of tensors is employed, so its performance would be better than that by using the periodic boundary condition in DFT. We demonstrate that the tensor product between two tensors by using DCT can be equivalent to the multiplication between a block Toeplitz-plus-Hankel matrix and a block vector. Numerical examples of low-rank tensor completion are further given to illustrate that the efficiency by using DCT is two times faster than that by using DFT and also the errors of video and multispectral image completion by using DCT are smaller than those by using DFT.
\end{abstract}


\end{frontmatter}

\section{Introduction}
A tensor is a multi-dimensional array of numbers, which is a generalization of a matrix. Compared to a ``flat'' matrix, a tensor provides a richer and more natural representation for many data. In this paper, we focus on the third-order tensor which looks like a magic cube. This format of data is widely used in color image and gray-scale video inpainting \cite{bertalmio2000image, komodakis2006image, liu2013tensor, korah2007spatiotemporal, chan2011an, jiang2017a}, hyperspectral image (HSI) data recovery \cite{li2012coupled, zhao2013deblurring, li2010tensor, xing2012dictionary}, personalized web search \cite{sun2005cubesvd:}, high-order web link analysis \cite{kolda2005higher-order}, magnetic resonance imaging (MRI) data recovery \cite{varghees2012adaptive}, and seismic data reconstruction \cite{kreimer2012a}.

Like the matrix decomposition, the tensor decomposition is an important multilinear algebra tool. There are many different tensor decompositions. The CANDECOMP/PAEAFAC (CP) decomposition \cite{CPdecomposition} and the Tucker decomposition \cite{tucker1966some} are the two most well-known ones. The CP decomposition can be considered as the higher order generalization of the matrix singular value decomposition (SVD). It tries to decompose a tensor into a sum of rank-one tensors. Similar to the rank-one matrix, third-order rank-one tensors can be written as the outer product of 3 vectors. The CP-rank of a tensor is defined as the minimum number of rank-one tensors whose sum generates the original tensor. This definition is an analog of the definition of matrix rank. The Tucker decomposition is the higher order generalization of the principal component analysis (PCA). It decomposes a tensor into a core tensor multiplied by a matrix along each mode. The Tucker rank based on Tucker decomposition is a vector whose $i$-th element is the mode-$i$ unfolding matrix rank.

Recent years, Kilmer and Martin \cite{kilmer2011factorization, martin2013an, kilmer2013third-order} proposed a third-order tensor decomposition called tensor singular value decomposition (t-SVD). This decomposition strategy is based on the definition of the tensor product (see Section 2). After performing one-dimensional discrete Fourier transformation (DFT) on the third dimension of the tensor, this tensor product makes tensor decomposition be an analog of matrix decomposition. This strategy avoids the loss of structure information in matricization of the tensor. But because of performing one-dimensional DFT on the third dimension, the obtained tensor is a complex tensor. These complex numbers lead to higher computational cost and are not required. Why don't we use another transformation instead of DFT to avoid its disadvantage? Discrete cosine transformation (DCT) \cite{ng1999a} is the first alternative which expresses a finite sequence in terms of a sum of the cosine functions.

DCT only produces the real number for real input. This feature greatly reduces the data in the process of t-SVD, thus saving a lot of time. And there is another difference: DFT implies periodic boundary conditions (BC) when DCT implies reflexive BCs which yields a continuous extension at the boundaries \cite{ng1999a}. If the signal satisfies reflexive BCs (real data often satisfies), the new t-SVD based on DCT can achieve better results than DFT. We give the theoretical derivation of using DCT for t-SVD and verify the superiority compared to DFT.

The rest of this paper is as follows. In Section 2, we introduce some related notations and the original t-SVD with DFT background. In Section 3, we propose the theoretical derivation of new t-SVD with DCT. Based on the new t-SVD, we introduce the new tensor nuclear norm in Section 4. We conduct extensive experiments to demonstrate the effectiveness of the proposed method in Section 5. In Section 6, we give some concluding remarks.

 \section{Notations and Preliminaries}

In this section, we introduce the basic notations and give the definitions related to the t-SVD.
We use non-bold lowercase letters for scalars, e.g., $x$, boldface lowercase letters for vectors, e.g., $\mathbf{x}$, boldface capital letters for matrices, e.g., $\mathbf{X}$, boldface Calligraphy letters for tensors, e.g., $\mathcal{X}$. $\mathbb{R}$ and $\mathbb{C}$ represent the field of real number and complex number, respectively. For a third-order tensor $\mathcal{X}$, we use the MATLAB notations $\mathcal{X}(i,:,:)$, $\mathcal{X}(:,j,:)$, and $\mathcal{X}(:,:,k)$ to denote the horizontal, lateral, and frontal slices, respectively, and $\mathcal{X}(:,j,k)$, $\mathcal{X}(i,:,k)$, and $\mathcal{X}(i,j,:)$ to denote the columns, rows, and tubes, respectively. For convenience, we use $\mathbf{X}^{(k)}$ for the $k$th \textbf{frontal slice} and $\mathbf{x}_{ij:}$ for the $(i,j)$-th \textbf{tube} $\mathcal{X}(i,j,:)$. Both $\mathcal{X}(i,j,k)$ and $x_{ijk}$ represent the $(i,j,k)$-th element. The Frobenius norm of $\mathcal{X}$ is defined as $\left \| \mathcal{X} \right \|_{F} := (\sum_{i,j,k}|x_{ijk}|^{2})^{\frac{1}{2}}$. It is easily to see that $\left \| \mathcal{X} \right \|_{F}^{2} = \sum_{n=1}^{k} \left \| \mathbf{X}^{(n)} \right \|_{F}^{2}$.

Next, we introduce some definitions that are closely related to t-SVD. We use $\tilde{\mathcal{X}} \in \mathbb{C}^{m_{1} \times m_{2} \times m_{3}}$ to represent the discrete Fourier transform of $\mathcal{X} \in \mathbb{C}^{m_{1} \times m_{2} \times m_{3}}$ along each tube, i.e., $\tilde{\mathcal{X}}=\mathrm{fft}(\mathcal{X},[ \thinspace],3)$. The block circulant matrix \cite{martin2013an, kilmer2013third-order} is defined as
\begin{equation} \label{bcirc}
\text{bcirc}(\mathcal{X}):=\left[
  \begin{array}{cccc}
    \mathbf{X}^{(1)} & \mathbf{X}^{(m_{3})} &  \cdots & \mathbf{X}^{(2)} \\
    \mathbf{X}^{(2)} & \mathbf{X}^{(1)} &  \cdots & \mathbf{X}^{(3)} \\
    \vdots      & \vdots      &  \ddots &  \vdots             \\
    \mathbf{X}^{(m_{3})} & \mathbf{X}^{(m_{3}-1)} &  \cdots & \mathbf{X}^{(1)} \\
  \end{array}
\right].
\end{equation}
The block diagonal matrix and the corresponding inverse operator \cite{martin2013an, kilmer2013third-order} are defined as
\begin{equation} \label{bdiag}
\text{bdiag}(\mathcal{X}):=\left[
  \begin{array}{cccc}
    \mathbf{X}^{(1)} &  &  &  \\
      & \mathbf{X}^{(2)} &  & \\
      &  &  \ddots &          \\
      &  &  & \mathbf{X}^{(m_{3})} \\
  \end{array}
\right],
\end{equation}
$$
    \text{unbdiag}(\text{bdiag}(\mathcal{X}))=\mathcal{X}.
$$
The unfold and fold operators in t-SVD \cite{martin2013an, kilmer2013third-order} are defined as
\begin{equation}\label{fold}
  \text{unfold}(\mathcal{X}):=\left [
   \begin{array}{c}
     \mathbf{X}^{(1)} \\
     \mathbf{X}^{(2)} \\
     \vdots \\
     \mathbf{X}^{(m_{3})}
   \end{array} \right ], \quad
   \text{fold}(\text{unfold}(\mathcal{X}))=\mathcal{X}.
\end{equation}
It is a important point that block circulant matrix can be block diagonalized.
\begin{theorem}[\cite{kilmer2011factorization}]\label{block diagonalized}
  \begin{equation}
  \text{bdiag}(\tilde{\mathcal{X}})=(\mathbf{F}_{m_{3}} \otimes \mathbf{I}_{m_{1}})\text{bcirc}(\mathcal{X})(\mathbf{F}^{H}_{m_{3}} \otimes \mathbf{I}_{m_{2}}),
\end{equation}
where $\otimes$ denotes the Kronecker product, $\mathbf{F}_{m_{3}}$ is an $m_{3} \times m_{3}$ DFT matrix and $\mathbf{I}_{m}$ is an $m \times m$ identity matrix.
\end{theorem}

\begin{definition}[t-product \cite{kilmer2013third-order}]
  Given $\mathcal{X} \in \mathbb{C}^{m_{1} \times m_{2} \times m_{3}}$ and $\mathcal{Y} \in \mathbb{C}^{m_{2} \times m_{4} \times m_{3}}$, the t-product $\mathcal{X} \ast \mathcal{Y}$ is a third-order tensor of size $m_{1} \times m_{4} \times m_{3}$
  \end{definition}
  \begin{equation}\label{tproduct}
    \mathcal{Z}=\mathcal{X} \ast \mathcal{Y} := \text{fold}(\text{bcirc}(\mathcal{X})\text{unfold}(\mathcal{Y})).
  \end{equation}

This definition is the core of t-SVD. It is like a one-dimensional convolution of two vectors under reflexive BCs, but the elements of vectors are the frontal slices of tensors. With Theorem \ref{block diagonalized}, equation (\ref{tproduct}) can be rewritten as
\begin{equation}\label{t-product DFT}
  \begin{split}
     \tilde{\mathcal{Z}} &= \text{fold}(\text{bdiag}(\tilde{\mathcal{X}})((\mathbf{F}_{m_{3}} \otimes \mathbf{I}_{m_{2}})\text{unfold}(\mathcal{Y}))) \\
       &=\text{fold}(\text{bdiag}(\tilde{\mathcal{X}})\text{unfold}(\tilde{\mathcal{Y}}))\\
       &=\text{unbdiag}(\text{bdiag}(\tilde{\mathcal{X}})\text{bdiag}(\tilde{\mathcal{Y}})).
  \end{split}
\end{equation}
Equation (\ref{t-product DFT}) means that the t-product in the spatial domain corresponds to the matrix multiplication of the frontal slices in the Fourier domain, which greatly simplifies the process of the algorithm.

\begin{definition}[identity tensor \cite{kilmer2013third-order}]
  The identity tensor $\mathcal{I} \in \mathbb{C}^{m_{1} \times m_{1} \times m_{3}}$ is a tensor whose first frontal slice is the identity matrix of size $m_{1} \times m_{1}$, and whose other frontal slices are all zeros.
\end{definition}

\begin{definition}[orthogonal tensor \cite{kilmer2013third-order}]
  A tensor $\mathcal{Q} \in \mathbb{C}^{m_{1} \times m_{1} \times m_{3}}$ is orthogonal if it satisfies $\mathcal{Q} \ast \mathcal{Q}^{H} = \mathcal{Q}^{H} \ast \mathcal{Q} = \mathcal{I}$, where $\mathcal{Q}^{H}$ is the tensor conjugate transpose of $\mathcal{Q}$, which is obtained by conjugate transposing each frontal slice of $\mathcal{Q}$.
\end{definition}

\begin{definition}[f-diagonal tensor \cite{kilmer2013third-order}]
  A tensor is called f-diagonal if each of its frontal slices is a diagonal matrix.
\end{definition}

\begin{theorem}[t-SVD \cite{kilmer2013third-order, kilmer2011factorization}]
  Given a tensor $\mathcal{X} \in \mathbb{C}^{m_{1} \times m_{2} \times m_{3}}$, the t-SVD of $\mathcal{X}$ is given by
  \begin{equation}\label{tSVD}
    \mathcal{X} = \mathcal{U} \ast \mathcal{S} \ast \mathcal{V}^{H} ,
  \end{equation}
  where $\mathcal{U} \in \mathbb{C}^{m_{1} \times m_{1} \times m_{3}}$,$\mathcal{V} \in \mathbb{C}^{m_{2} \times m_{2} \times m_{3}}$ are orthogonal tensors, and $\mathcal{S} \in \mathbb{C}^{m_{1} \times m_{2} \times m_{3}}$ is a f-diagonal tensor.
\end{theorem}
\begin{figure}[!htp]
  \centering
  \includegraphics[width=0.8\textwidth]{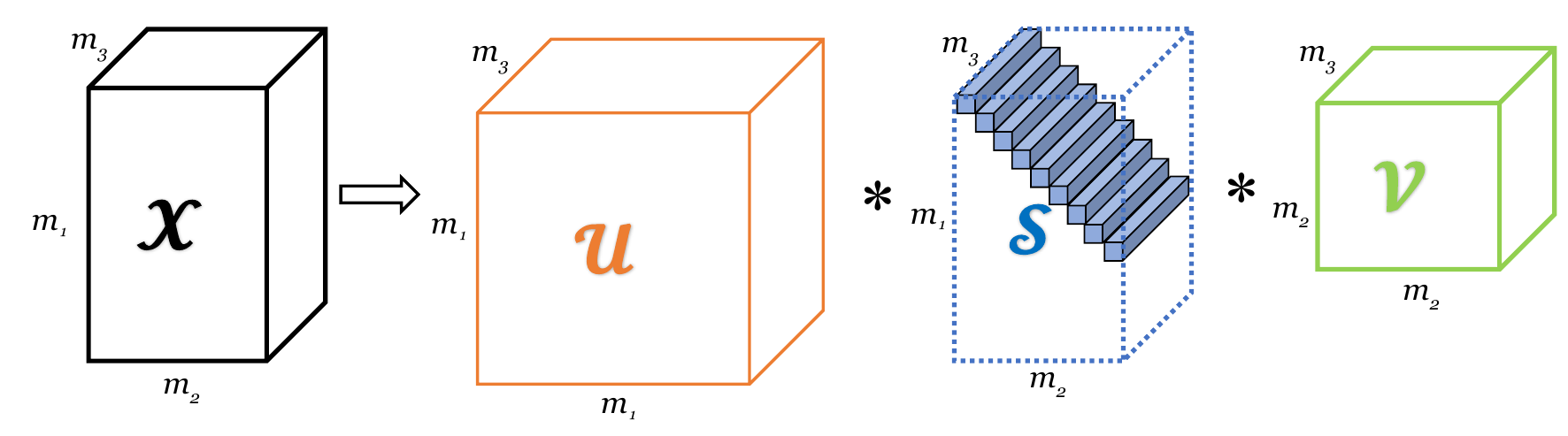}
  \caption{the t-SVD of an $m_{1} \times m_{2} \times m_{3}$ tensor. }\label{figT-SVD}
\end{figure}

\begin{definition}[tensor multi-rank and tubal rank \cite{zhang2014novel}]
  Given $\mathcal{X} \in \mathbb{C}^{m_{1} \times m_{2} \times m_{3}}$, its multi-rank is a vector $\mathbf{r} \in \mathbb{R}^{m_{3}}$ whose $i$-th element is the rank of the $i$-th frontal slice of $\tilde{\mathcal{X}}$, i.e., $\mathbf{r}_{i} = rank(\tilde{\mathbf{X}}^{(i)})$. Its tubal rank is defined as the number of nonzero singular tubes, where the singular tubes of $\mathcal{X}$ are the nonzero tubes of $\mathcal{S}$.
\end{definition}

%\begin{definition}[tensor tubal rank \cite{zhang2014novel}]
%  Given $\mathcal{X} \in \mathbb{C}^{m_{1} \times m_{2} \times m_{3}}$, it has a t-SVD as $\mathcal{X} = \mathcal{U} \ast \mathcal{S} \ast \mathcal{V}^{H}$. Then the tubal rank of $\mathcal{X}$ is defined as the number of nonzero singular tubes.
%\end{definition}
The tensor tubal rank is actually the largest element of multi-rank.
 \begin{definition}[tensor nuclear norm \cite{lu2016tensor, semerci2014tensor-based}]
   Given $\mathcal{X} \in \mathbb{C}^{m_{1} \times m_{2} \times m_{3}}$, based on the tensor multi-rank, the tensor nuclear norm (TNN) of $\mathcal{X}$ is defined as
   \begin{equation}\label{tnn}
     \left \| \mathcal{X} \right \|_{\ast} := \frac{1}{m_{3}} \sum_{k=1}^{m_{3}} \left \| \tilde{\mathbf{X}}^{(k)} \right \|_{\ast} .
   \end{equation}
 \end{definition}
 In order to avoid confusion with the new definition of TNN we proposed later, we call this definition TNN-F in this paper.

The computation of t-SVD on an $m_{1} \times m_{2} \times m_{3}$ tensor needs two steps. Firstly, the first step is to
perform DFT by fast Fourier transformation (FFT) along each tube. The time complexity of the first step is $O(m_{1} m_{2} m_{3}\log(m_{3}))$. After DFT, the obtained tensor is a complex tensor which can be divided into a real number tensor and an imaginary number tensor.
The computation of SVD along each frontal slice on the obtained tensor is actually equivalent to performing on the real number tensor and the imaginary number tensor respectively. The time complexity of the second step is $O(2m_{3} \min(m_{1} m_{2}^{2}, m_{2} m_{1}^{2}))$,
which is about the computational cost of the first step.

 \section{Cosine Transform Based Tensor Singular Value Decomposition}

We discuss the DCT-based t-SVD and the resulting structure in this section. Since the corresponding block circulant matrices can be diagonalized by DFT, the DFT based t-SVD can be efficiently implemented via fast Fourier transform (fft). We will show the corresponding structure of DCT-based t-SVD can be diagonalized by DCT.

We define the shift of tensor $\mathcal{A} = \text{fold}\left [
   \begin{array}{c}
     \mathbf{A}^{(1)} \\
     \mathbf{A}^{(2)} \\
     \vdots \\
     \mathbf{A}^{(m_{3})}
   \end{array} \right ]$ as $\sigma(\mathcal{A}) = \text{fold}\left [
   \begin{array}{c}
     \mathbf{A}^{(2)} \\
     \mathbf{A}^{(3)} \\
     \vdots \\
     \mathbf{A}^{(m_{3})}\\
     \mathbf{O}
   \end{array} \right ]$.
It is easy to prove that any tensor $\mathcal{X}$ can be uniquely divided into $\mathcal{A} + \sigma(\mathcal{A})$.
We use $\bar{\mathcal{X}} \in \mathbb{R}^{m_{1} \times m_{2} \times m_{3}}$ to represent the DCT along each tube of $\mathcal{X}$, i.e., $\bar{\mathcal{X}}=\mathrm{dct}(\mathcal{X},[\thinspace],3)=\mathrm{dct}(\mathcal{A}+\sigma(\mathcal{A}),[ \thinspace],3)$. We define the block Toeplitz matrix of $\mathcal{A}$ as
\begin{equation} \label{btplz}
\text{bt}(\mathcal{A}):=\left[
  \begin{array}{ccccc}
    \mathbf{A}^{(1)} & \mathbf{A}^{(2)} &  \cdots & \mathbf{A}^{(m_{3}-1)} & \mathbf{A}^{(m_{3})} \\
    \mathbf{A}^{(2)} & \mathbf{A}^{(1)} &  \cdots & \mathbf{A}^{(m_{3}-2)} & \mathbf{A}^{(m_{3}-1)} \\
    \vdots      & \vdots      &  \ddots &  \vdots             \\
    \mathbf{A}^{(m_{3}-1)} & \mathbf{A}^{(m_{3}-2)} &  \cdots & \mathbf{A}^{(1)} & \mathbf{A}^{(2)} \\
    \mathbf{A}^{(m_{3})} & \mathbf{A}^{(m_{3}-1)} &  \cdots & \mathbf{A}^{(2)} & \mathbf{A}^{(1)} \\
  \end{array}
\right].
\end{equation}
The block Hankel matrix is defined as
\begin{equation} \label{bhkl}
\text{bh}(\mathcal{A}):=\left[
  \begin{array}{ccccc}
    \mathbf{A}^{(2)} & \mathbf{A}^{(3)} &  \cdots & \mathbf{A}^{(m_{3})} & \mathbf{O} \\
    \mathbf{A}^{(3)} & \mathbf{A}^{(4)} &  \cdots & \mathbf{O} & \mathbf{A}^{(m_{3})} \\
    \vdots      & \vdots      &  \ddots &  \vdots             \\
    \mathbf{A}^{(m_{3})} & \mathbf{O} &  \cdots & \mathbf{A}^{(4)} & \mathbf{A}^{(3)} \\
    \mathbf{O} & \mathbf{A}^{(m_{3})} &  \cdots & \mathbf{A}^{(3)} & \mathbf{A}^{(2)} \\
  \end{array}
\right].
\end{equation}
The block Toeplitz-plus-Hankel matrix of $\mathcal{A}$ is defined as
\begin{equation} \label{btph}
\text{btph}(\mathcal{A}):= \text{bt}(\mathcal{A}) + \text{bh}(\mathcal{A}).
\end{equation}
The block Toeplitz-plus-Hankel matrix can be diagonalized. The following theorem can by similarly established as \cite{ng1999a}.
\begin{theorem}\label{dct block diagonalized}
  \begin{equation}
  \text{bdiag}(\bar{\mathcal{X}})=(\mathbf{C}_{m_{3}} \otimes \mathbf{I}_{m_{1}})\text{btph}(\mathcal{A})(\mathbf{C}^{T}_{m_{3}} \otimes \mathbf{I}_{m_{2}}),
\end{equation}
where $\otimes$ denotes the Kronecker product, $\mathbf{C}_{m_{3}}$ is an $m_{3} \times m_{3}$  DCT matrix.
\end{theorem}

The proof of Theorem 3 can be obtained by using the similar argument in \cite{ng1999a}.
We briefly illustrate this theorem with an example.

\begin{example}
  Let the frontal slice of $\mathcal{X} \in \mathbb{R}^{2 \times 2 \times 2}$ be
  $$
    \mathbf{X}^{(1)} = \left[
        \begin{array}{cc}
          1 & 2 \\
          3 & 4
        \end{array}
\right], \quad   \mathbf{X}^{(2)} = \left[
        \begin{array}{cc}
          5 & 6 \\
          7 & 8
        \end{array}
\right].
  $$
  So the component $\mathcal{A}$ is
  $$
  \mathbf{A}^{(1)} =\mathbf{X}^{(1)}-\mathbf{X}^{(2)} = \left[
        \begin{array}{cc}
          -4 & -4 \\
          -4 & -4
        \end{array}
\right], \quad   \mathbf{A}^{(2)} =
  \mathbf{X}^{(2)} = \left[
        \begin{array}{cc}
          5 & 6 \\
          7 & 8
        \end{array}
\right].
  $$
  The block Toeplitz matrix is
  $$
    \text{bt}(\mathcal{A}) =\left [
    \begin{array}{cc}
\mathbf{A}^{(1)} & \mathbf{A}^{(2)} \\
\mathbf{A}^{(2)} & \mathbf{A}^{(1)} \end{array}
    \right ] = \left[
     \begin{array}{cccc}
                                  -4 & -4 & 5 & 6 \\
                                  -3 & -4 & 7 & 8 \\
                                  5 & 6 & -4 & -4 \\
                                  7 & 8 & -4 & -4
                                \end{array}
                                \right],
  $$
  and the block Hankel matrix is
   $$
    \text{bh}(\mathcal{A}) =\left [
    \begin{array}{cc}
\mathbf{A}^{(2)} & 0 \\
0 & \mathbf{A}^{(2)} \end{array}
    \right ] = \left[
     \begin{array}{cccc}
                                  5 & 6 & 0 & 0 \\
                                  7 & 8 & 0 & 0 \\
                                  0 & 0 & 5 & 6 \\
                                  0 & 0 & 7 & 8
                                \end{array}
                                \right].
  $$
  Then the block Toeplitz-plus-Hankel matrix is
  $$
    \text{btph}(\mathcal{A}) = \text{bt}(\mathcal{A})+\text{bh}(\mathcal{A}) = \left[
     \begin{array}{cccc}
                                  1 & 2 & 5 & 6 \\
                                  3 & 4 & 7 & 8 \\
                                  5 & 6 & 1 & 2 \\
                                  7 & 8 & 3 & 4
                                \end{array}
                                \right].
  $$
  By using stride permutations, we get
  $$
  \mathbf{P} \text{btph}(\mathcal{A}) \mathbf{P} = \left[
     \begin{array}{cccc}
                                  1 & 5 & 2 & 6 \\
                                  5 & 1 & 6 & 2 \\
                                  3 & 7 & 4 & 8 \\
                                  7 & 3 & 8 & 4
                                \end{array}
                                \right] = \left[
                                \begin{array}{cc}
                                  \mathbf{A} & \mathbf{B} \\
                                  \mathbf{C} & \mathbf{D}
                                \end{array}
                                \right],
  $$
  where $\mathbf{P} = \left[
     \begin{array}{cccc}
                                  1 & 0 & 0 & 0 \\
                                  0 & 0 & 1 & 0 \\
                                  0 & 1 & 0 & 0 \\
                                  0 & 0 & 0 & 1
                                \end{array}
                                \right]$
  and $\mathbf{A}$, $\mathbf{B}$, $\mathbf{C}$, and $\mathbf{D}$ are Toeplitz-plus-Hankel matrices.
  So we have
  $$
    (\mathbf{C}_{2} \otimes \mathbf{I}_{2})\text{btph}(\mathcal{A})(\mathbf{C}^{T}_{2} \otimes \mathbf{I}_{2})=(\mathbf{C}_{2} \otimes \mathbf{I}_{2})\mathbf{P}\mathbf{P}\text{btph}(\mathcal{A})\mathbf{P}\mathbf{P}(\mathbf{C}^{T}_{2} \otimes \mathbf{I}_{2}),
  $$
  where $\mathbf{C}_{2}$ is a $2 \times 2$ DCT matrix. In this equation, it is easy to see that
  $$
  \mathbf{P}(\mathbf{C}_{2} \otimes \mathbf{I}_{2})\mathbf{P}= \left[
    \begin{array}{cc}
      \mathbf{C}_{2} & 0  \\
      0 & \mathbf{C}_{2}
    \end{array}
  \right].
  $$
  Similarly,
  $$
  \mathbf{P}(\mathbf{C}_{2}^{T} \otimes \mathbf{I}_{2})\mathbf{P} = \left[
    \begin{array}{cc}
      \mathbf{C}_{2}^{T} & 0 \\
      0 & \mathbf{C}_{2}^{T}
    \end{array}
  \right].
  $$
  Hence, we have

  \begin{align}
     (\mathbf{C}_{2} \otimes \mathbf{I}_{2})\text{btph}(\mathcal{A})(\mathbf{C}^{T}_{2} \otimes \mathbf{I}_{2}) &=  \mathbf{P}\left[
    \begin{array}{cc}
      \mathbf{C}_{2} & 0  \\
      0 & \mathbf{C}_{2}
    \end{array}
  \right]\left[
                                \begin{array}{cc}
                                  \mathbf{A} & \mathbf{B} \\
                                  \mathbf{C} & \mathbf{D}
                                \end{array}
                                \right]\left[
    \begin{array}{cc}
      \mathbf{C}_{2}^{T} & 0 \\
      0 & \mathbf{C}_{2}^{T}
    \end{array}
  \right]\mathbf{P} \nonumber\\
       &= \mathbf{P} \left[
  \begin{array}{cc}
    \mathbf{C}_{2} \mathbf{A} \mathbf{C}_{2}^{T} & \mathbf{C}_{2} \mathbf{B} \mathbf{C}_{2}^{T} \\
    \mathbf{C}_{2} \mathbf{C} \mathbf{C}_{2}^{T} & \mathbf{C}_{2} \mathbf{D} \mathbf{C}_{2}^{T}
  \end{array}
  \right] \mathbf{P} \nonumber\\
  & = \left[
     \begin{array}{cccc}
                                  6 & 8 & 0 & 0 \\
                                  10 & 12 & 0 & 0 \\
                                  0 & 0 & -4 & -4 \\
                                  0 & 0 & -4 & -4
                                \end{array}
                                \right]. \nonumber
  \end{align}
Now, it is easy to verify
 \begin{align}
     \text{bdiag}(\bar{\mathcal{X}}) &= \text{bdiag}(\text{dct}(\mathcal{A}+\sigma(\mathcal{A}),[\thinspace],3)) \nonumber \\
       &= (\mathbf{C}_{2} \otimes \mathbf{I}_{2})\text{btph}(\mathcal{A})(\mathbf{C}^{T}_{2} \otimes \mathbf{I}_{2}). \nonumber
  \end{align}
\end{example}
%\textbf{Remark}: The block Toeplitz-plus-Hankel matrix is made up by $\mathcal{A}$.

\begin{definition}[DCT-based t-product]
  Given $\mathcal{X} \in \mathbb{C}^{m_{1} \times m_{2} \times m_{3}}$ and $\mathcal{Y} \in \mathbb{C}^{m_{2} \times m_{4} \times m_{3}}$, the t-product $\mathcal{X} \ast \mathcal{Y}$ is a third-order tensor of size $m_{1} \times m_{4} \times m_{3}$
  \end{definition}
  \begin{equation}\label{ntproduct}
    \mathcal{Z}=\mathcal{X} \ast \mathcal{Y} := \text{fold}(\text{btph}(\mathcal{A})\text{unfold}(\mathcal{Y})),
  \end{equation}
  where $\mathcal{X} = \mathcal{A}+\sigma(\mathcal{A})$.

Equation (\ref{ntproduct}) can be rewritten as
\begin{equation}\label{tproduct dct}
  \begin{split}
     \bar{\mathcal{Z}} &= \text{fold}(\text{bdiag}(\bar{\mathcal{X}})((\mathbf{C}_{m_{3}} \otimes \mathbf{I}_{m_{2}})\text{unfold}(\mathcal{Y}))) \\
       &=\text{fold}(\text{bdiag}(\bar{\mathcal{X}})\text{unfold}(\bar{\mathcal{Y}})).
  \end{split}
\end{equation}
Based on this new t-product, the DCT-based t-SVD can be defined as follows:

\begin{theorem}[DCT-based t-SVD]
  Given a tensor $\mathcal{X} \in \mathbb{R}^{m_{1} \times m_{2} \times m_{3}}$, the DCT-based t-SVD of $\mathcal{X}$ is given by
  \begin{equation}\label{ctSVD}
    \mathcal{X} = \mathcal{U} \ast \mathcal{S} \ast \mathcal{V}^{T} ,
  \end{equation}
  where $\mathcal{U} \in \mathbb{R}^{m_{1} \times m_{1} \times m_{3}}$,$\mathcal{V} \in \mathbb{R}^{m_{2} \times m_{2} \times m_{3}}$ are orthogonal tensors, $\mathcal{S} \in \mathbb{R}^{m_{1} \times m_{2} \times m_{3}}$ is a f-diagonal tensor, and $\mathcal{V}^{T}$ is the tensor transpose of $\mathcal{V}$, which is obtained by transposing each frontal slice of $\mathcal{V}$.
\end{theorem}

The proof of Theorem 4 can be obtained by using the similar argument in \cite{kilmer2013third-order}.

By exploiting the beautiful structure, the DCT-based t-SVD can be efficiently calculated by performing the matrix singular value decomposition for each frontal slice of the third-order tensor after DCT along each tube. For an $m_{1} \times m_{2} \times m_{3}$ tensor, the time complexity of performing DCT along each tube in the first step is $O(m_{1} m_{2} m_{3}\log(m_{3}))$ for DCT-based t-SVD, which is the same as that DFT-based t-SVD. Since DCT only produces the real number, the time complexity of calculating SVDs is $O(m_{3} \min(m_{1} m_{2}^{2}, m_{2} m_{1}^{2}))$ for DCT-based t-SVD, which is half that of DFT-based t-SVD.
\begin{table}[htbp]
  \small
  \centering
  \setlength{\abovecaptionskip}{0pt}%
    \setlength{\belowcaptionskip}{10pt}%
    \renewcommand\arraystretch{0.9}
  \caption{The time complexity of t-SVD and DCT-based t-SVD on an $m_{1} \times m_{2} \times m_{3}$ tensor.}
    \begin{tabular}{cc}
    \hline
    tensor & $m_{1} \times m_{2} \times m_{3}$ \bigstrut\\
    \hline
    DFT   & $O(m_{1} m_{2} m_{3}\log(m_{3}))$ \bigstrut[t]\\
    SVD after DFT & $O(2m_{3} \min(m_{1} m_{2}^{2}, m_{2} m_{1}^{2}))$ \\
    t-SVD & $O(m_{1} m_{2} m_{3}\log(m_{3}))+O(2m_{3} \min(m_{1} m_{2}^{2}, m_{2} m_{1}^{2}))$ \bigstrut[b]\\
    \hline
    \hline
    DCT   & $O(m_{1} m_{2} m_{3}\log(m_{3}))$ \bigstrut[t]\\
    SVD after DCT & $O(m_{3} \min(m_{1} m_{2}^{2}, m_{2} m_{1}^{2}))$ \\
    new t-SVD & $O(m_{1} m_{2} m_{3}\log(m_{3}))+O(m_{3} \min(m_{1} m_{2}^{2}, m_{2} m_{1}^{2}))$ \bigstrut[b]\\
    \hline
    \end{tabular}%
  \label{tab:addlabel}%

  \label{tabTestCost1}%
\end{table}%

\section{Low-rank Tensor Completion by TNN-C}
Based on the DCT-based t-SVD, we propose the new definition of TNN called TNN-C in this section. Then, we establish the low-rank tensor completion model \cite{jiang2017a} based on TNN-C and develop the alternating direction method of multipliers (ADMM) to tackle the corresponding low-rank tensor completion model.

\begin{definition}[TNN-C]
Given $\mathcal{X} \in \mathbb{R}^{m_{1} \times m_{2} \times m_{3}}$, TNN-C of $\mathcal{X}$ is defined as
  \begin{equation}\label{equTNN1}
\left \| \mathcal{X}\right \|_{\ast} =\frac{1}{m_{3}} \sum_{i=1}^{m_{3}} \left \| \bar{\mathbf{X}}^{(i)} \right \|_{\ast}.
\end{equation}
\end{definition}
It is easy to see that TNN-C of $\mathcal{X}$ is the sum of singular values of all frontal slices of $\bar{\mathcal{X}}$. Meanwhile, the $i$-th element of multi-rank is the rank of the $i$-th frontal slice of $\bar{\mathcal{X}}$. Thus, TNN-C is a convex surrogate of the $l_{1}$ norm of a third-order tensor's multi-rank.

 %TNN-C is a convex surrogate for multi-rank defined as
 The low-rank tensor completion model is defined as
  \begin{equation}\label{model}
   \min_{\mathcal{X}} \left \| \mathcal{X} \right \|_{\ast}, \quad s.t. \quad \mathcal{X}_{\Omega} = \mathcal{B}_{\Omega}.
 \end{equation}
Letting
$$
    l_{\mathbb{S}}(\mathcal{X})= \begin{cases}
        0,&  \text{if } \mathcal{X} \in \mathbb{S},    \\
        \infty,& \text{otherwise},
    \end{cases}
$$
where $\mathbb{S} := \{ \mathcal{X} \in \mathbb{R}^{m_{1} \times m_{2} \times m_{3}} , \mathcal{X}_{\Omega} = \mathcal{B}_{\Omega} \} $, (\ref{model}) can be rewritten as the following unconstrained problem:
\begin{equation}\label{unconstrained}
  \min_{\mathcal{X}} \left \| \mathcal{X} \right \|_{\ast} + l_{\mathbb{S}}(\mathcal{X}).
\end{equation}

By introducing an auxiliary variable $\mathcal{Y}=\mathcal{X}$, the augmented Lagrangian function of (\ref{unconstrained}) is
\begin{equation}\label{Lagrangian}
\begin{split}
  L(\mathcal{X},\mathcal{Y},\mathcal{M}) & := \left \| \mathcal{Y} \right \|_{\ast} + l_{\mathbb{S}}(\mathcal{X}) + \langle \mathcal{Y}-\mathcal{X}, \mathcal{M} \rangle + \frac{\beta}{2}\left \| \mathcal{Y}-\mathcal{X} \right \|^{2}_{F} \\
  & = \left \| \mathcal{Y} \right \|_{\ast} + l_{\mathbb{S}}(\mathcal{X}) + \frac{\beta}{2} \left \| \mathcal{Y}-\mathcal{X}+\frac{1}{\beta} \mathcal{M} \right \|^{2}_{F}-\frac{1}{2\beta}\langle\mathcal{M},\mathcal{M\rangle},
\end{split}
\end{equation}
where $ \mathcal{M} \in \mathbb{R}^{m_{1} \times m_{2} \times m_{3}}$ is the Lagrangian multiplier, and $\beta$ is the balance parameter. According to the framework of ADMM \cite{boyd2011distributed, lin2010augmented, he2012alternating}, $\mathcal{X}$, $\mathcal{Y}$, and $\mathcal{M}$ are iteratively updated as
\begin{equation}\label{iterative}
  \begin{cases}
    \begin{aligned}
        \text{Step 1: } &  \mathcal{Y}^{l+1} \in \arg\min_{\mathcal{Y}} L(\mathcal{X}^{l},\mathcal{Y},\mathcal{M}^{l}),    \\
        \text{Step 2: } &  \mathcal{X}^{l+1} \in \arg\min_{\mathcal{X}} L(\mathcal{X},\mathcal{Y}^{l+1},\mathcal{M}^{l}),    \\
        \text{Step 3: } &  \mathcal{M}^{l+1} = \mathcal{M}^{l} + \beta (\mathcal{Y}^{l+1}- \mathcal{X}^{l+1}).
    \end{aligned}
  \end{cases}
\end{equation}
Now, we give the details for solving each subproblem.

In \textbf{Step 1}, the $\mathcal{Y}$-subproblem is:
\begin{equation}\label{step 1}
  \arg\min_{\mathcal{Y}} \left \| \mathcal{Y} \right \|_{\ast} + \frac{\beta}{2} \left \| \mathcal{Y} - \mathcal{X}^{l} + \frac{1}{\beta} \mathcal{M}^{l} \right \|^{2}_{F},
\end{equation}
which can be solved by the following theorem \cite{lu2016tensor, semerci2014tensor-based}.
 \begin{theorem}
  Given $\mathcal{Z} \in \mathbb{C}^{m_{1} \times m_{2} \times m_{3}}$, a minimizer to
  \begin{equation}\label{minimizer}
    \min_{\mathcal{Y}} \left \| \mathcal{Y} \right \|_{\ast} + \frac{\beta}{2} \left \| \mathcal{Y} - \mathcal{Z} \right \|^{2}_{F}
  \end{equation}
 is given by the tensor singular value thresholding
 \begin{equation}\label{wtsvt}
   \mathcal{Y} = \mathcal{U} \ast \mathcal{D}_{\frac{1}{\beta}} \ast \mathcal{V}^{T},
 \end{equation}
 where $\mathcal{Z} = \mathcal{U} \ast \mathcal{S} \ast \mathcal{V}^{T}$ and $\mathcal{D}_{\frac{1}{\beta}}$ is an $\mathbb{R}^{m_{1} \times m_{2} \times m_{3}}$ f-diagonal tensor whose each frontal slice in the discrete cosine domain is $\bar{\mathcal{D}}_{\frac{1}{\beta}}(i,i,j) = (\bar{\mathcal{S}}(i,i,j) - \frac{1}{\beta} )_{+}$.
 \end{theorem}

In \textbf{Step 2}, we solve the following problem:
\begin{equation}\label{step 2}
  \arg\min_{\mathcal{X}} l_{\mathbb{S}}(\mathcal{X}) + \frac{\beta}{2} \left \| \mathcal{Y}^{l+1} - \mathcal{X} + \frac{1}{\beta} \mathcal{M}^{l} \right \|^{2}_{F},
\end{equation}
which has a closed-form solution
\begin{equation}\label{x solution}
  \mathcal{X}^{l+1} = (\mathcal{Y}^{l+1} + \frac{1}{\beta} \mathcal{M}^{l})_{\Omega^{C}} + \mathcal{B},
\end{equation}
where $\Omega ^{C}$ is the complementary set of the index set $\Omega$.

We summarize the proposed ADMM procedure in Algorithm 1. Every step of ADMM has an explicit solution. Thus, the proposed method is efficiently implementable. The convergence of the ADMM method of convex functions of separable variables with linear constraints is guaranteed \cite{afonso2011an, han2012a}.\\[1ex]

\begin{tabular}{l}
\hline
\textbf{Algorithm 1} {\small ADMM for solving the proposed model (\ref{model}).}\\
\hline
\textbf{Input:} Observed data $\mathcal{B}$, index set $\Omega$, parameters $\beta$.\\
\textbf{Initialize:} $\mathcal{X}=\mathcal{B}$, $\mathcal{Y}=\mathbf{0}$, $\mathcal{M}=\mathbf{0}$, $\text{tol}=10^{-5}$, and $L=500$.\\
\indent 1: \textbf{while} $l<L$ and $\left \| \mathcal{X}^{l+1} - \mathcal{X}^{l} \right \|_{F} / \left \| \mathcal{X}^{l} \right \|_{F} > tol$ \textbf{do} \\
\indent 2: \quad $\mathcal{Z}=\mathcal{X}^{l} - \frac{1}{\beta} \mathcal{M}^{l}$; \\
\indent 3: \quad $\bar{\mathcal{Z}} = \mathrm{dct}(\mathcal{Z},[\thinspace],3)$;\\
\indent 4: \quad \textbf{for} $k=1$ to $m_{3}$ \textbf{do} \\
\indent 5: \quad \quad \indent $[\bar{\mathbf{U}}^{(k)},\bar{\mathbf{S}}^{(k)},\bar{\mathbf{V}}^{(k)}]=\mathrm{SVD}(\bar{\mathbf{Z}}^{(k)});$\\
\indent 6: \quad \quad \indent $\bar{\mathbf{D}}^{(k)} = (\bar{\mathbf{S}}^{(k)}-1/ \beta)_{+};$ \\
\indent 7: \quad \quad \indent $\bar{\mathbf{Z}}^{(k),l+1} = \mathbf{U}^{(k)}\bar{\mathbf{D}}^{(k)}\mathbf{V}^{(k) H};$ \\
\indent 8: \quad \textbf{end for}\\
\indent 9: \quad  $\mathcal{Y}^{l+1} = \mathrm{idct}(\bar{\mathcal{Z}}^{l+1},[\thinspace],3);$\\
\indent 10: \quad $\mathcal{X}^{l+1} = (\mathcal{Y}^{l+1} + \frac{1}{\beta} \mathcal{M}^{l})_{\Omega^{c}} + \mathcal{B}$;\\
\indent 11: \quad  $\mathcal{M}^{l+1} = \mathcal{M}^{l} + \beta (\mathcal{Y}^{l+1}- \mathcal{X
}^{l+1}).$\\
\indent 12
: \textbf{end while}\\
\textbf{Output:} The recovered tensor $\mathcal{X}$. \\
\hline
\end{tabular}

\section{Numerical Examples}
In this section, all experiments are implemented on Windows 10 and Matlab (R2017a) with an Intel(R) Core(TM) i7-7700k CPU at 4.20 GHz and 16 GB RAM.

\subsection{The Computational Time}

Saving time is the most important advantage of DCT-based t-SVD. We illustrate this advantage of the new t-SVD by operating on random tensors.
We set 4 groups of random tensors of different size and performed 1000 runs to get the average time required.
Tab.\thinspace\ref{tabTestCost1} shows that average time cost of performing t-SVD and DCT-based t-SVD, and confirms our point that DCT-based t-SVD only needs half the time of t-SVD.
% Table generated by Excel2LaTeX from sheet 'Sheet2'

\begin{table}[htbp]
 % \small
  \centering
  \setlength{\abovecaptionskip}{0pt}%
\setlength{\belowcaptionskip}{10pt}%
    \renewcommand\arraystretch{0.9}
  \caption{The time cost of t-SVD and DCT-based t-SVD on the random tensors of different size.}
    \begin{tabular}{ccccc}
    \hline
    size  & \textit{100*100*100} & \textit{100*100*400} & \textit{200*200*100} & \textit{400*400*100} \bigstrut\\
    \hline
    FFT   & 0.0041  & 0.0175  & 0.0176  & 0.0653  \bigstrut[t]\\
    SVD after FFT & 0.0818  & 0.3250  & 0.3641  & 1.9015  \\
    original t-SVD & 0.0859  & 0.3425  & 0.3817  & 1.9668  \bigstrut[b]\\
    \hline
    \hline
    DCT   & 0.0042  & 0.0150  & 0.0162  & 0.0601  \bigstrut[t]\\
    SVD after DCT & 0.0439  & 0.1649  & 0.1978  & 0.8922  \\
    new t-SVD & 0.0481  & 0.1799  & 0.2140  & 0.9523  \bigstrut[b]\\
    \hline
    \end{tabular}%

  \label{tabTestCost1}%
\end{table}%

\subsection{Real Data}
We conduct the video and multispectral image (MSI) completion experiments and compare TNN-C with the TNN-F \cite{lu2016tensor}. In our experiments, the quality of the recovered image is measured by the average of highest peak signal-to-noise ratio (PSNR) and structural similarity index (SSIM) of all bands. PSNR of a band is defined as follows:
$$
    \text{PSNR}=10\log_{10} \frac{m_{1}m_{2} \mathbf{X}^{2}_{\max}}{\left \| \hat{\mathbf{X}} - \mathbf{X} \right \|^{2}_{F}},
$$
where $\mathbf{X}$ is the masked matrix, $\hat{\mathbf{X}}$ is the recovered matrix, and $\mathbf{X}_{max}$ is the maximum pixel value of the original matrix $\mathbf{X}$. SSIM can measure the similarity between the recovered image and the masked image. This indicator can reflect the similarities in brightness, contrast, and structure of two images and is defined as
$$
    \text{SSIM} = \frac{(2 \mu_{\mathbf{x}} \mu_{\hat{\mathbf{x}}} + c_{1}) (2 \sigma_{\mathbf{x} \hat{\mathbf{x}}} + c_{2}) }{( \mu^{2}_{\mathbf{x}} + \mu^{2}_{\hat{\mathbf{x}}} + c_{1}) ( \sigma^{2}_{\mathbf{x}} + \sigma^{2}_{\hat{\mathbf{x}}} + c_{1}) },
$$
where $\mu_{\mathbf{x}}$ and $\mu_{\hat{\mathbf{x}}}$ represent the average values of the original matrix and the estimated matrix, respectively, $\sigma_{\mathbf{x}}$ and $\sigma_{\hat{\mathbf{x}}}$ represent the standard deviation of $\mathbf{X}$ and $\hat{\mathbf{X}}$, respectively.

For all the following experiments, we set the maximum number of iterations to 500 and the tolerance to $1 \times 10^{-8}$. This algorithm only needs one parameter $\beta$, and we set it to $1 \times 10^{-2}$.

\begin{figure}
%[!htp]
 \begin{center}
 \includegraphics[width=1.0\textwidth,height=3.5cm]{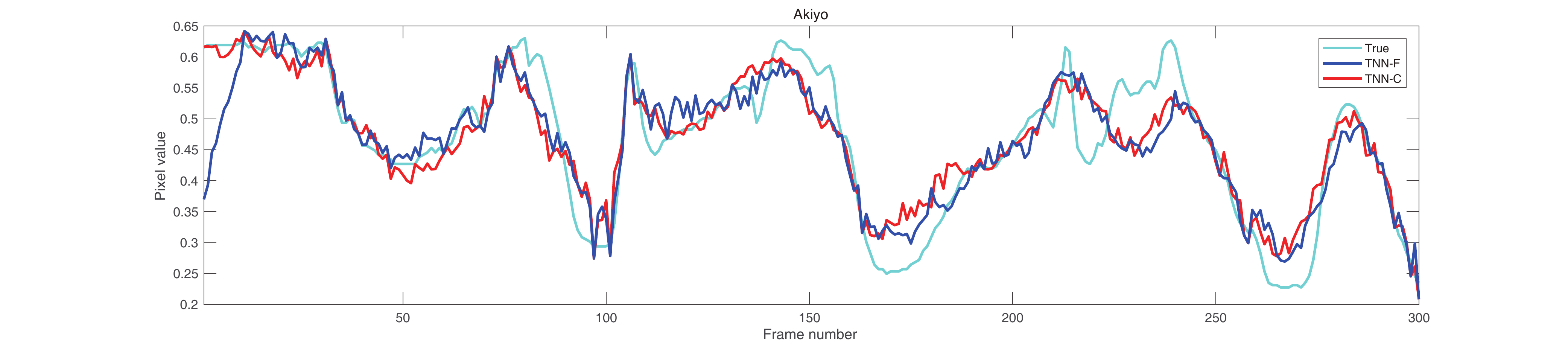}
 \includegraphics[width=1.0\textwidth,height=3.5cm]{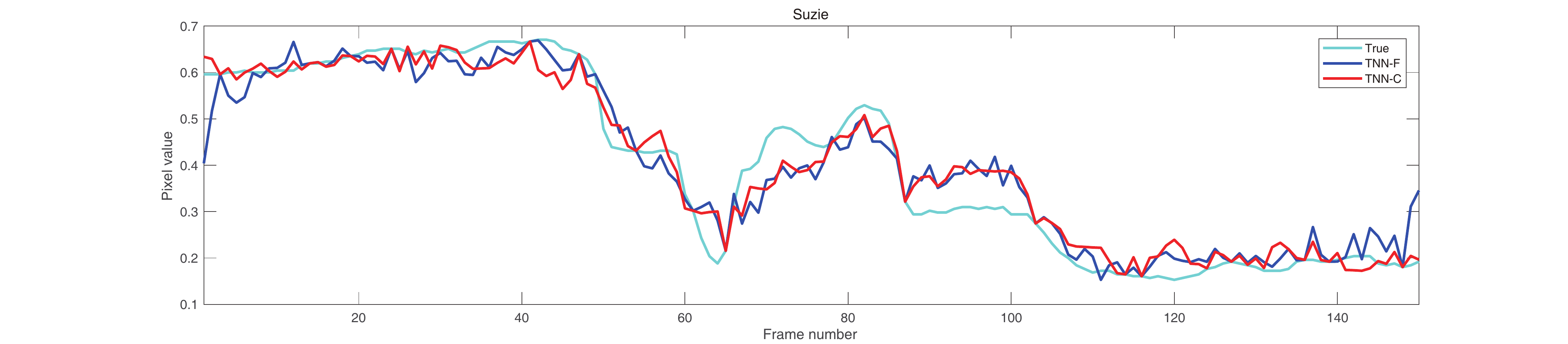}
 \includegraphics[width=1.0\textwidth,height=3.5cm]{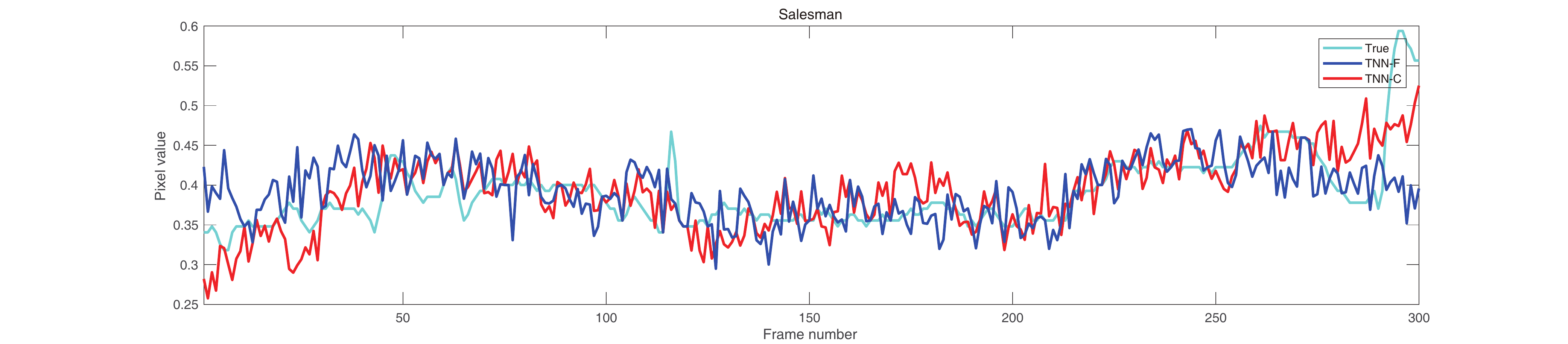}
 \end{center}
 \caption{The pixel value of a selected tube of videos \textit{Akiyo}, \textit{Suzie}, and \textit{Salesman}. }
 \label{figVideoTube}
\end{figure}

\begin{figure}
%[!htp]
 \begin{center}
 \includegraphics[width=1.0\textwidth,height=3.5cm]{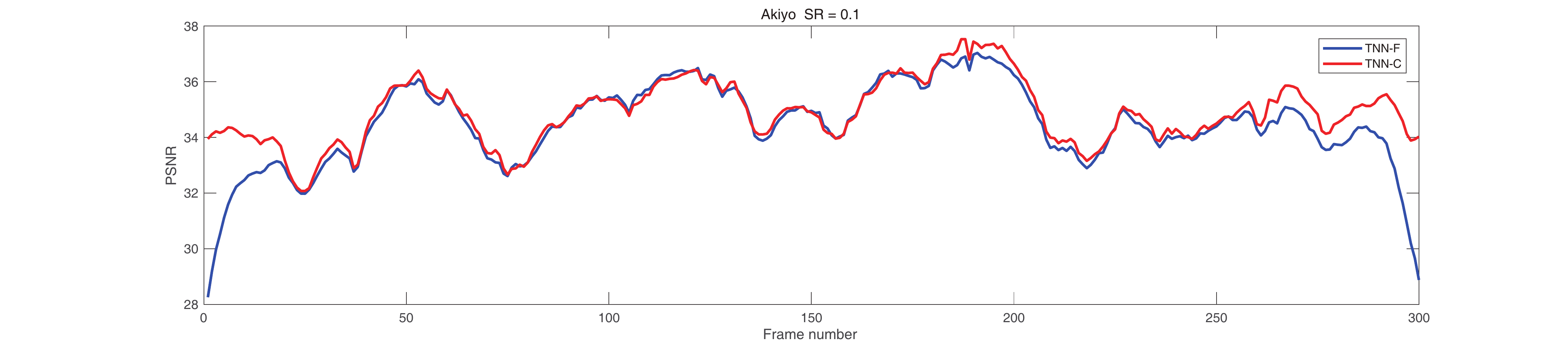}
 \includegraphics[width=1.0\textwidth,height=3.5cm]{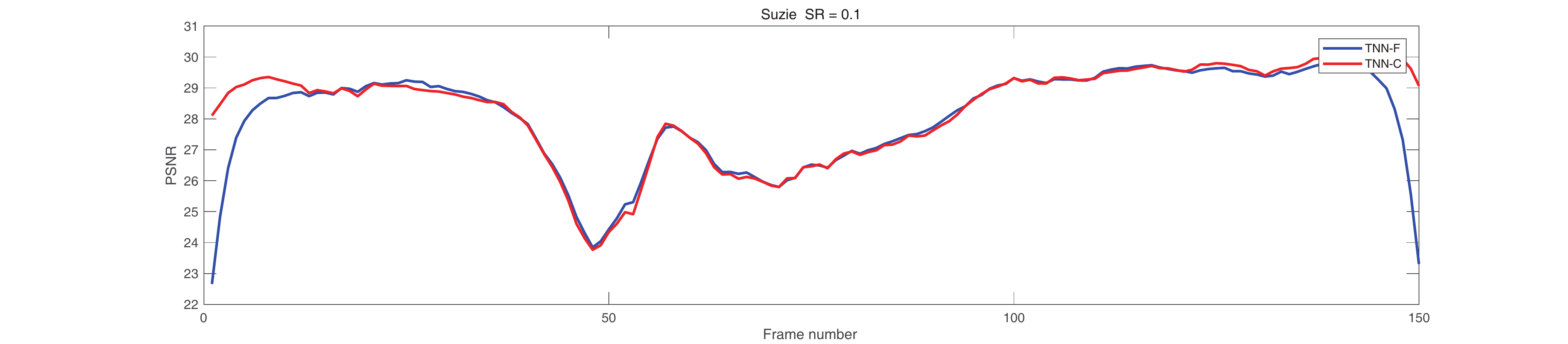}
 \includegraphics[width=1.0\textwidth,height=3.5cm]{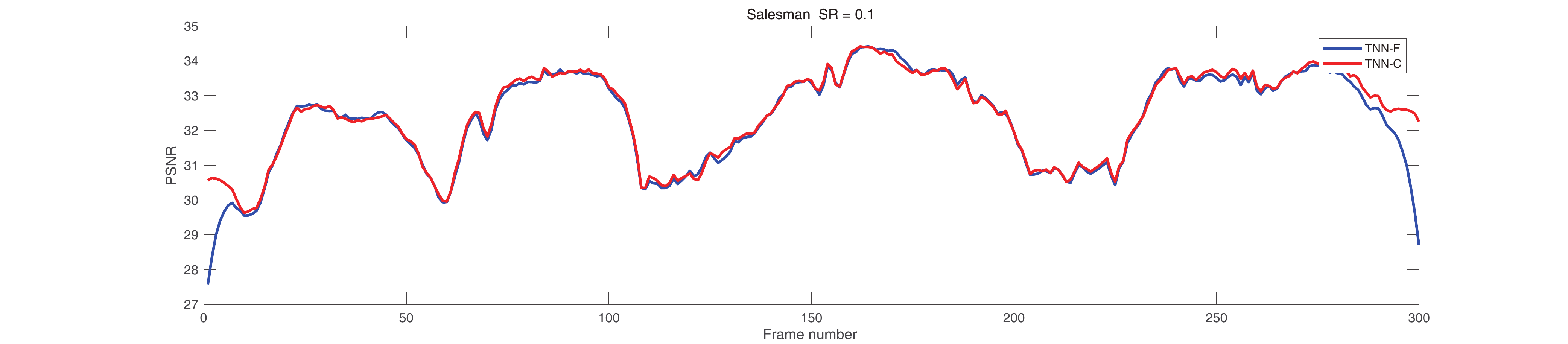}
 \end{center}
 \caption{The PSNR values of each frame of the recovered videos \textit{Akiyo}, \textit{Suzie}, and \textit{Salesman} obtained by TNN-F and TNN-C. }
 \label{figVideoPsnr}
\end{figure}

\begin{figure}
%[!htp]
 \begin{center}
 \includegraphics[width=0.23\textwidth]{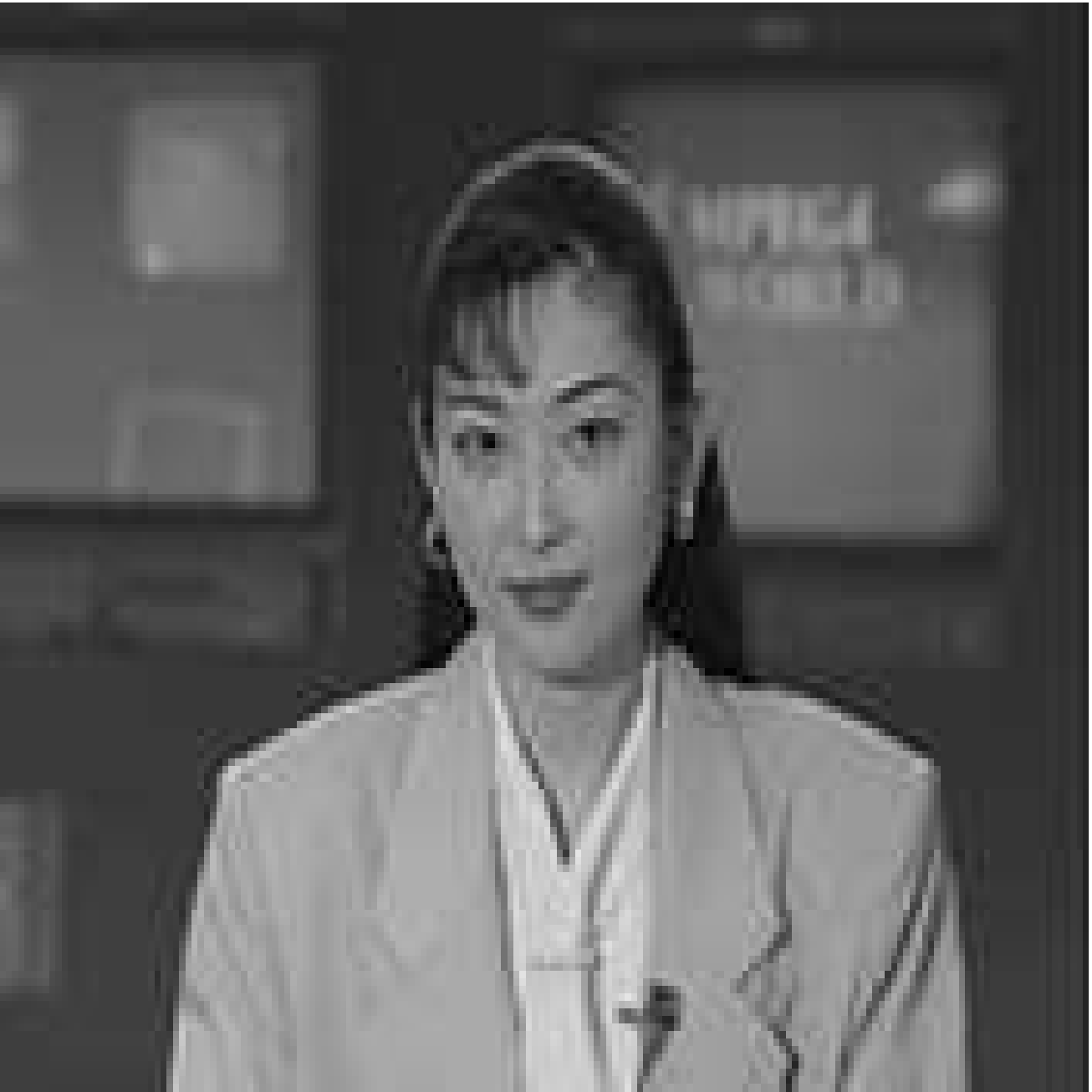}
 \includegraphics[width=0.23\textwidth]{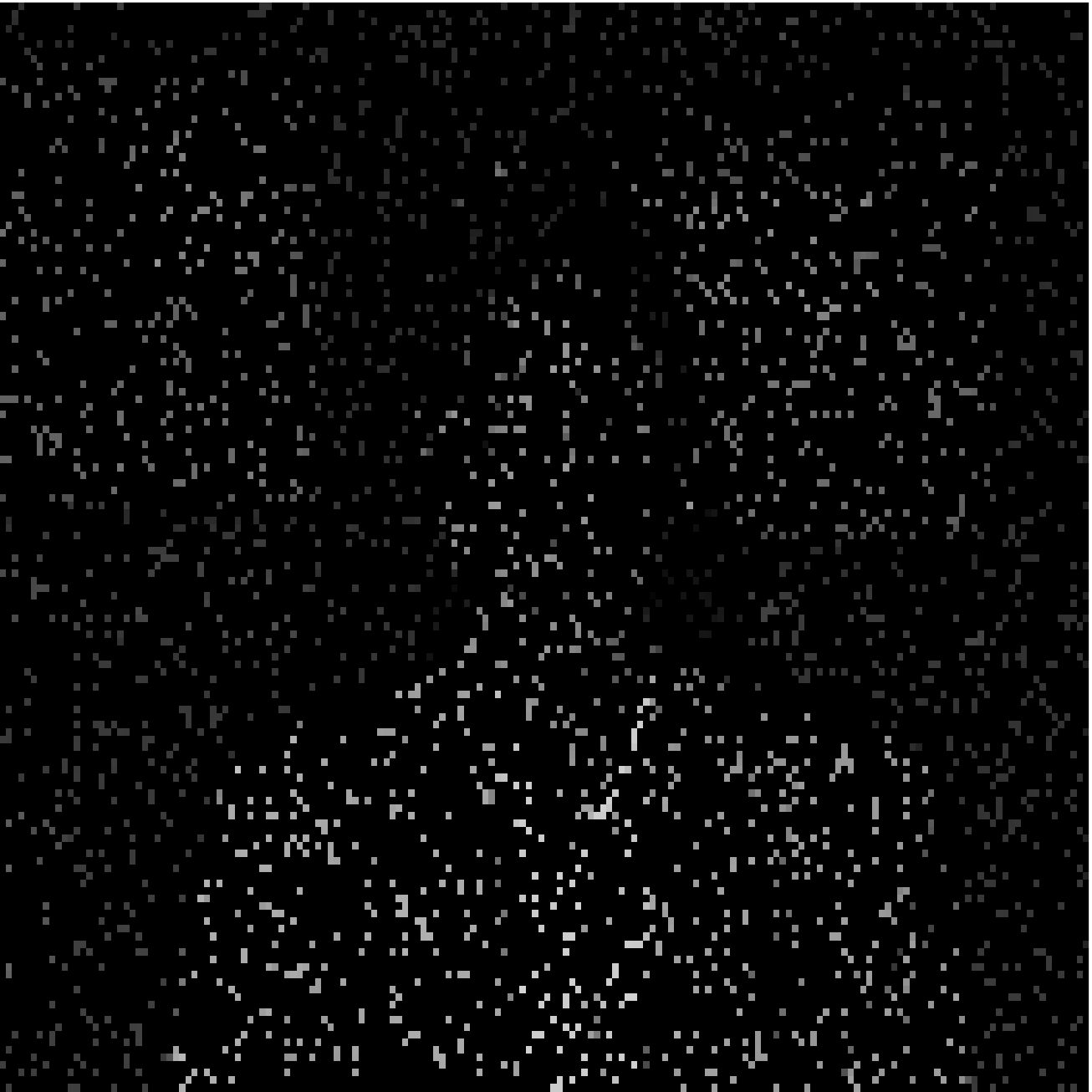}
 \includegraphics[width=0.23\textwidth]{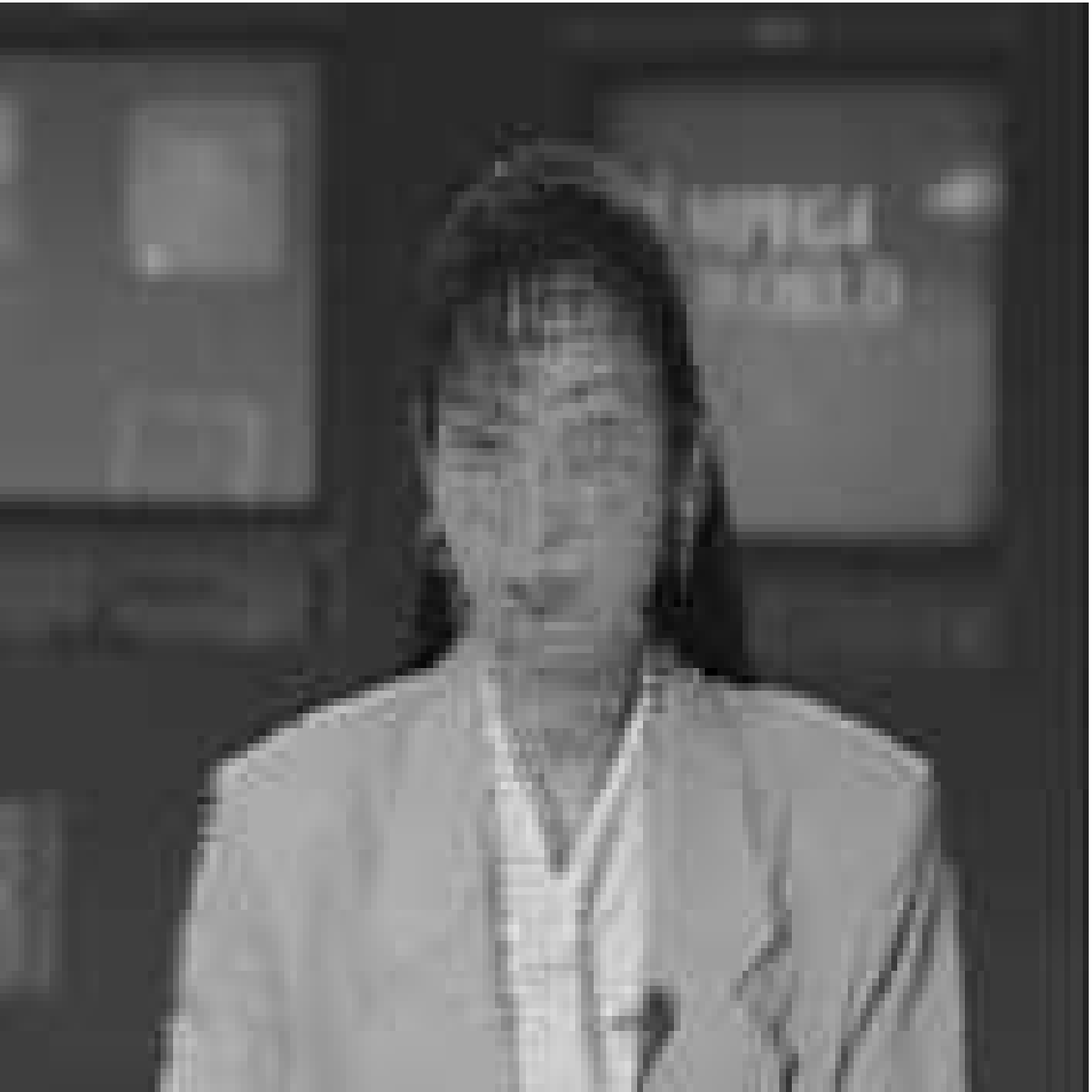}
 \includegraphics[width=0.23\textwidth]{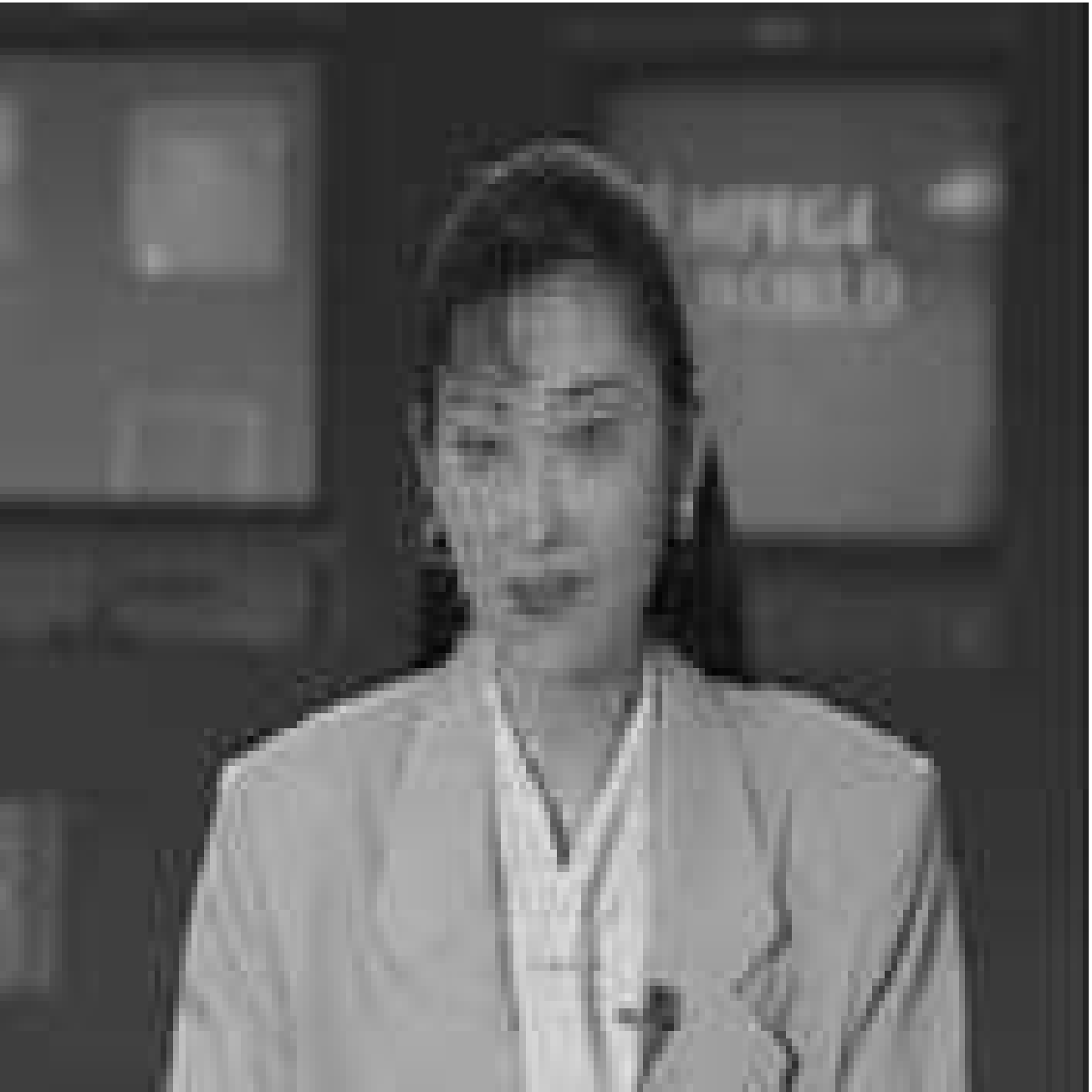}

 \includegraphics[width=0.23\textwidth]{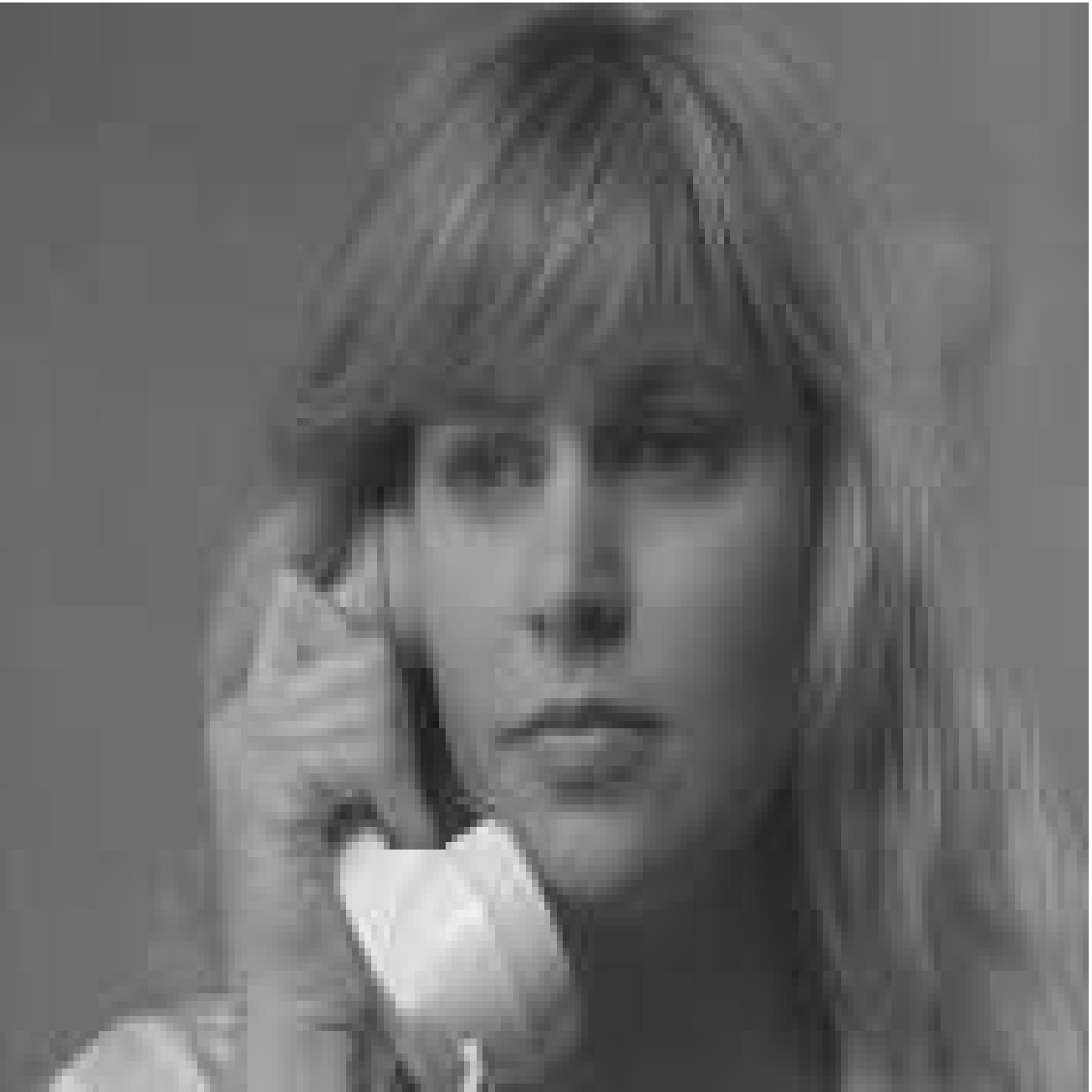}
 \includegraphics[width=0.23\textwidth]{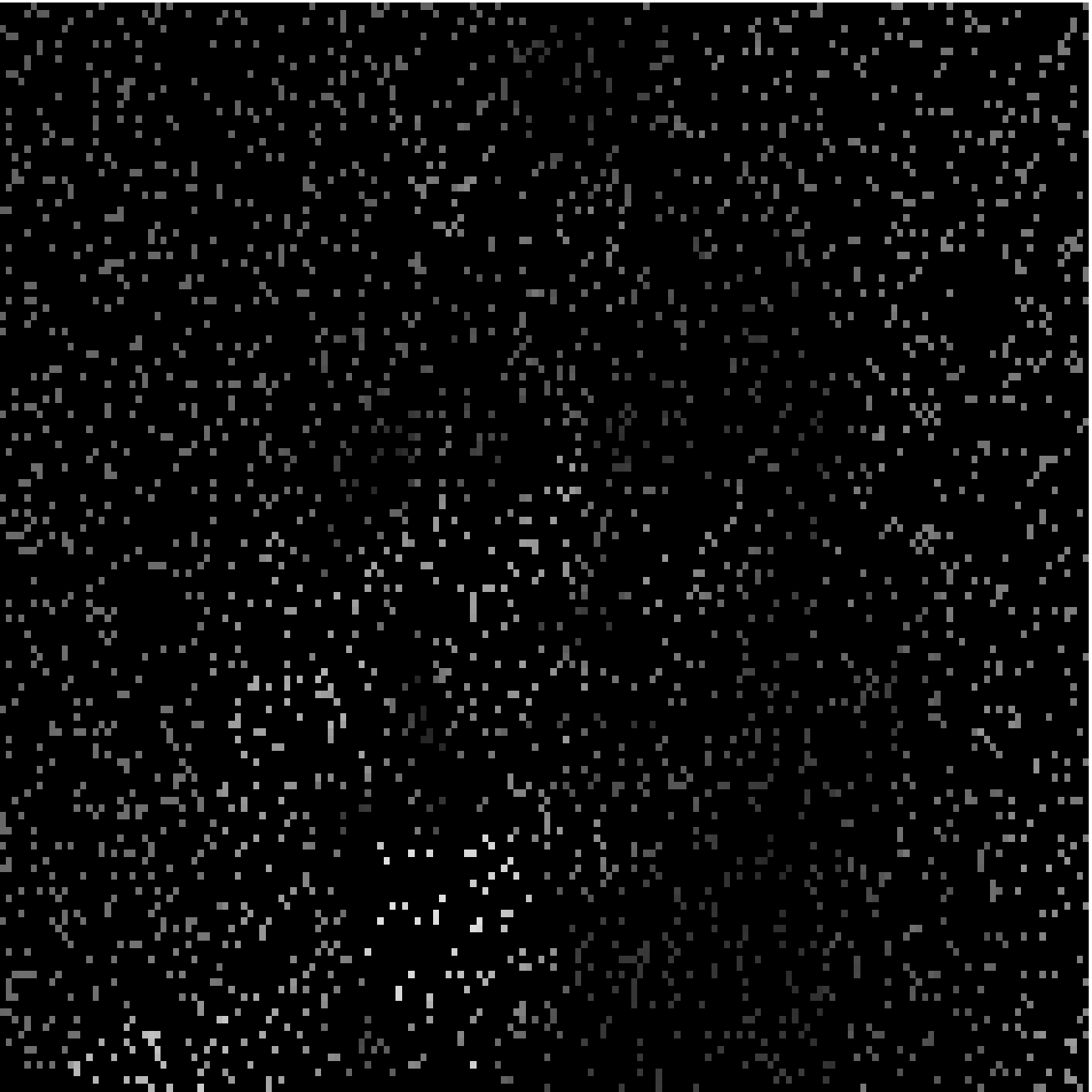}
 \includegraphics[width=0.23\textwidth]{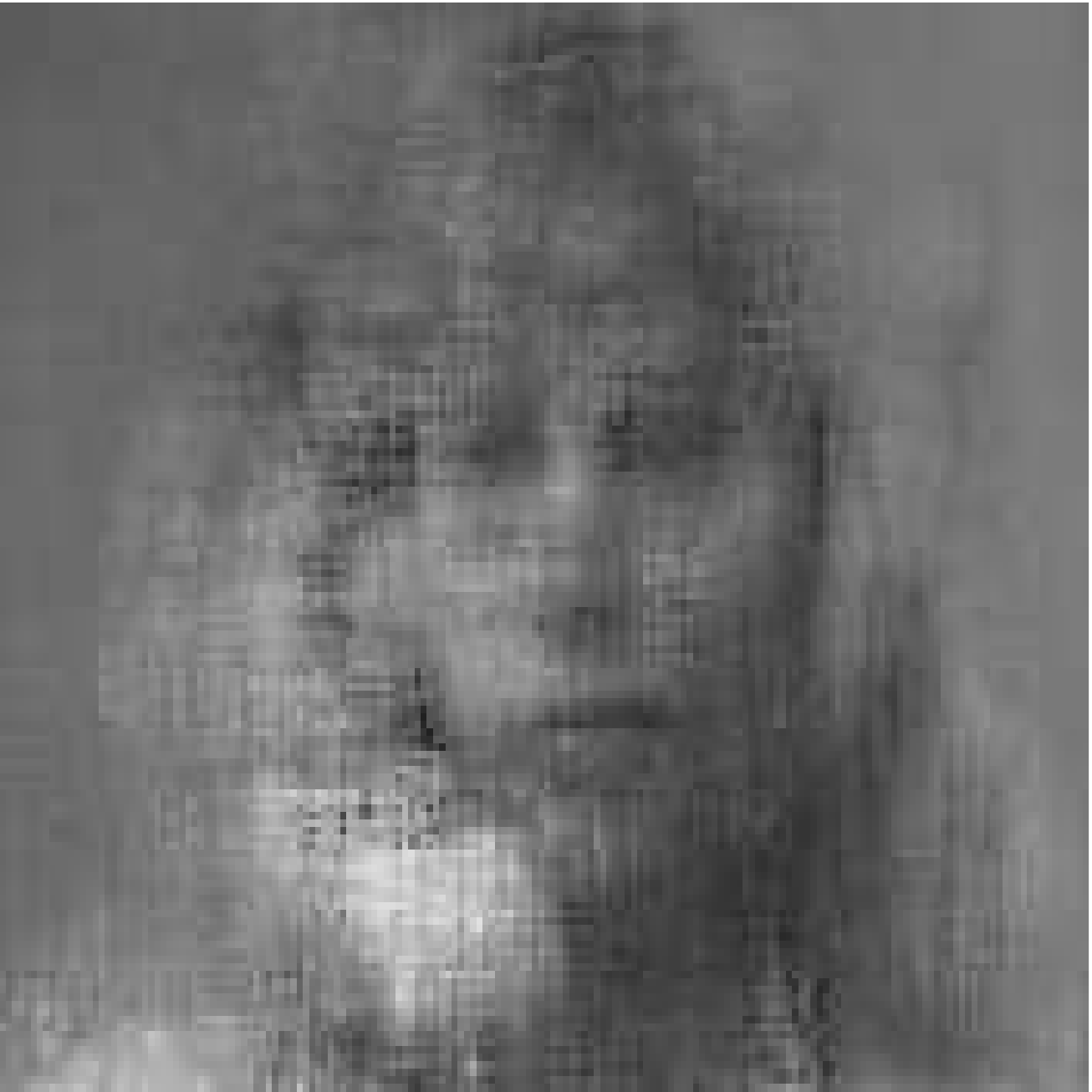}
 \includegraphics[width=0.23\textwidth]{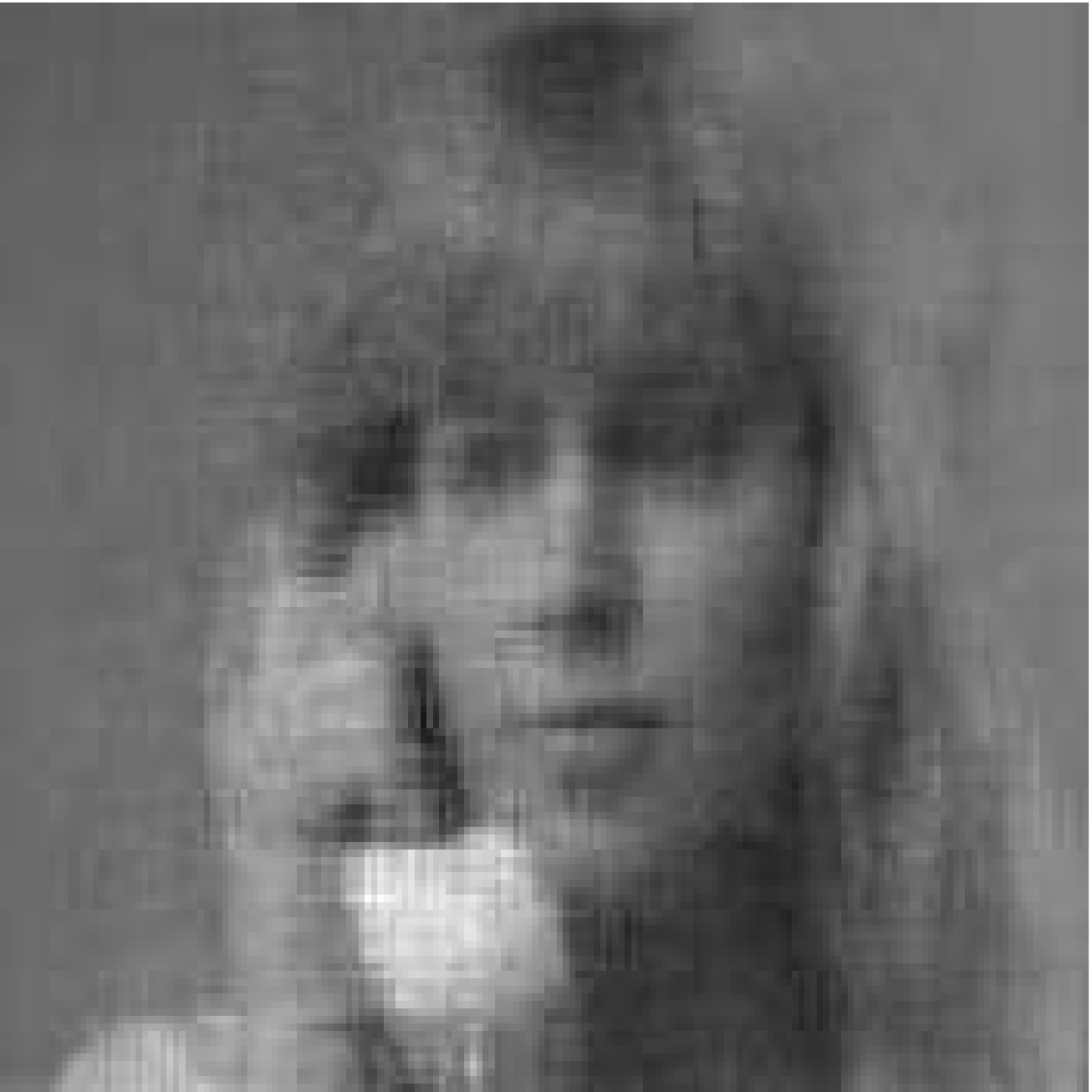}

 \includegraphics[width=0.23\textwidth]{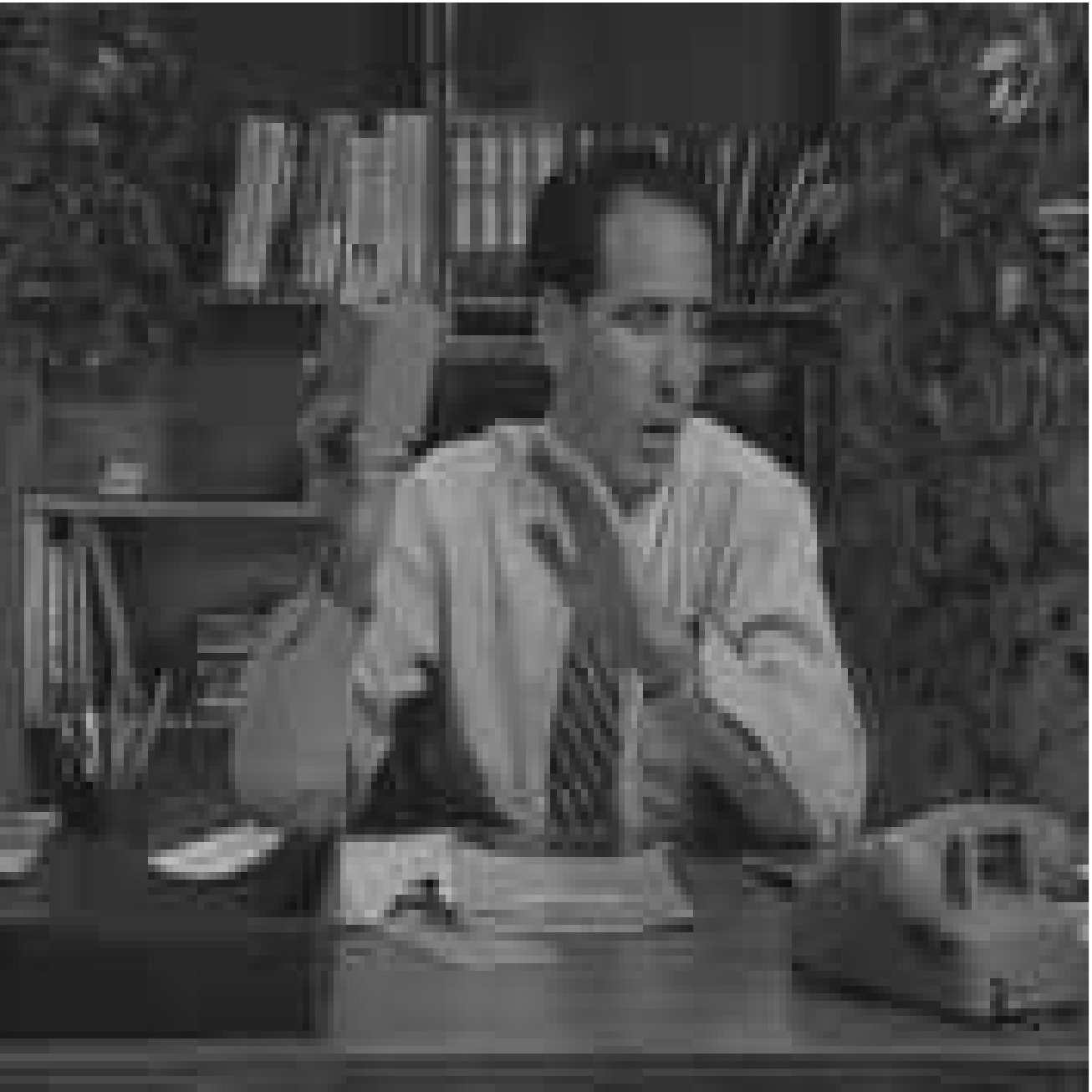}
 \includegraphics[width=0.23\textwidth]{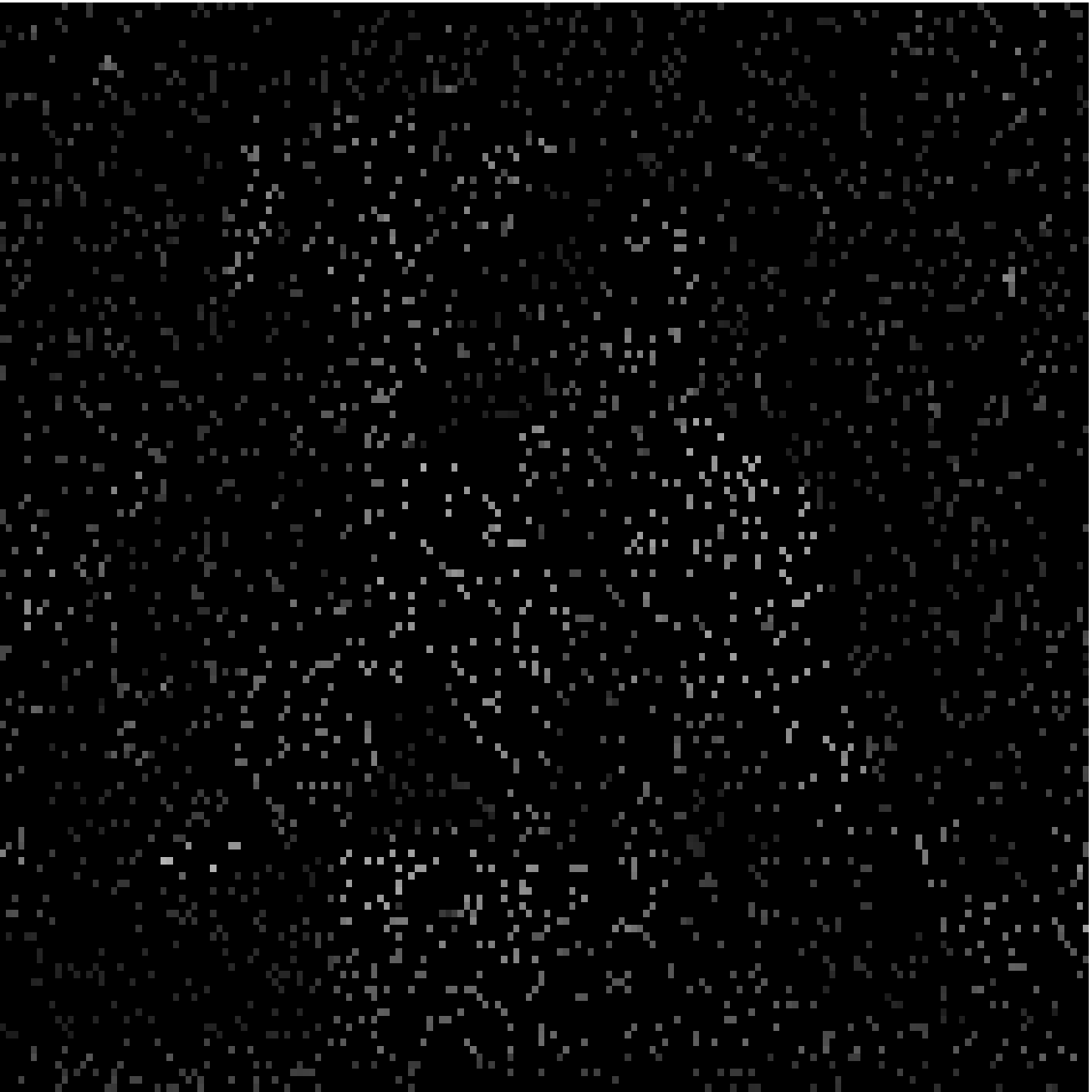}
 \includegraphics[width=0.23\textwidth]{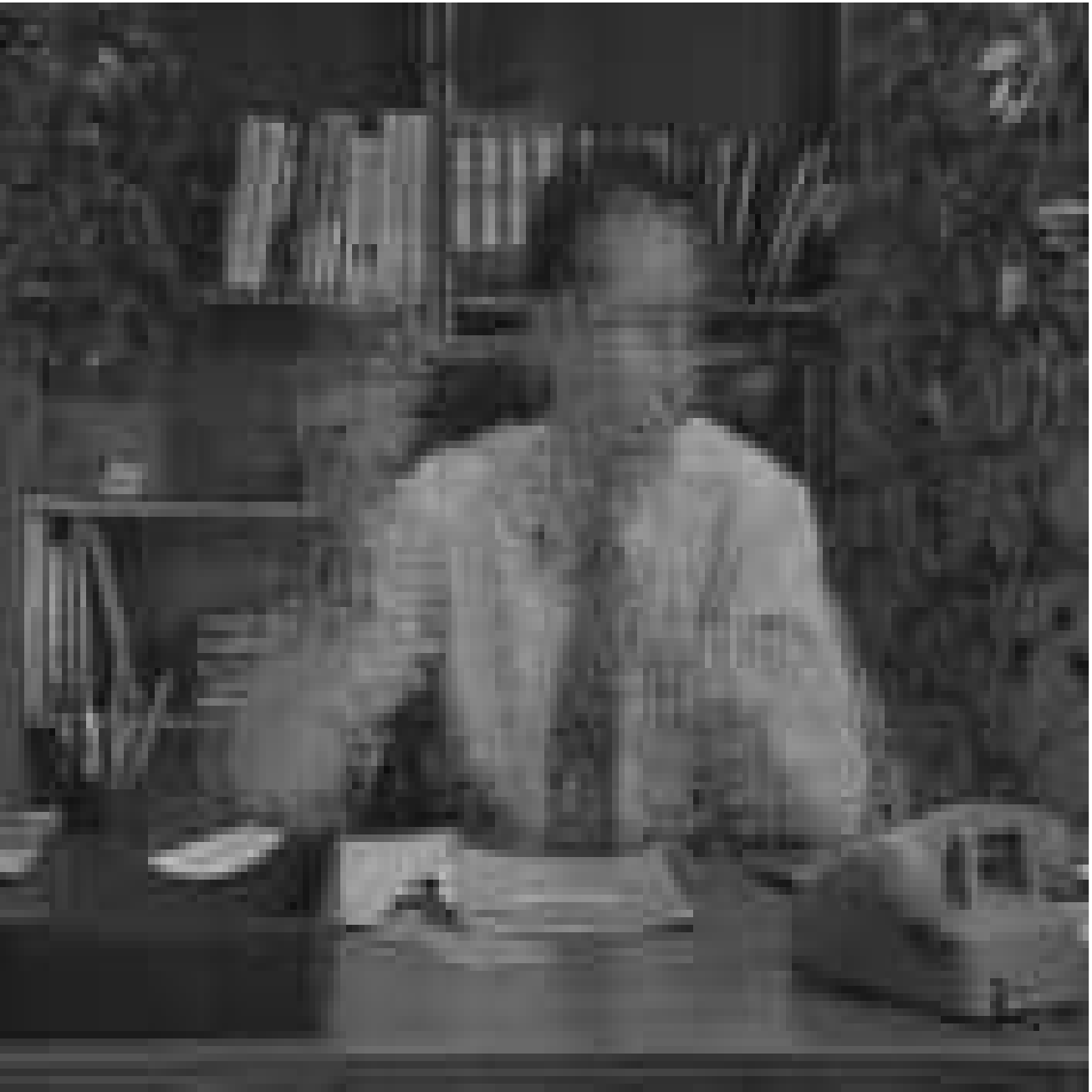}
 \includegraphics[width=0.23\textwidth]{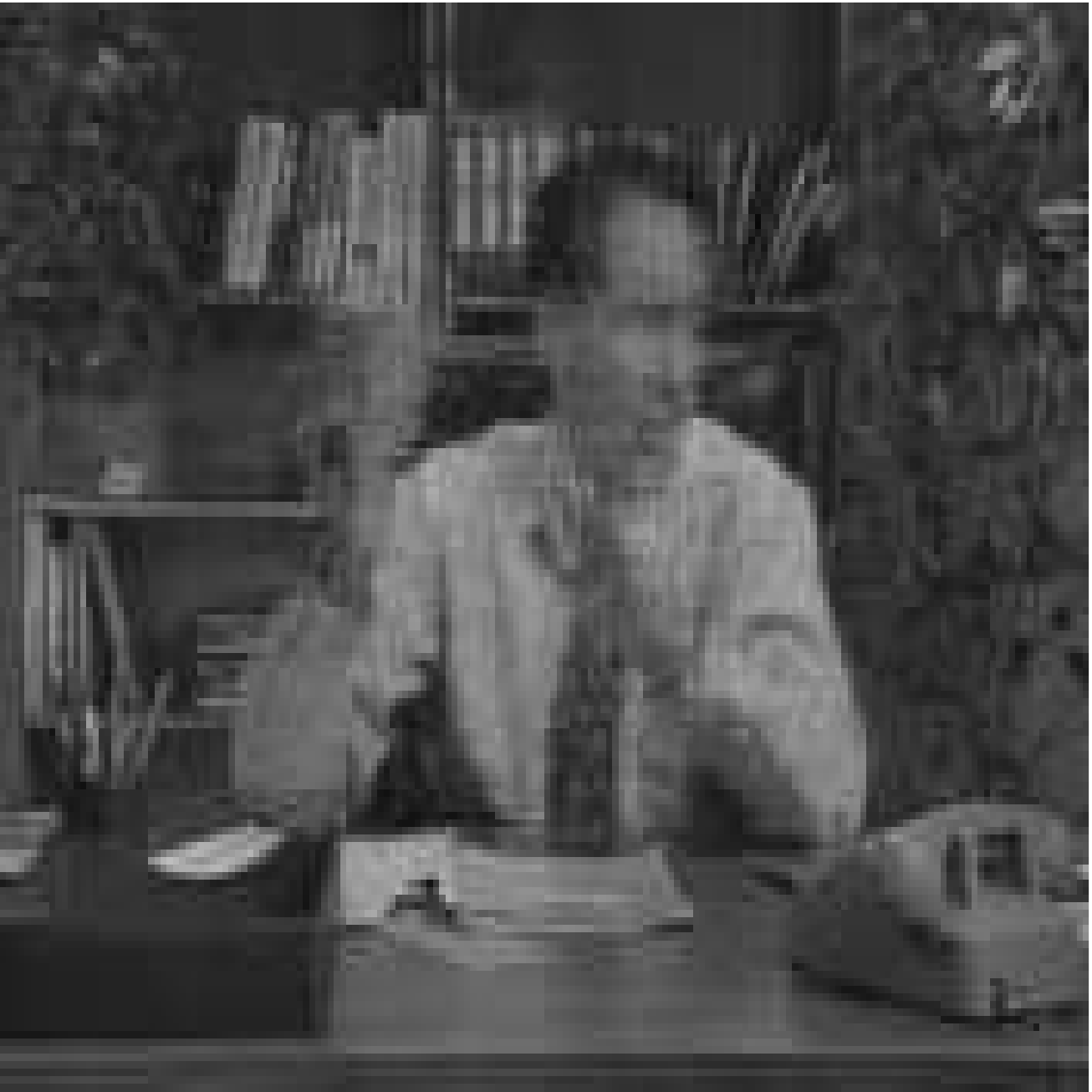}

 \end{center}
 \caption{A frame of the recovered videos with $\text{SR}=0.1$. From top to bottom: \textit{Akiyo}, \textit{Suzie}, and \textit{Salesman}. From left to right: the original image, the masked image, the results by TNN-F, and TNN-C. }
 \label{figTestVideo}
\end{figure}

\textbf{Video completion.} We test 3 videos: \textit{Akiyo}, \textit{Suzie}, and \textit{Salesman}. The size of \textit{Akiyo} and \textit{Salesman} is $144 \times 176 \times 300$. The size of \textit{Suzie} is $144 \times 176 \times 150$. Tab.\thinspace\ref{tabTestVideo} shows PSNR, SSIM, and time cost of TNN-F and TNN-C. TNN-C achieves better results and costs much less time than TNN-F in all experiments. Fig.\thinspace\ref{figVideoTube} shows one selected tube. We can observe that the tube of recovered video by TNN-C is more closely to the true tube than that by TNN-F, especially near the boundary. Fig.\thinspace\ref{figVideoPsnr} shows the PSNR values of each frame of recovered videos by TNN-F and TNN-C. We can observe that when the sampling rate (SR) is $0.1$, the PSNR values of TNN-C are higher than those of TNN-F, especially for the first and last few frames. This observation is consistent with our interpretation of BCs. Fig.\thinspace\ref{figTestVideo} shows the results recovered by TNN-F and TNN-C with $\text{SR} = 0.1$. TNN-C is visually better than TNN-F.

% Table generated by Excel2LaTeX from sheet 'Sheet2'
\begin{table}[htbp]
  \setlength{\abovecaptionskip}{0pt}%
\setlength{\belowcaptionskip}{10pt}%
  \caption{PSNR, SSIM, and time of two methods in video completion. In brackets, they are the time required for transformation and time required for performing SVD. The best results are highlighted in bold.}
    %\small
  \centering

    %\renewcommand\arraystretch{0.9}
    %\resizebox{\textwidth}{24mm}{
\begin{tabular}{|cc|cc|cc|cc|}
%cc}
\hline
\multicolumn{2}{|c|}{video} & \multicolumn{2}{c|}{\textit{akiyo}} & \multicolumn{2}{c|}{\textit{suzie}} & \multicolumn{2}{c|}{\textit{salesman}}
%& \multicolumn{2}{c}{average}
\bigstrut\\
\hline
SR    & metric & TNN-F & TNN-C & TNN-F & TNN-C & TNN-F & TNN-C \\
%& TNN-F & TNN-C \bigstrut\\
\hline
\hline
\multirow{4}[2]{*}{0.05} & PSNR  & 32.00  & \textbf{32.57 } & 25.50  & \textbf{26.02 } & 30.12  & \textbf{30.22 }
% 29.21  & \textbf{29.60 }
 \bigstrut[t]\\
      & SSIM  & 0.934  & \textbf{0.941 } & 0.681  & \textbf{0.700 } & 0.895  & \textbf{0.897 }
      %& 0.837  & \textbf{0.846 }
      \\
      & \multirow{2}[1]{*}{time} & 156.2  & \textbf{91.9 } & 69.6  & \textbf{40.1 } & 148.5  & \textbf{85.6 } \\
      %& 124.8  & \textbf{72.5 } \\
      &       & (8.8+137.0) & \textbf{(6.2+70.9)} & (4.0+60.6) & \textbf{(2.9+30.6)} & (8.6+128.9) & \textbf{(6.0+65.3)}
      %& (7.1+108.8) & \textbf{(5.0+55.6)}
      \bigstrut[b]\\
\hline
\multirow{4}[2]{*}{0.1} & PSNR  & 34.20  & \textbf{34.75 } & 27.73  & \textbf{27.93 } & 32.13  & \textbf{32.29 }
%& 31.35  & \textbf{31.66 }
\bigstrut[t]\\
      & SSIM  & 0.958  & \textbf{0.963 } & 0.759  & \textbf{0.766 } & 0.928  & \textbf{0.931 }
      %& 0.882  & \textbf{0.887 }
      \\
      & \multirow{2}[1]{*}{time} & 141.8  & \textbf{86.3 } & 64.5  & \textbf{39.3 } & 139.5  & \textbf{84.9 }
      %& 115.3  & \textbf{70.2 }
      \\
      &       & (8.1+122.9) & \textbf{(5.8+66.6)} & (3.8+55.2) & \textbf{(2.8+30.2)} & (8.3+120.3) & \textbf{(5.8+64.9)}
      %& (6.7+99.5) & \textbf{(4.8+53.9)}
      \bigstrut[b]\\
\hline
\multirow{4}[2]{*}{0.2} & PSNR  & 37.44  & \textbf{38.11 } & 30.29  & \textbf{30.51 } & 35.01  & \textbf{35.20 }
%& 34.25  & \textbf{34.61 }
\bigstrut[t]\\
      & SSIM  & 0.979  & \textbf{0.983 } & 0.838  & \textbf{0.844 } & 0.960  & \textbf{0.961 } \\
      %& 0.926  & \textbf{0.929 } \\
      & \multirow{2}[1]{*}{time} & 145.2  & \textbf{79.8 } & 62.5  & \textbf{37.2 } & 135.1  & \textbf{81.3 } \\
      %& 114.3  & \textbf{66.1 } \\
      &       & (8.1+125.6) & \textbf{(5.4+60.3)} & (3.6+53.3) & \textbf{(2.8+28.6)} & (8.1+116.3) & \textbf{(5.5+61.6)}
      %& (6.6+98.4) & \textbf{(4.6+50.2)}
      \bigstrut[b]\\
\hline
\end{tabular}%
  %  }
  \label{tabTestVideo}%
\end{table}%

\begin{figure}
%[!htp]
 \begin{center}
 \includegraphics[width=0.4\textwidth]{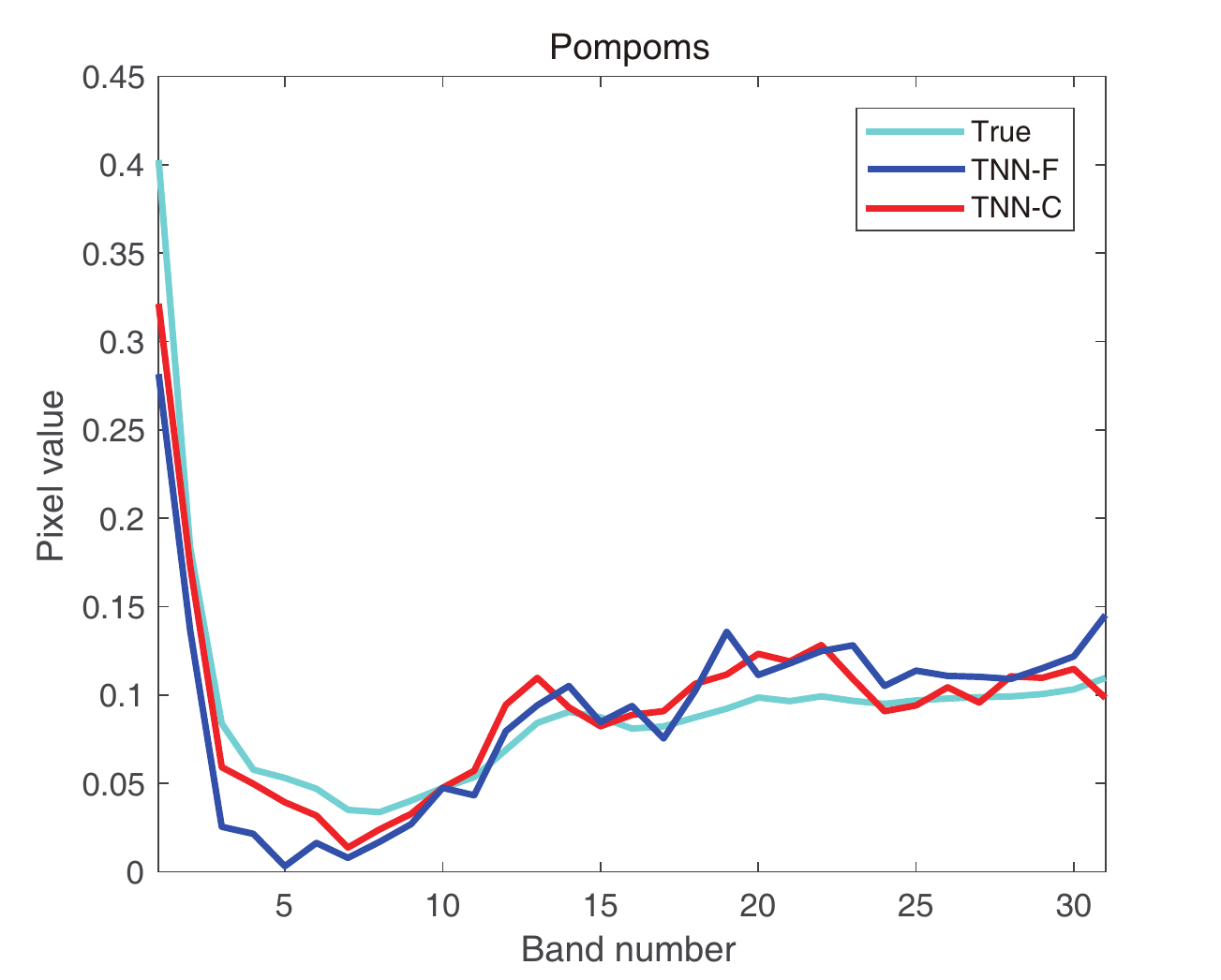}
 \includegraphics[width=0.4\textwidth]{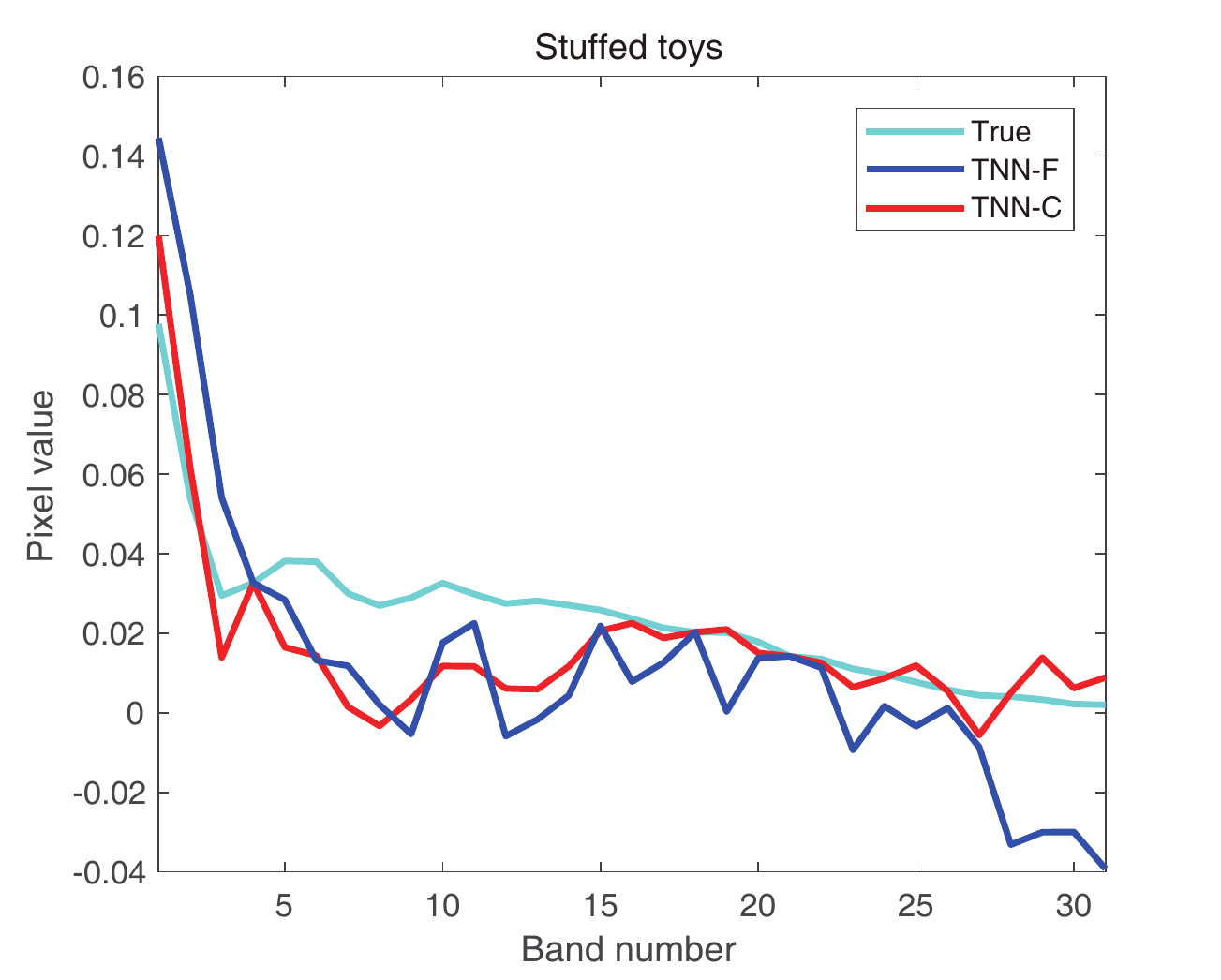}
 \includegraphics[width=0.4\textwidth]{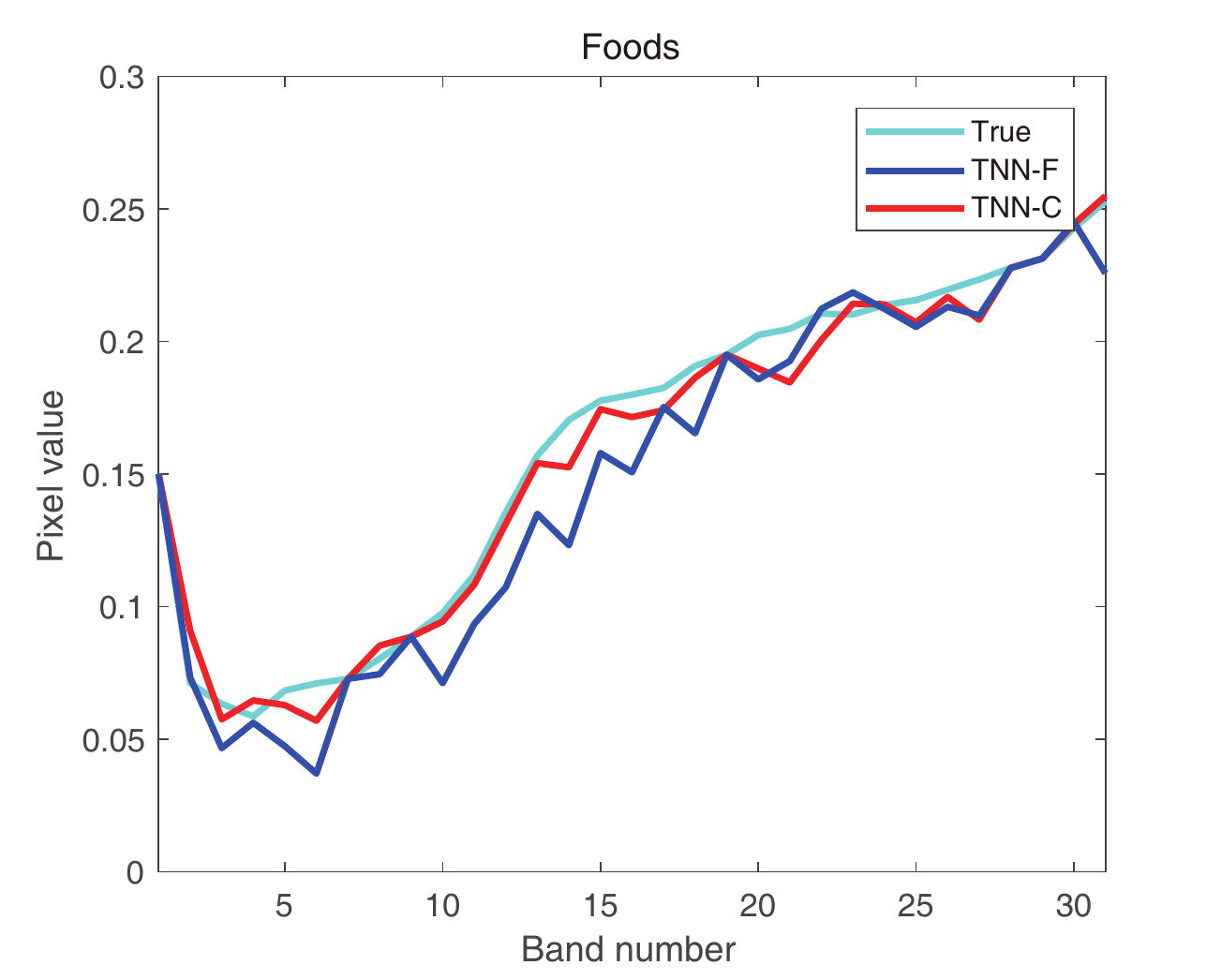}
 \includegraphics[width=0.4\textwidth]{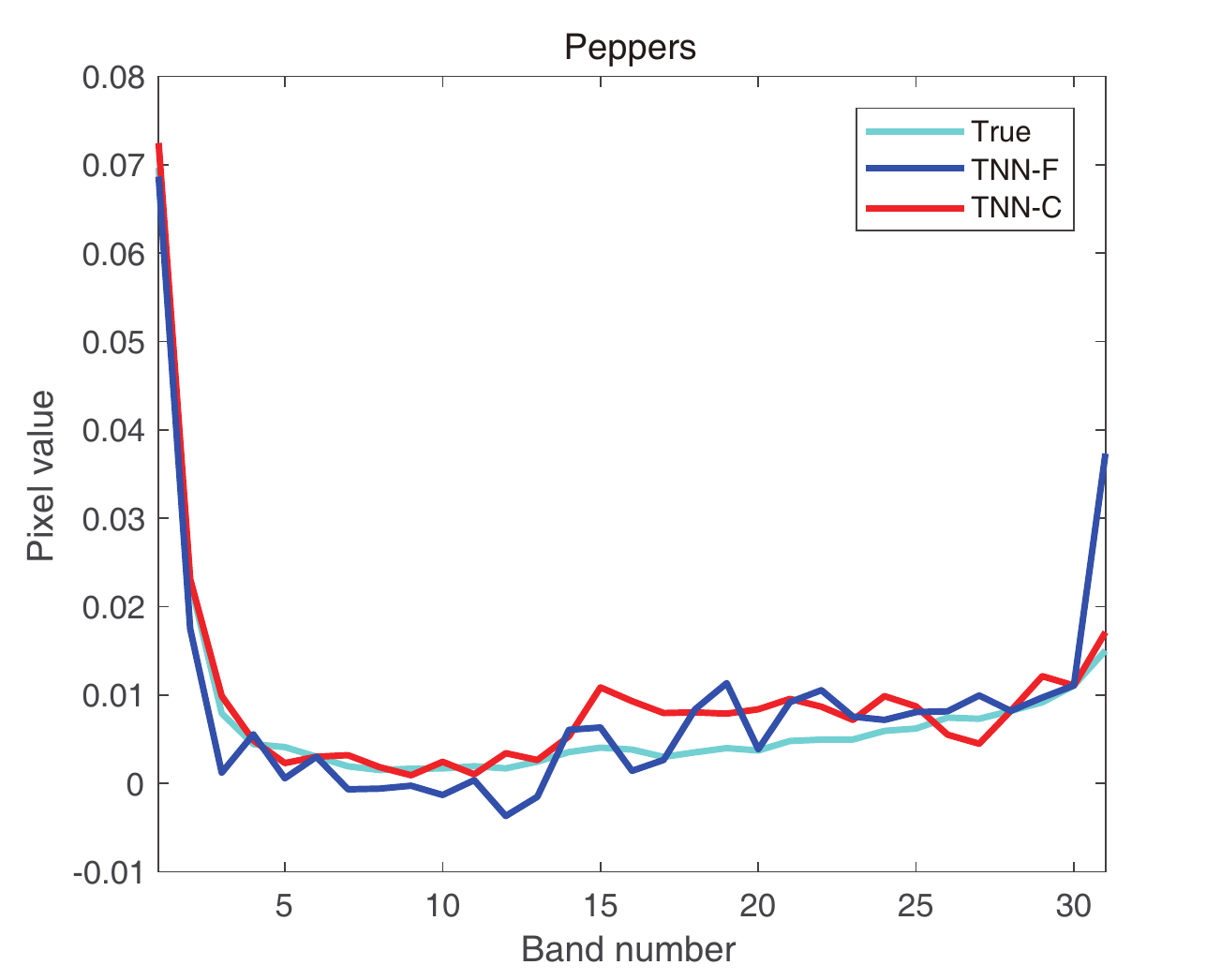}
 \end{center}
 \caption{The pixel values of a random tube of MSI \textit{Pompoms}, \textit{Stuffed toys}, \textit{Foods}, and \textit{Peppers}. }
 \label{figMsiTube}
\end{figure}

\begin{figure}
%[!htp]
 \begin{center}
 \includegraphics[width=0.4\textwidth]{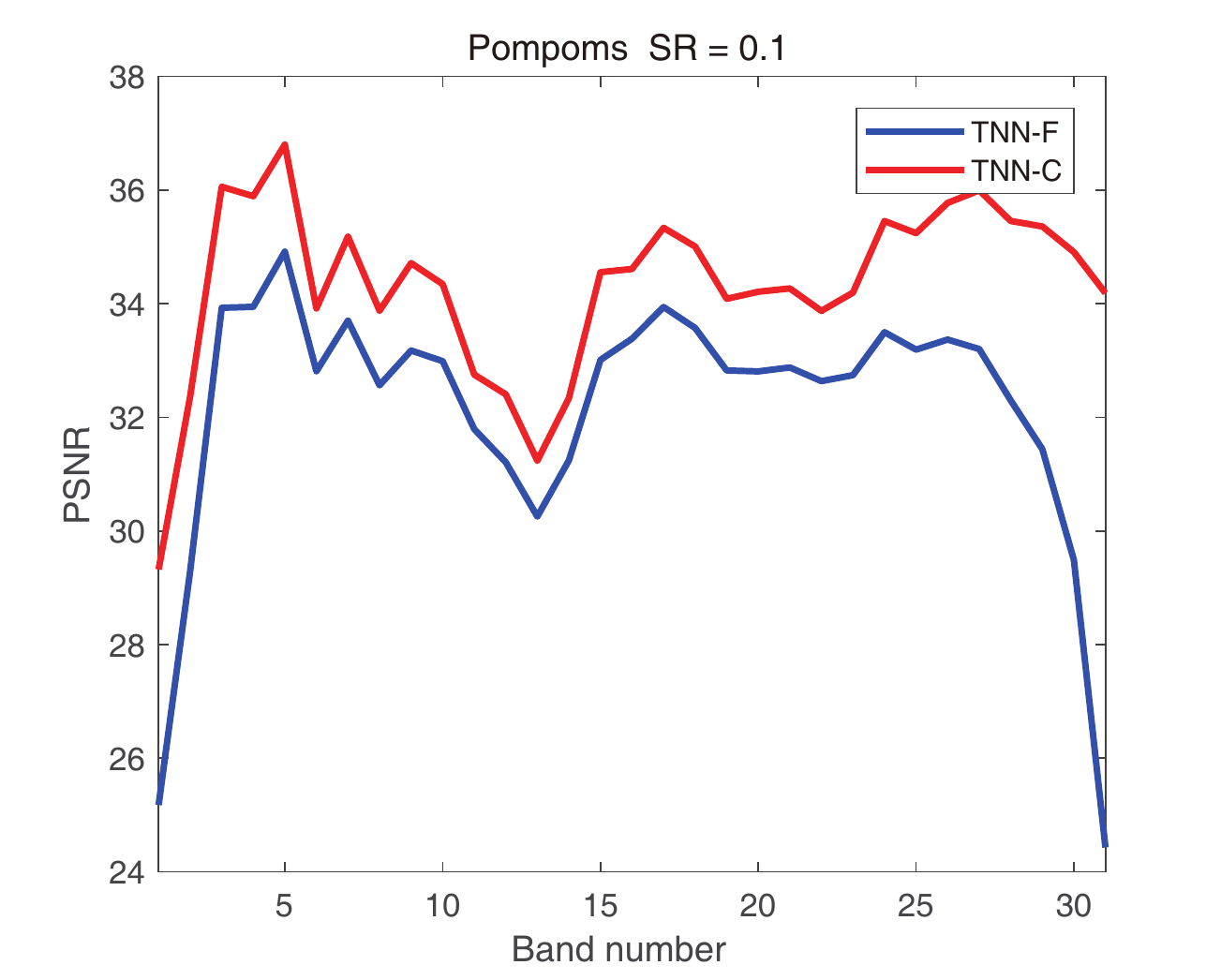}
 \includegraphics[width=0.4\textwidth]{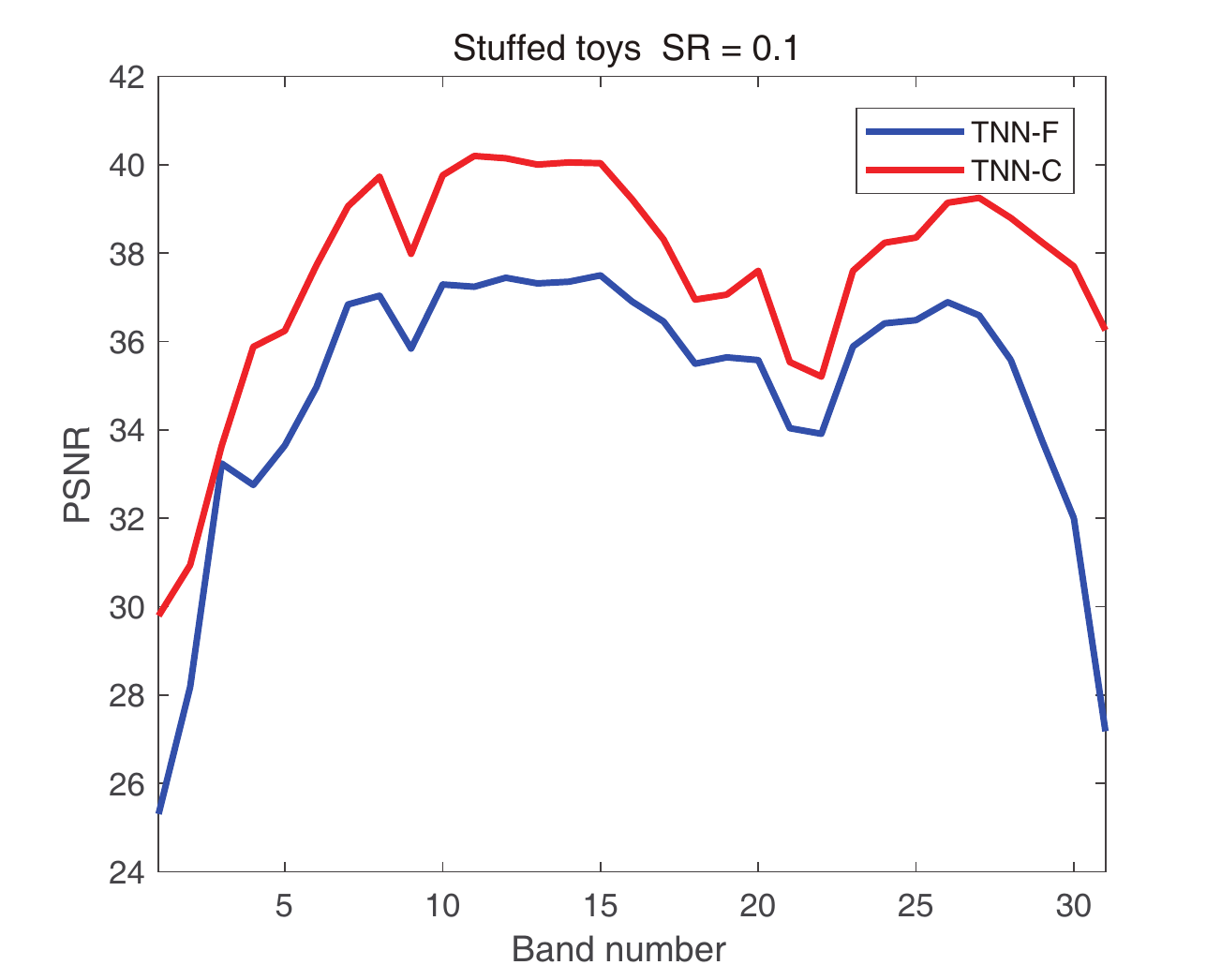}
 \includegraphics[width=0.4\textwidth]{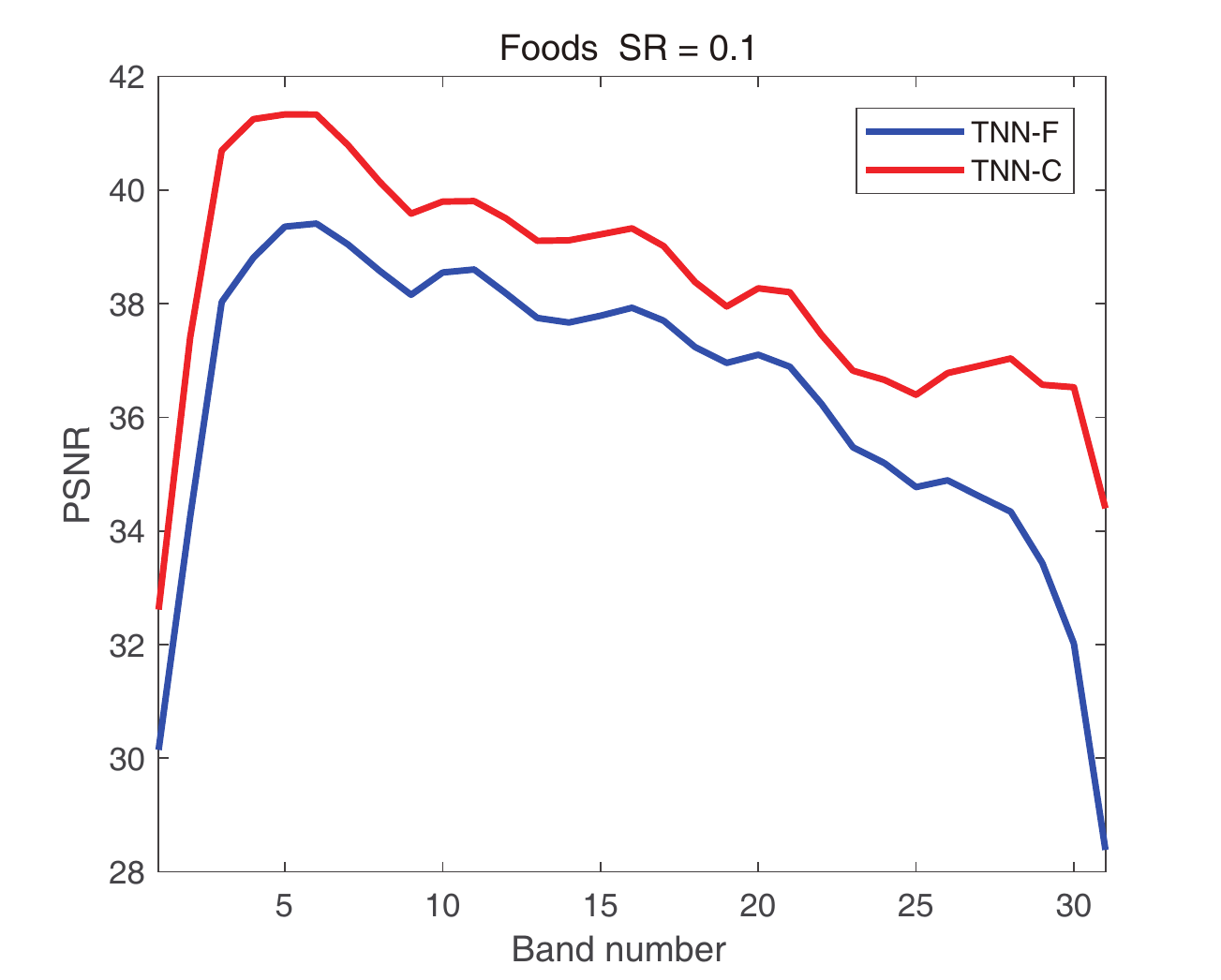}
 \includegraphics[width=0.4\textwidth]{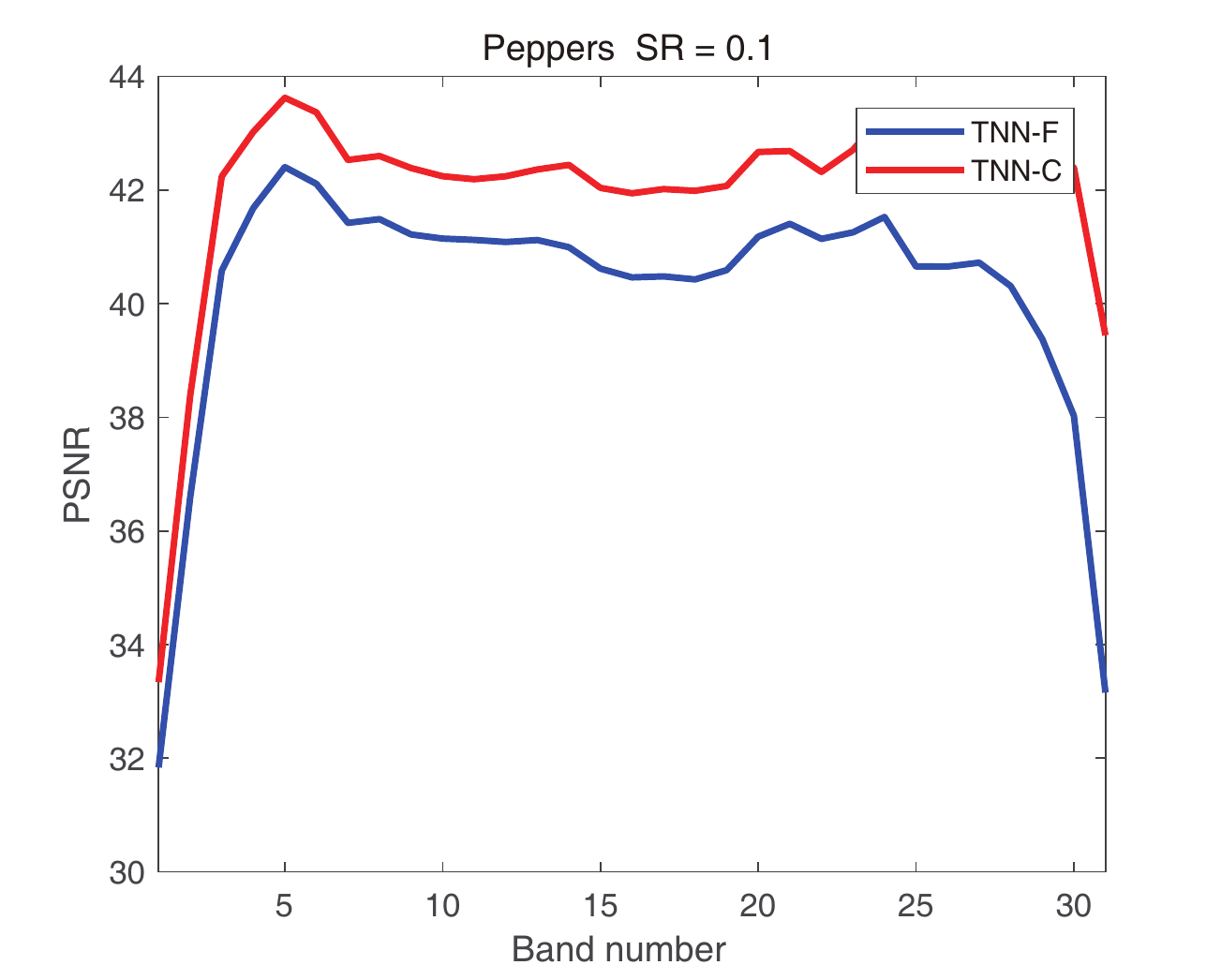}
 \end{center}
 \caption{The PSNR values of each band of the recovered MSIs \textit{Pompoms}, \textit{Stuffed toys}, \textit{Foods}, and \textit{Peppers} obtained by TNN-F and TNN-C. }
 \label{figMsiPsnr}
\end{figure}

\begin{figure}
%[!htp]
 \begin{center}
 \includegraphics[width=0.23\textwidth]{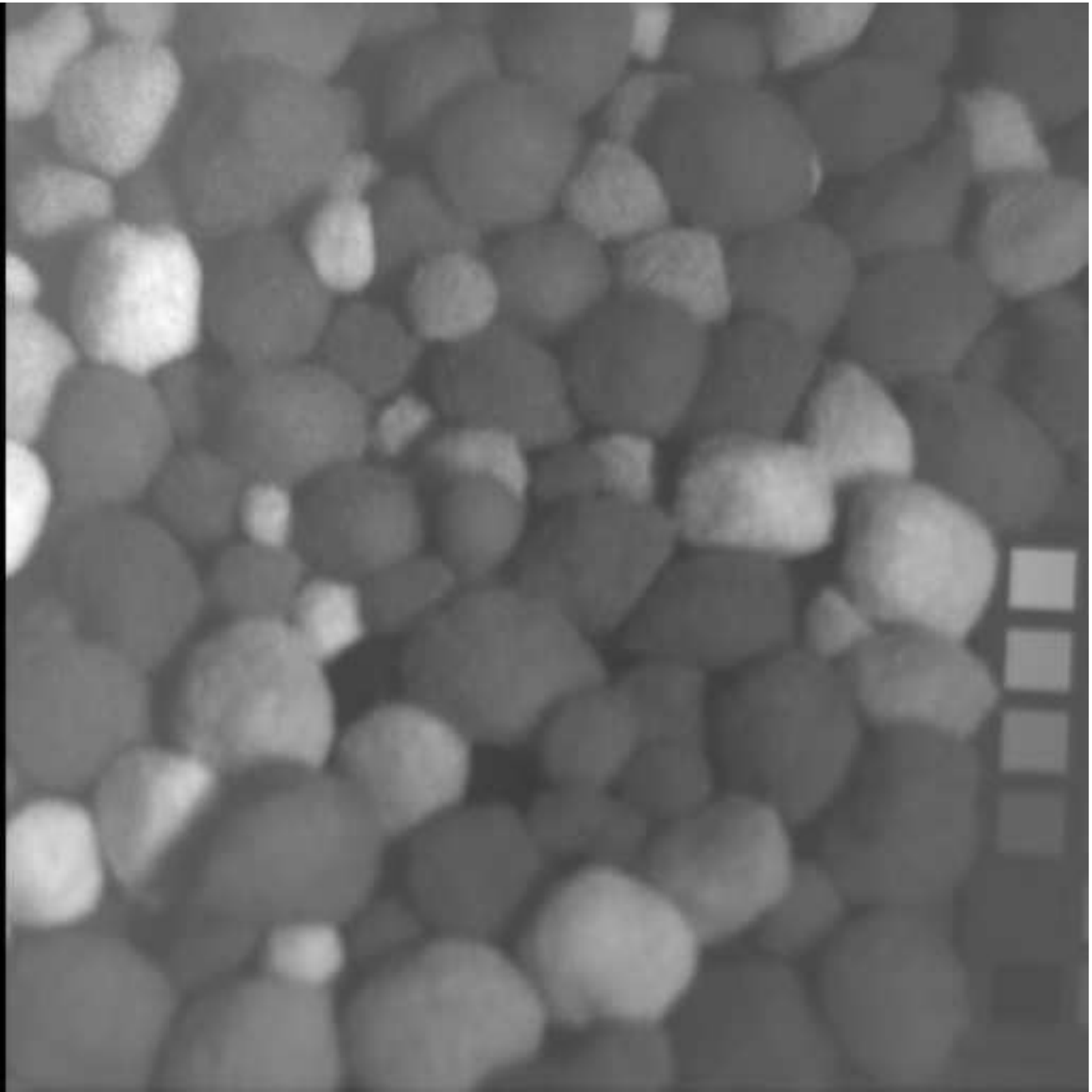}
 \includegraphics[width=0.23\textwidth]{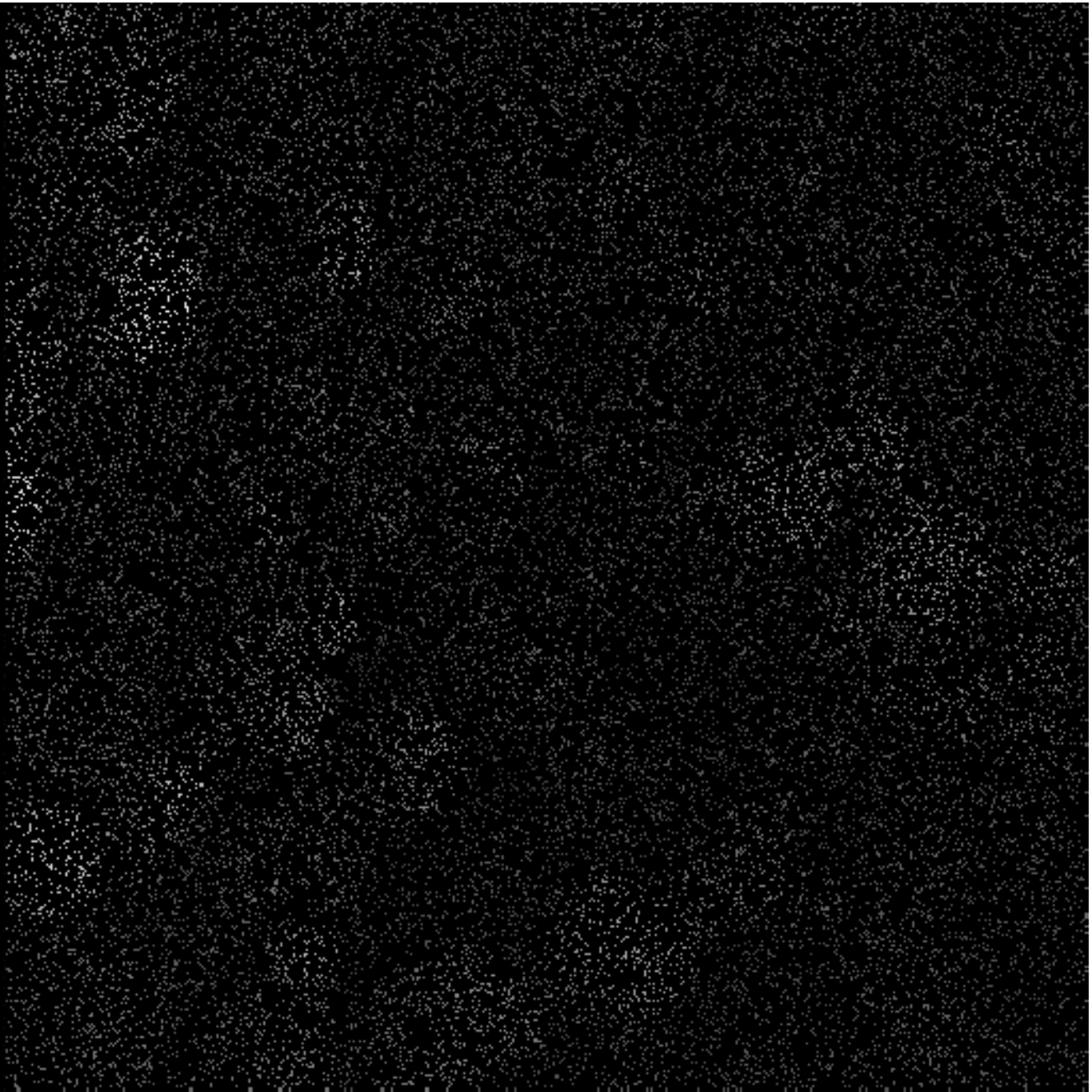}
 \includegraphics[width=0.23\textwidth]{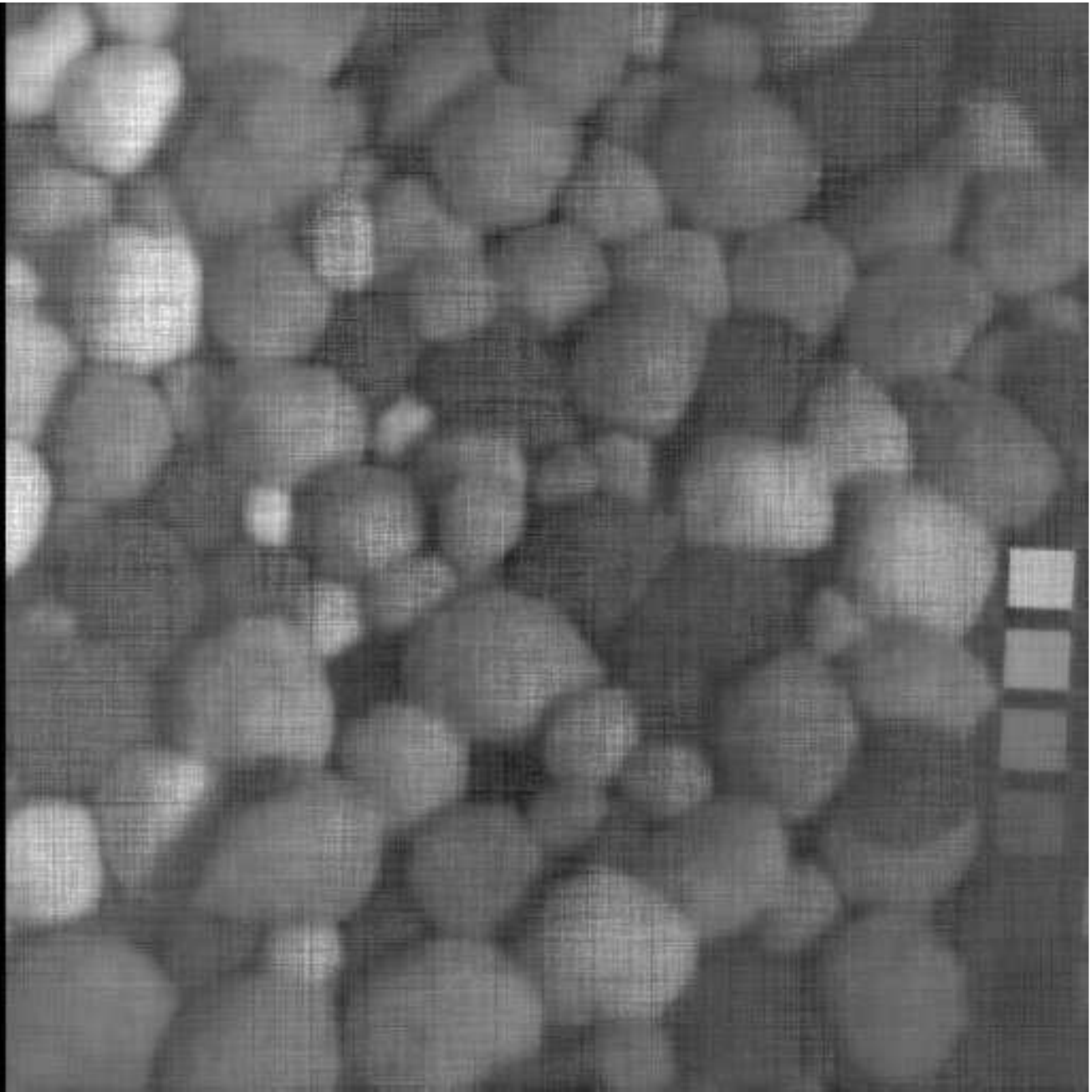}
 \includegraphics[width=0.23\textwidth]{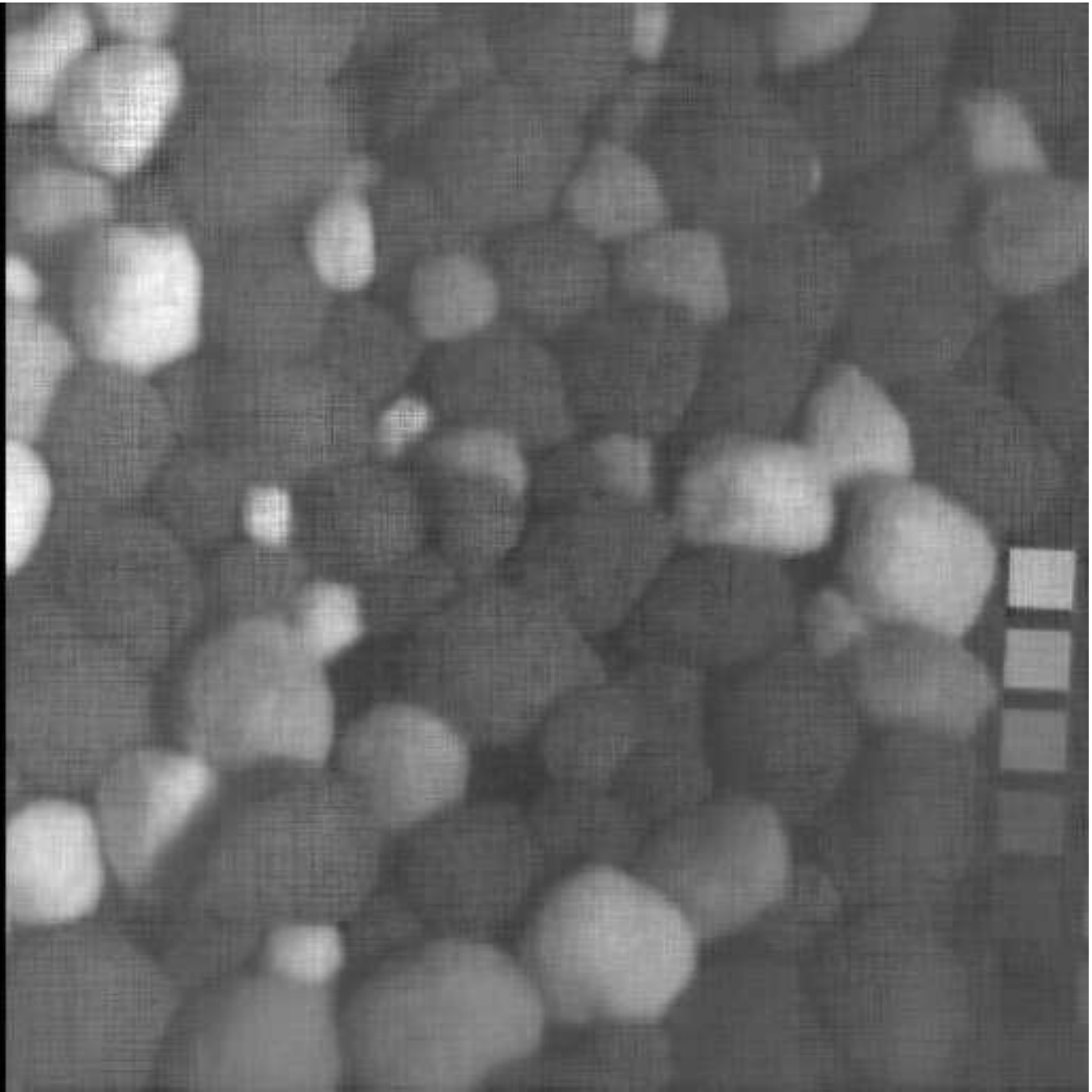}

 \includegraphics[width=0.23\textwidth]{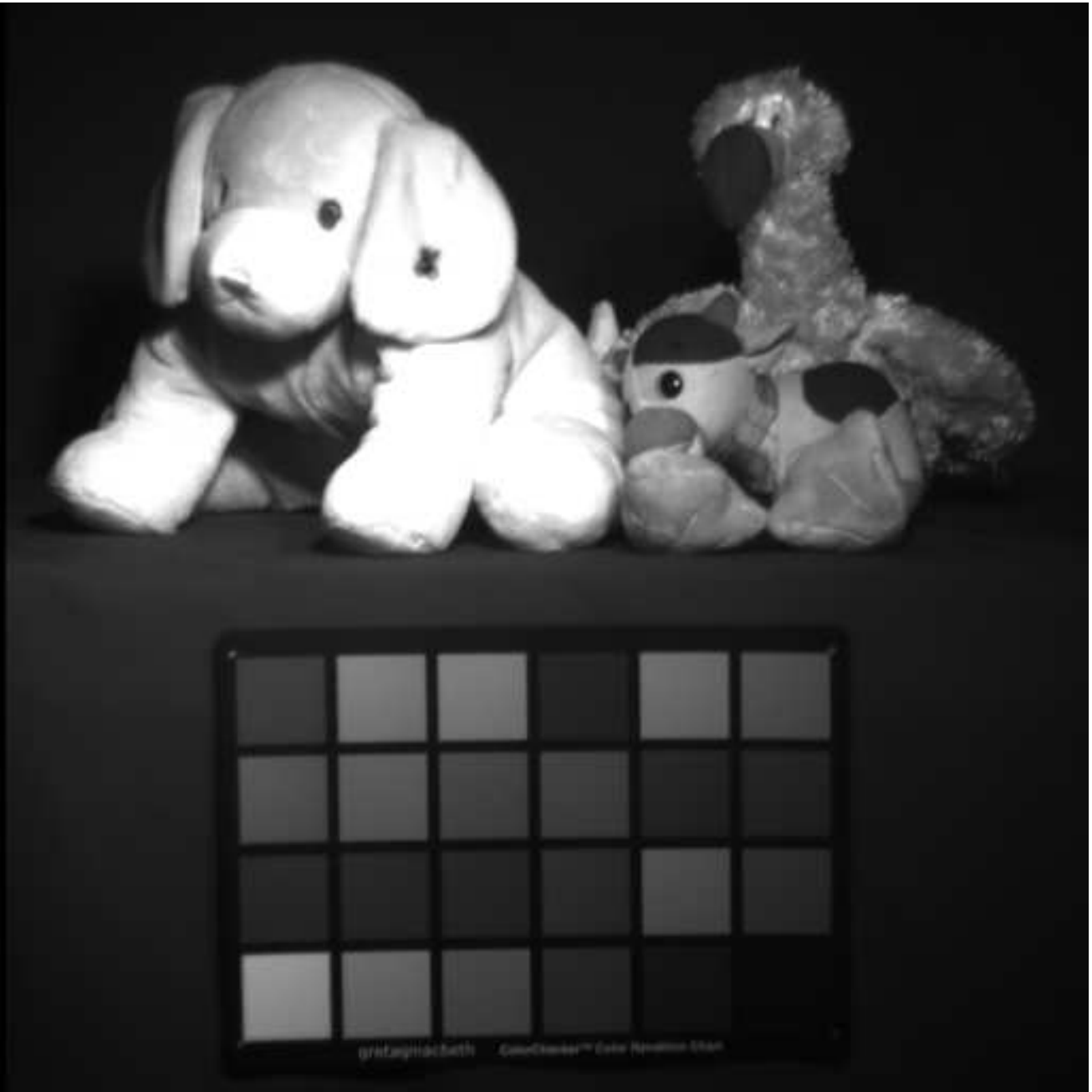}
 \includegraphics[width=0.23\textwidth]{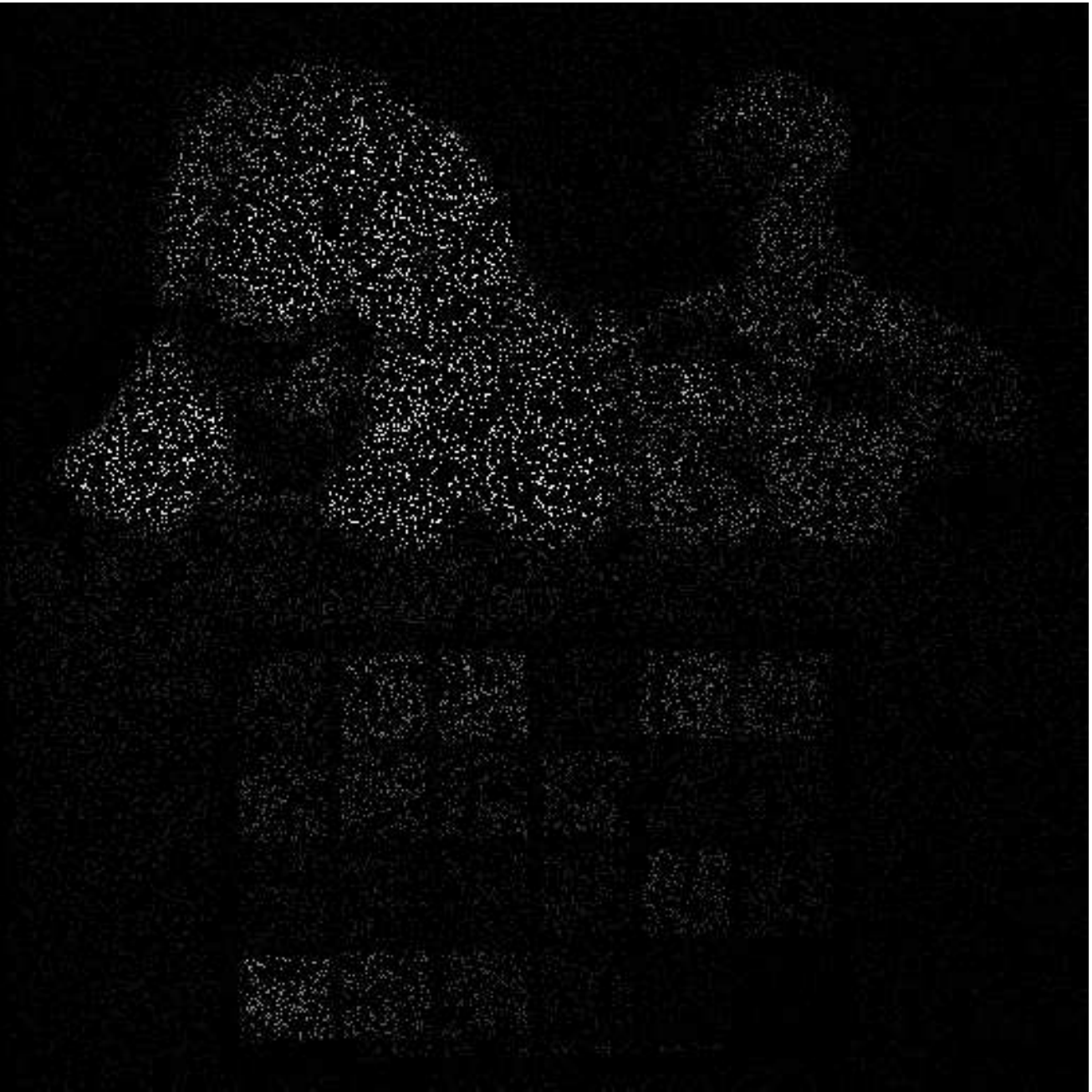}
 \includegraphics[width=0.23\textwidth]{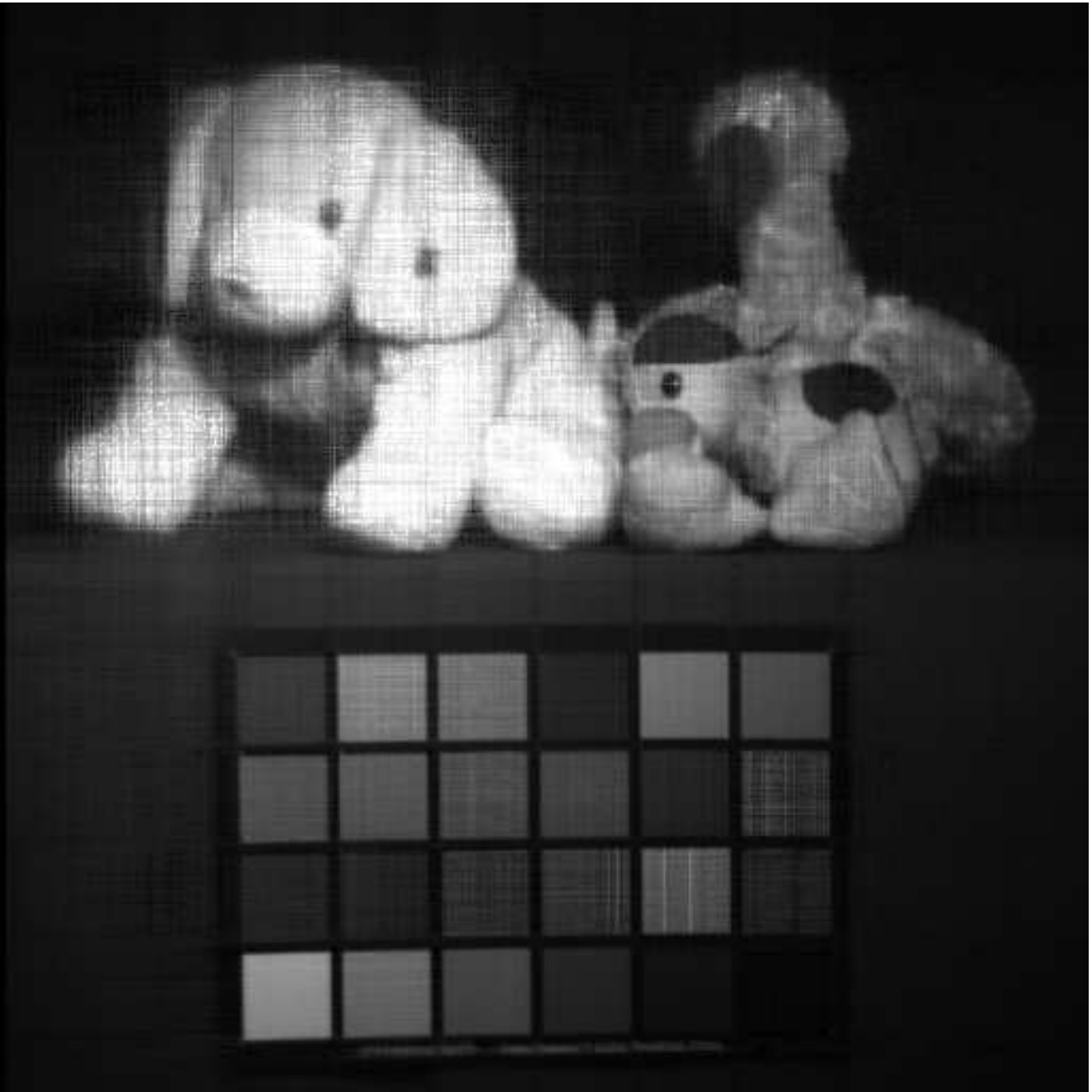}
 \includegraphics[width=0.23\textwidth]{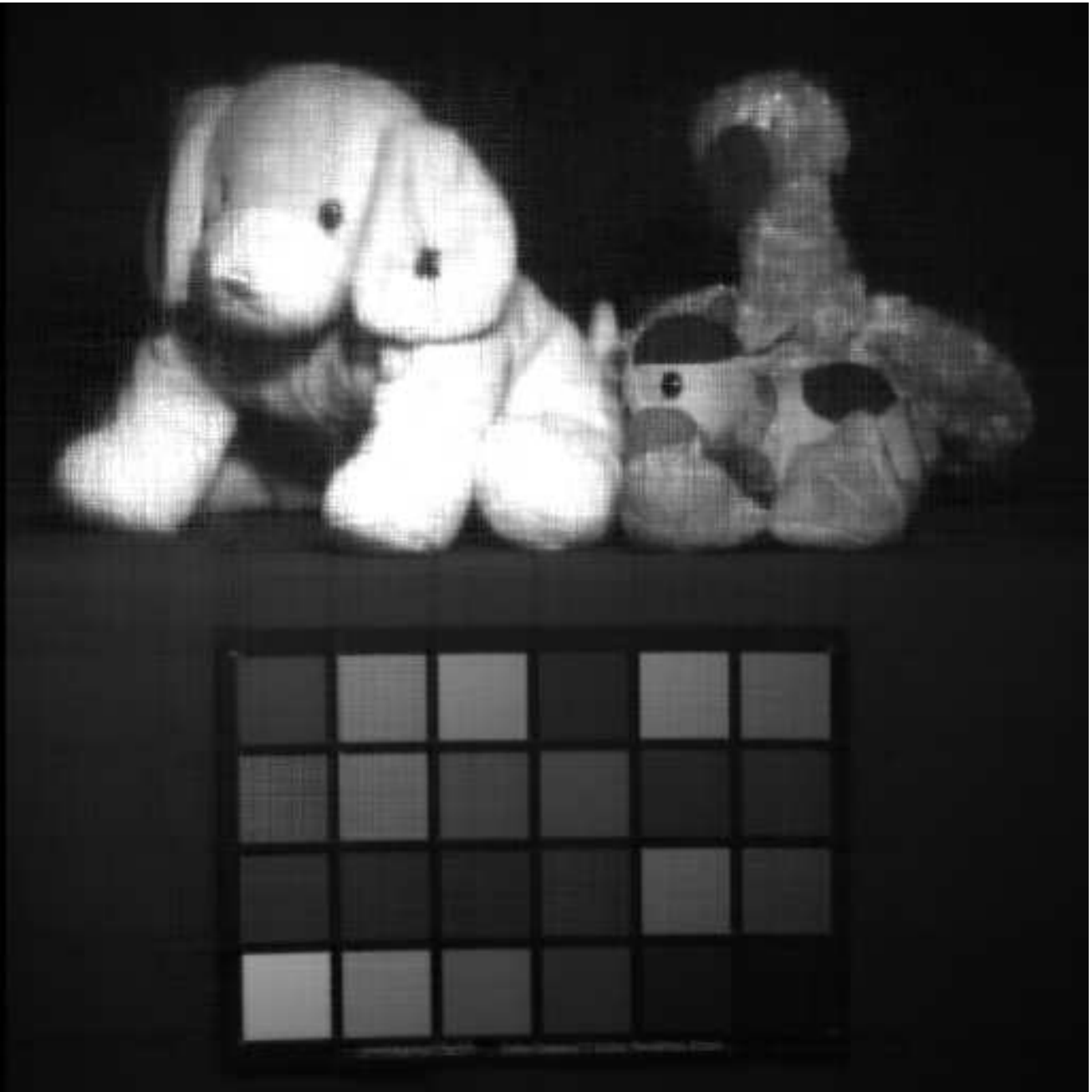}

 \includegraphics[width=0.23\textwidth]{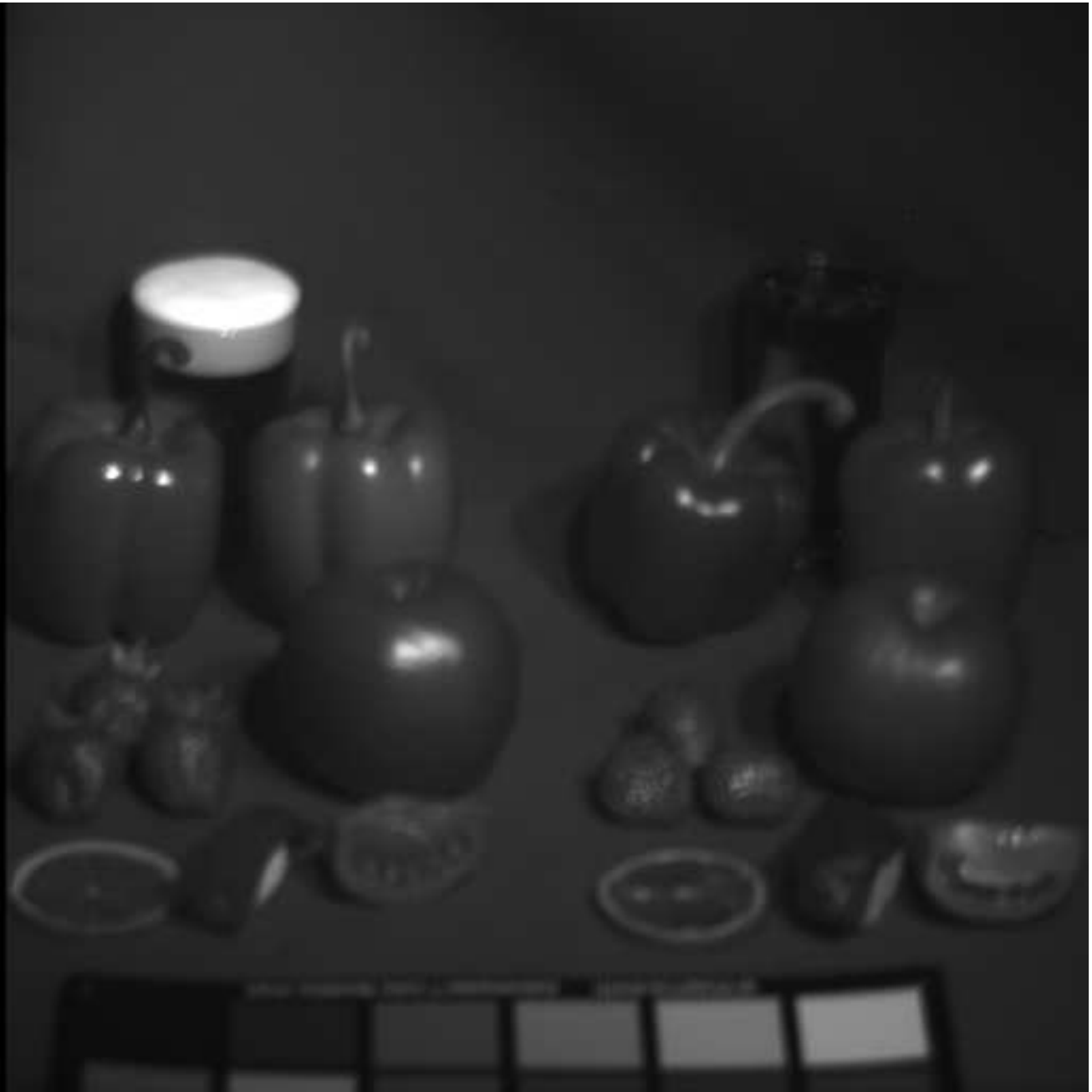}
 \includegraphics[width=0.23\textwidth]{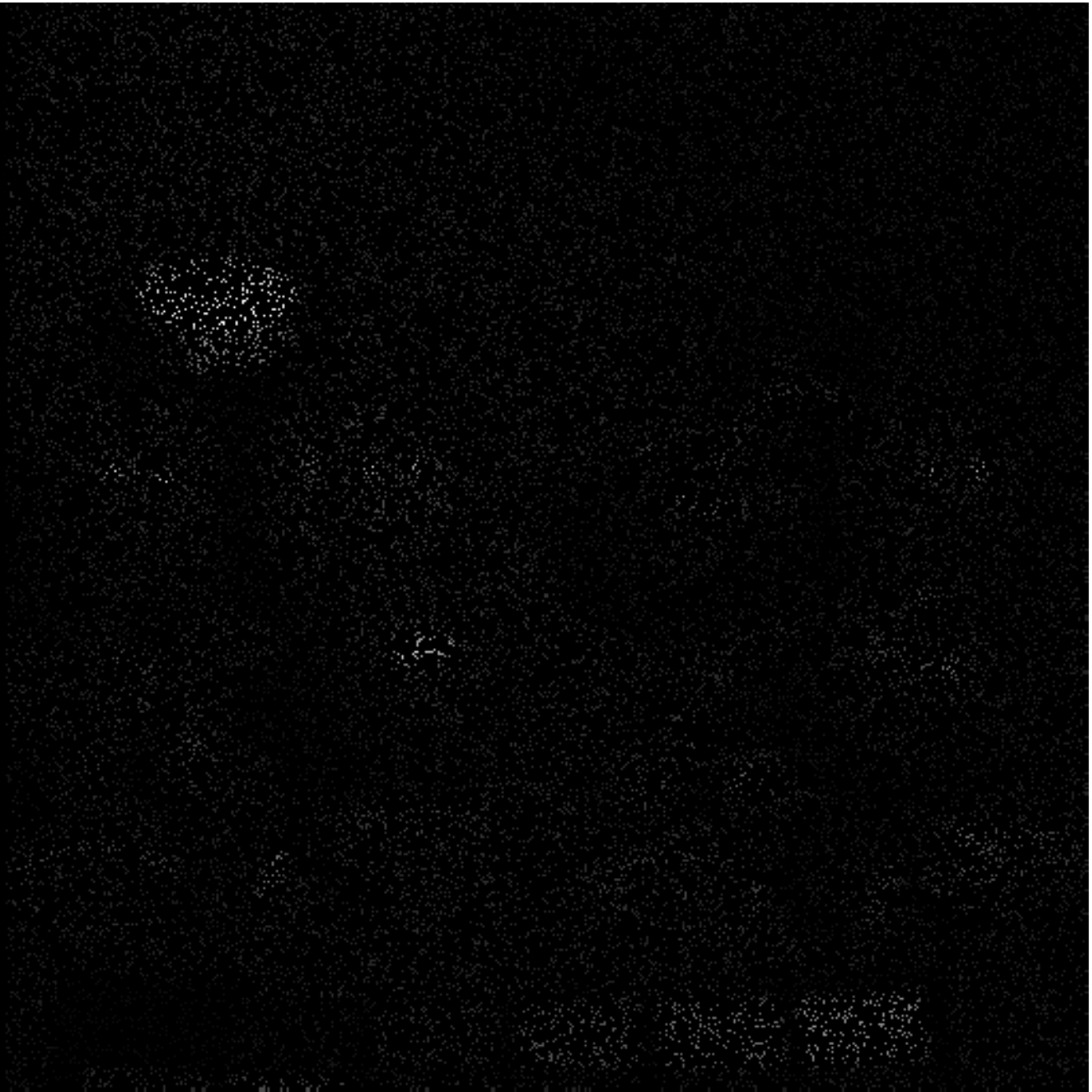}
 \includegraphics[width=0.23\textwidth]{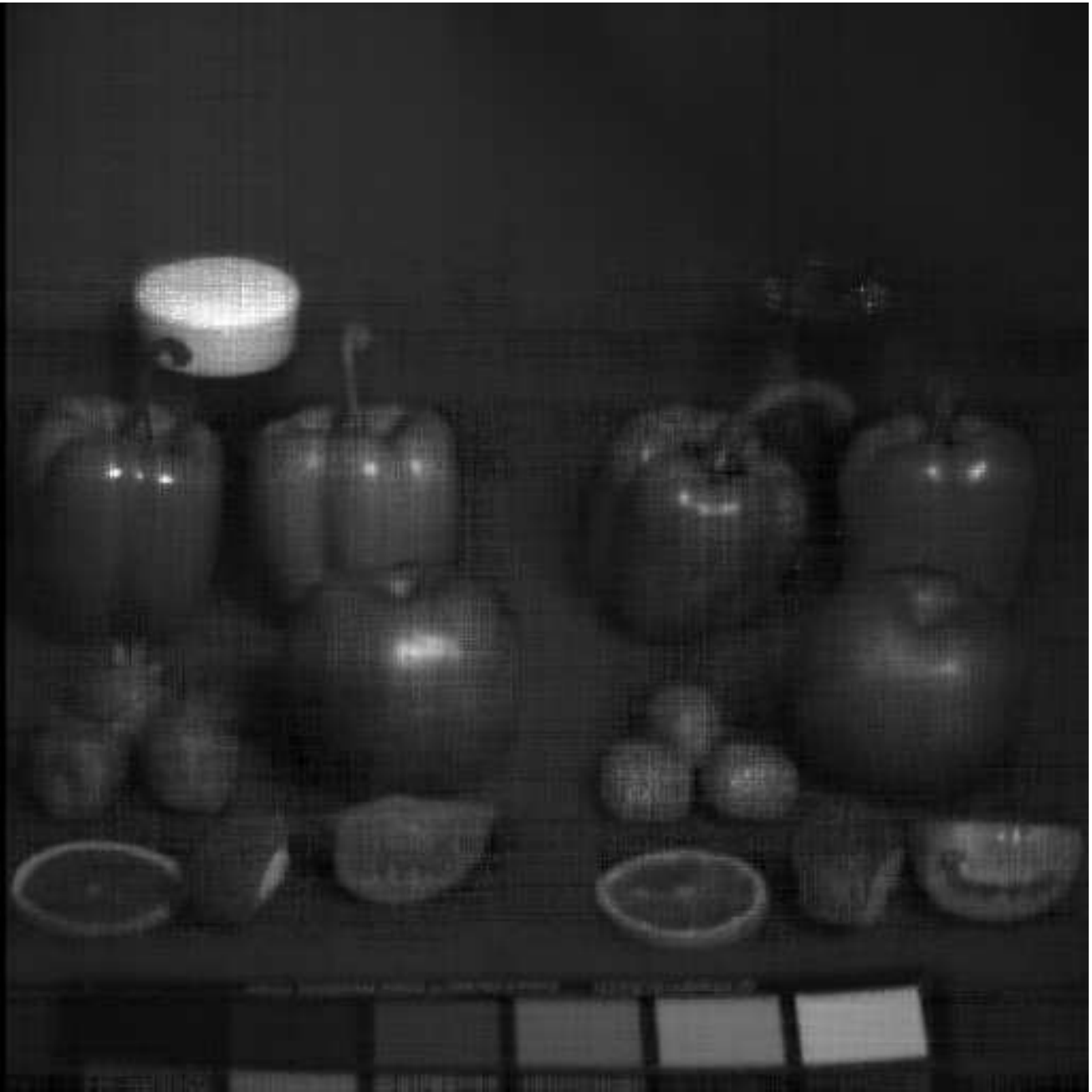}
 \includegraphics[width=0.23\textwidth]{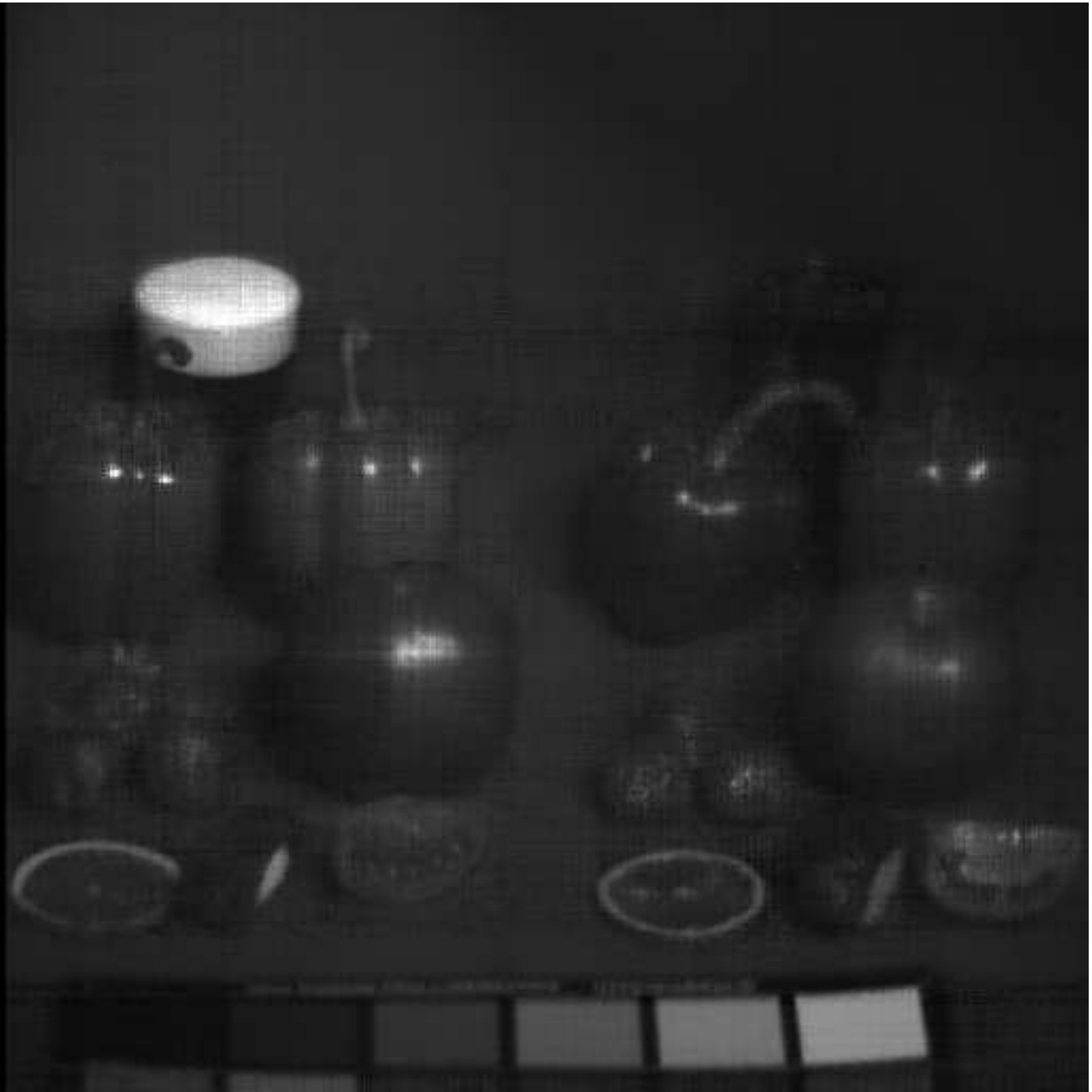}

 \includegraphics[width=0.23\textwidth]{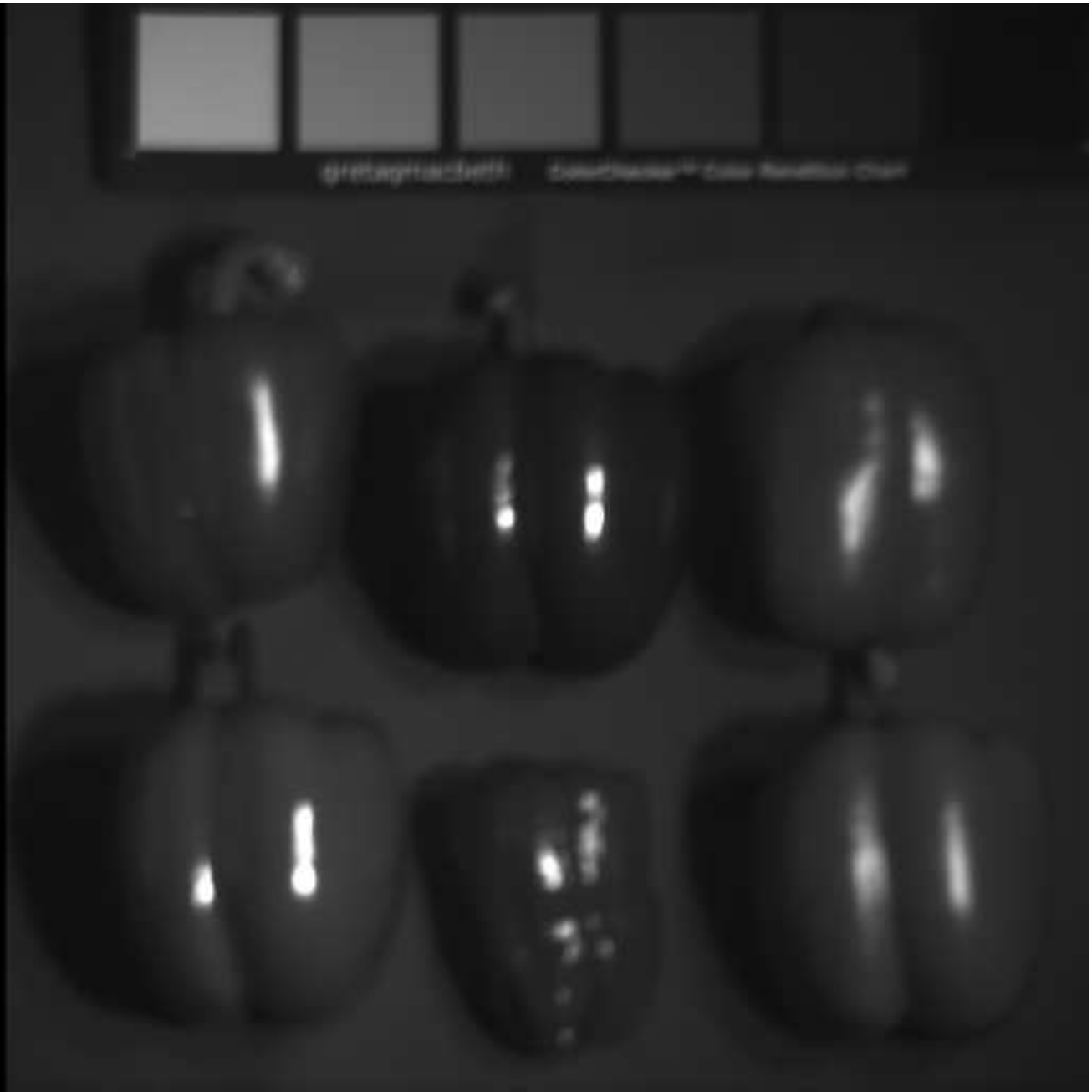}
 \includegraphics[width=0.23\textwidth]{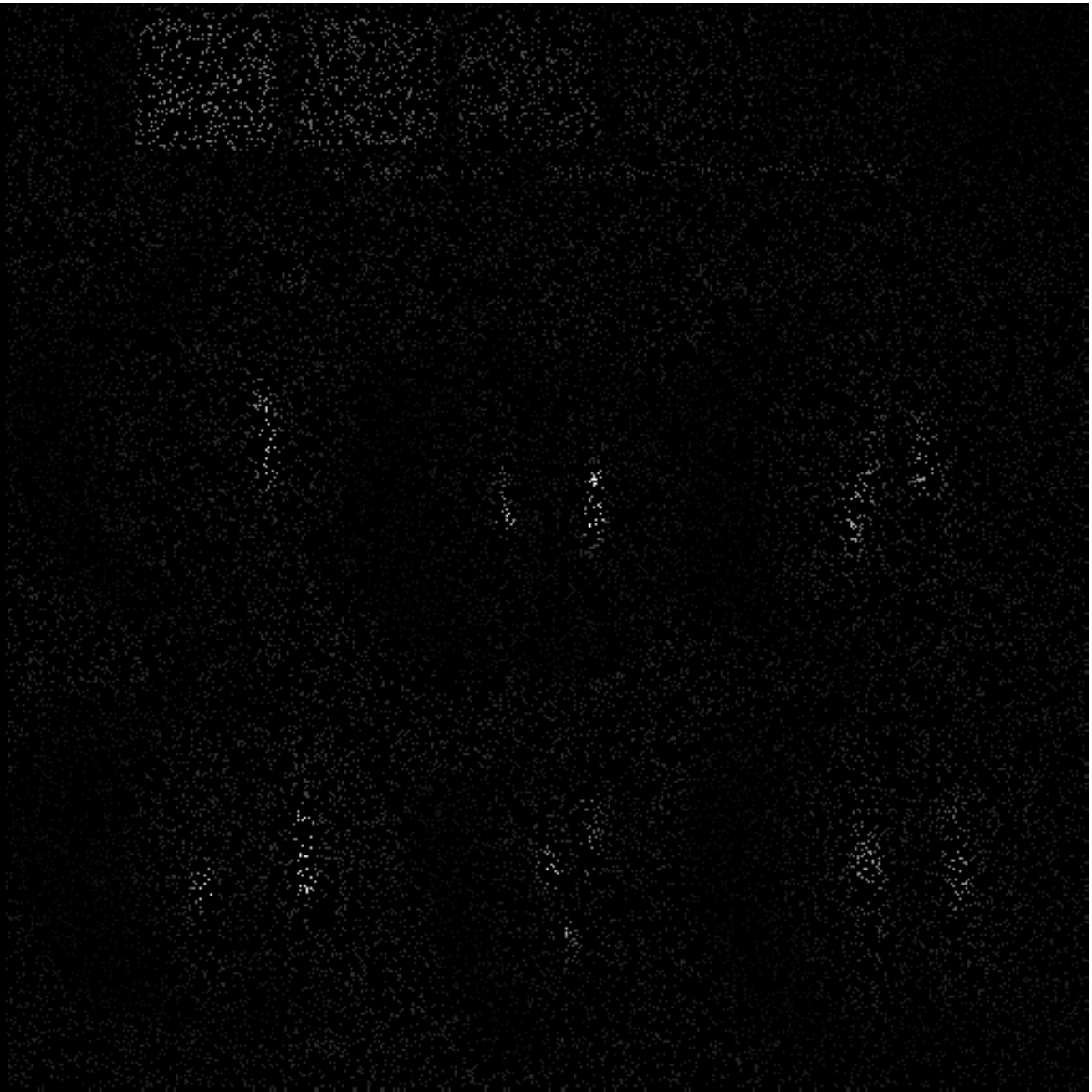}
 \includegraphics[width=0.23\textwidth]{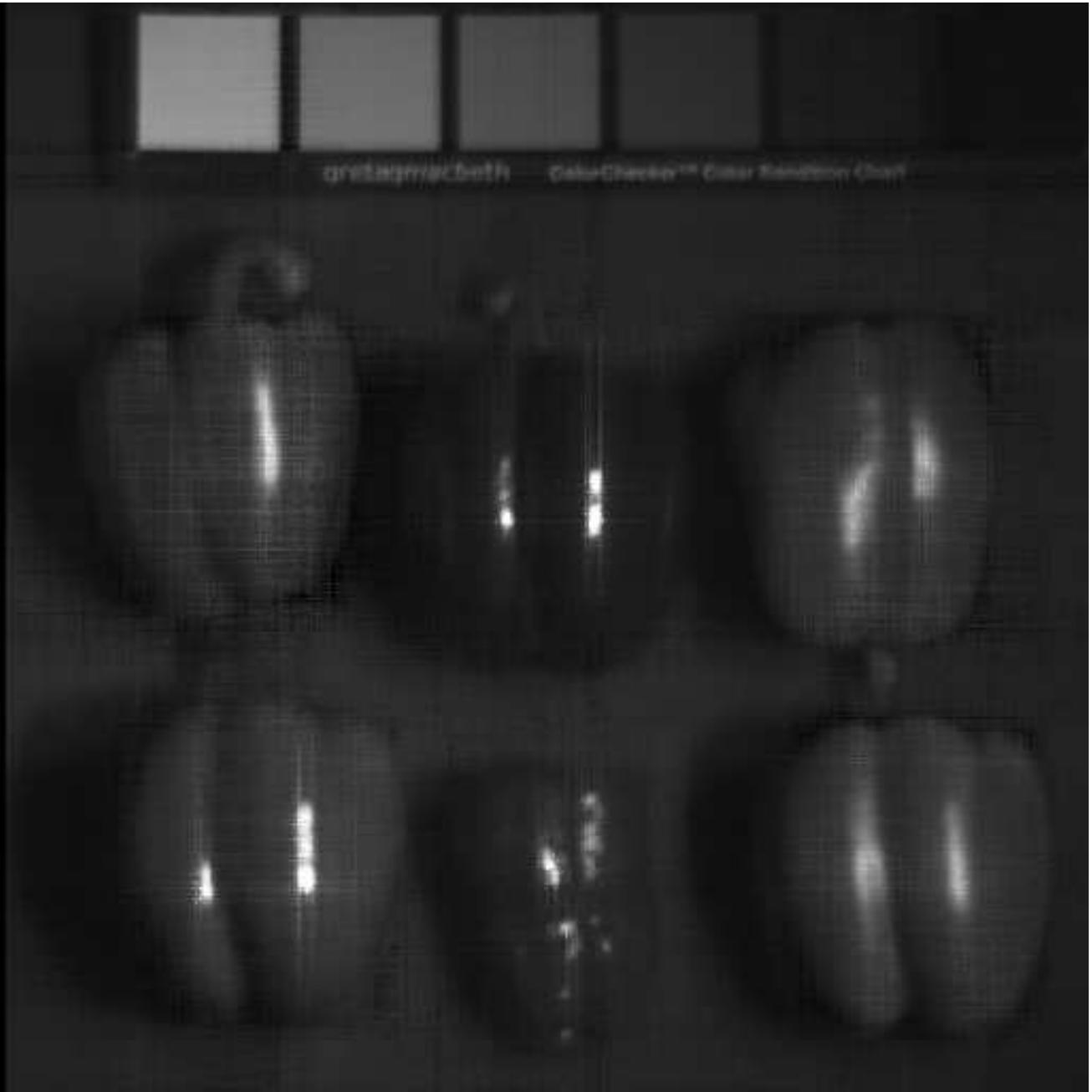}
 \includegraphics[width=0.23\textwidth]{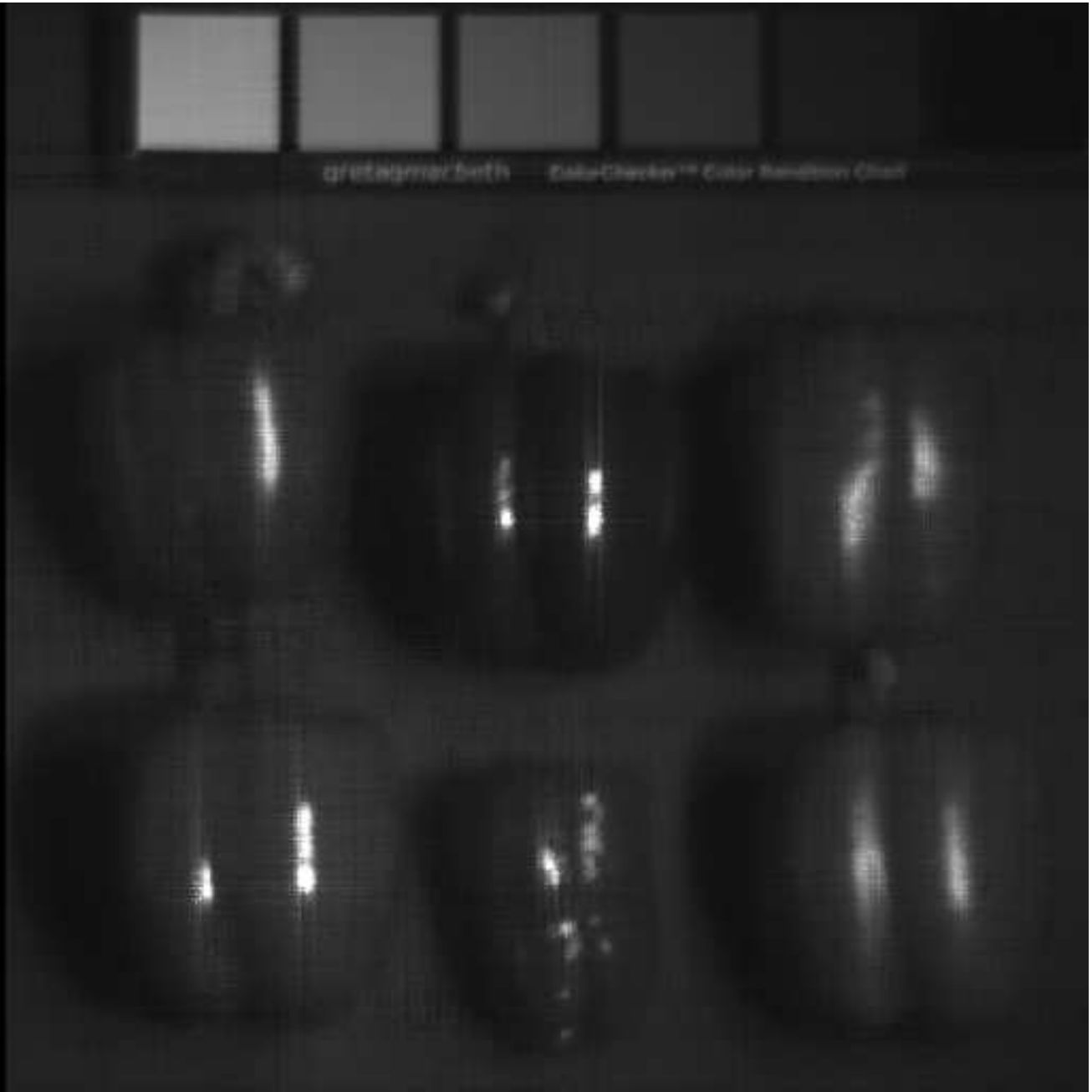}

 \end{center}
 \caption{The first band of recovered MSI images with $\text{SR}=0.1$. From top to bottom: \textit{Pompoms}, \textit{Stuffed toys}, \textit{Foods}, and \textit{Peppers}. From left to right: the original image, the masked image, the results by TNN-F, and TNN-C. }
 \label{figTestMsi}
\end{figure}

\textbf{MSI completion.} For MSI data, we add spectral angle mapper (SAM) and erreur relative globale adimensionnelle de synth$\grave{e}$se (ERGAS) which are common quality metrics for MSI data. SAM calculates the angle in spectral space between pixels and a set of reference tensor on spectral similarity. ERGAS measures fidelity of the recovered tensor based on the weighted sum of mean squared error (MSE) of all bands. The lower the value of these two indicators, the better the results. The size of the MSI data from CAVE database is $512 \times 512 \times 31$ with the wavelengths in the range of $400-700$ nm at an interval of 10nm.
We display one selected tube in Fig.\thinspace\ref{figMsiTube}. We can observe that the tube of recovered tensor by TNN-C is more closely to the true tube than that by TNN-F, especially near the boundary. Moreover, we plot the PSNR values of recovered tensor by TNN-C and TNN-F in Fig.\thinspace\ref{figMsiPsnr}. In general, we can observe that the PSNR values of TNN-C are higher than those of TNN-F, especially for the first and last few bands. Those observations verify TNN-C can produce more natural results as compared to TNN-F when more reasonable BCs is implied in TNN-C.
In Fig.\thinspace\ref{figTestMsi}, we show the first band of testing data recovered by the two methods with $\text{SR} = 0.1$. Obviously, TNN-C achieves better visual results than TNN-F. Tabs.\thinspace\ref{tabTestMsi1}-\ref{tabTestMsi2} give the more detailed data of other testing images. We can see that TNN-C not only has a better performance in PSNR, SSIM, SAM, and ERGAS, but also significantly reduces the time cost compared to TNN-F.

\begin{table}
%[htbp]
    \setlength{\abovecaptionskip}{0pt}%
    \setlength{\belowcaptionskip}{10pt}%
    \caption{PSNR, SSIM, SAM, ERGAS, and time of two methods in MSI completion. In brackets, they are time required for transformation and time required for performing SVD. The best results are highlighted in bold.}
 %  \small
   \centering
%\resizebox{\textwidth}{24mm}{
\begin{tabular}{|cc|cc|cc|}
%cc|cc|cc|}
\hline
\multicolumn{2}{|c|}{MSI} & \multicolumn{2}{c|}{\textit{Pompoms}} & \multicolumn{2}{c|}{\textit{Stuffed toys}}
%& \multicolumn{2}{c}{\textit{Foods}} & \multicolumn{2}{c}{\textit{Peppers}} & \multicolumn{2}{c}{average}
\bigstrut\\
\hline
SR    & metric & TNN-F & TNN-C & TNN-F & TNN-C
%& TNN-F & TNN-C & TNN-F & TNN-C & TNN-F & TNN-C
\bigstrut\\
\hline
\hline
\multirow{6}[2]{*}{0.05} & PSNR  & 26.56  & \textbf{29.00 } & 28.44  & \textbf{31.84 }
%& 31.48  & \textbf{33.33 } & 34.89  & \textbf{36.87 } & 30.34  & \textbf{32.76 }
\bigstrut[t]\\
      & SSIM  & 0.818  & \textbf{0.876 } & 0.892  & \textbf{0.941 } \\
      %& 0.904  & \textbf{0.932 } & 0.946  & \textbf{0.965 } & 0.89  & \textbf{0.93 } \\
      & SAM   & 0.22  & \textbf{0.16 } & 0.30  & \textbf{0.22 } \\
      %& 0.27  & \textbf{0.21 } & 0.21  & \textbf{0.15 } & 0.25  & \textbf{0.19 } \\
      & ERGAS & 10.28  & \textbf{8.00 } & 9.80  & \textbf{6.74 } \\
      %& 9.52  & \textbf{8.01 } & 6.31  & \textbf{5.21 } & 8.98  & \textbf{6.99 } \\
      & \multirow{2}[1]{*}{time} & 309.4  & \textbf{161.0 } & 320.6  & \textbf{183.4 } \\
      %& 281.0  & \textbf{164.8 } & 284.9  & \textbf{155.0 } & 299.0  & \textbf{166.1 } \\
      &       & (11.0+285.7) & \textbf{(8.9+135.3)} & (11.4+296.0) & \textbf{(10.3+153.4)}
      %& \multicolumn{1}{p{5.625em}}{(10.3+258.7)} & \textbf{(9.2+137.9)} & (10.4+255.2) & \textbf{(8.8+128.7)} & (10.8+273.9) & \multicolumn{1}{p{5.625em}}{\textbf{(9.3+138.8)}}
      \bigstrut[b]\\
\hline
\multirow{6}[2]{*}{0.1} & PSNR  & 31.26  & \textbf{33.98 } & 33.37  & \textbf{36.63 }
%& 35.31  & \textbf{37.73 } & 39.25  & \textbf{41.27 } & 34.80  & \textbf{37.40 }
\bigstrut[t]\\
      & SSIM  & 0.922  & \textbf{0.952 } & 0.955  & \textbf{0.978 } \\
      %& 0.957  & \textbf{0.974 } & 0.980  & \textbf{0.989 } & 0.95  & \textbf{0.97 } \\
      & SAM   & 0.13  & \textbf{0.09 } & 0.19  & \textbf{0.14 } \\
      %& 0.18  & \textbf{0.13 } & 0.13  & \textbf{0.09 } & 0.16  & \textbf{0.11 } \\
      & ERGAS & 5.96  & \textbf{4.52 } & 5.53  & \textbf{3.84 } \\
      %& 6.14  & \textbf{4.91 } & 3.86  & \textbf{3.18 } & 5.37  & \textbf{4.11 } \\
      & \multirow{2}[1]{*}{time} & 271.7  & \textbf{171.1 } & 320.2  & \textbf{164.5 } \\
      %& 291.4  & \textbf{167.7 } & 278.3  & \textbf{146.8 } & 290.4  & \textbf{162.5 } \\
      &       & (9.6+251.5) & \textbf{(9.6+143.9)} & (11.2+295.8) & \textbf{(9.2+138.1)}
      %& (10.7+267.9) & \textbf{(9.4+140.2)} & (10.0+256.6) & \textbf{(8.6+124.9)} & (10.4+268.0) & \textbf{(9.2+136.8)}
      \bigstrut[b]\\
\hline
\multirow{6}[2]{*}{0.2} & PSNR  & 37.13  & \textbf{39.55 } & 39.14  & \textbf{41.94 }
 %& 43.13  & \textbf{40.30 } & 44.30  & \textbf{46.22 } & 40.93  & \textbf{42.00 }
 \bigstrut[t]\\
      & SSIM  & 0.976  & \textbf{0.986 } & 0.986  & \textbf{0.994 } \\
      %& 0.993  & \textbf{0.986 } & 0.995  & \textbf{0.997 } & 0.99  & \textbf{0.99 } \\
      & SAM   & 0.07  & \textbf{0.05 } & 0.11  & \textbf{0.09 } \\
      %& 0.11  & \textbf{0.08 } & 0.07  & \textbf{0.05 } & 0.09  & \textbf{0.07 } \\
      & ERGAS & 3.04  & \textbf{2.39 } & 2.82  & \textbf{2.06 } \\
      %& 3.49  & \textbf{2.68 } & 2.19  & \textbf{1.82 } & 2.89  & \textbf{2.24 } \\
      & \multirow{2}[1]{*}{time} & 308.1  & \textbf{184.0 } & 278.9  & \textbf{165.8 } \\
      %& 289.7  & \textbf{164.0 } & 286.2  & \textbf{153.6 } & 290.7  & \textbf{166.8 } \\
      &       & (10.9+284.4) & \textbf{(10.2+154.2)} & (10.2+256.4) & \textbf{(9.2+138.7)}
      %& (10.6+266.7) & \textbf{(9.3+137.4)} & (10.4+264.2) & \textbf{(9.0+138.5)} & (10.5+267.9) & \textbf{(9.4+142.2)}
      \bigstrut[b]\\
\hline
\end{tabular}%
%}
  \label{tabTestMsi1}%
\end{table}%

\begin{table}
%[htbp]
    \setlength{\abovecaptionskip}{0pt}%
    \setlength{\belowcaptionskip}{10pt}%
    \caption{PSNR, SSIM, SAM, ERGAS, and time of two methods in MSI completion. In brackets, they are time required for transformation and time required for performing SVD. The best results are highlighted in bold.}
 %  \small
   \centering
%\resizebox{\textwidth}{24mm}{
\begin{tabular}{|cc|cc|cc|}
%cc|cc|cc|}
\hline
\multicolumn{2}{|c|}{MSI}
%& \multicolumn{2}{c|}{\textit{Pompoms}} & \multicolumn{2}{c|}{\textit{Stuffed toys}}
& \multicolumn{2}{c|}{\textit{Foods}} & \multicolumn{2}{c|}{\textit{Peppers}}
%& \multicolumn{2}{c}{average}
\bigstrut\\
\hline
SR    & metric & TNN-F & TNN-C & TNN-F & TNN-C
%& TNN-F & TNN-C & TNN-F & TNN-C & TNN-F & TNN-C
\bigstrut\\
\hline
\hline
\multirow{6}[2]{*}{0.05} & PSNR
%& 26.56  & \textbf{29.00 } & 28.44  & \textbf{31.84 }
& 31.48  & \textbf{33.33 } & 34.89  & \textbf{36.87 }
%& 30.34  & \textbf{32.76 }
\bigstrut[t]\\
      & SSIM
      %& 0.818  & \textbf{0.876 } & 0.892  & \textbf{0.941 } \\
      & 0.904  & \textbf{0.932 } & 0.946  & \textbf{0.965 } \\
      %& 0.89  & \textbf{0.93 } \\
      & SAM
      %& 0.22  & \textbf{0.16 } & 0.30  & \textbf{0.22 } \\
      & 0.27  & \textbf{0.21 } & 0.21  & \textbf{0.15 } \\
      %& 0.25  & \textbf{0.19 } \\
      & ERGAS
      %& 10.28  & \textbf{8.00 } & 9.80  & \textbf{6.74 } \\
      & 9.52  & \textbf{8.01 } & 6.31  & \textbf{5.21 } \\
      %& 8.98  & \textbf{6.99 } \\
      & \multirow{2}[1]{*}{time}
      %& 309.4  & \textbf{161.0 } & 320.6  & \textbf{183.4 } \\
      & 281.0  & \textbf{164.8 } & 284.9  & \textbf{155.0 } \\
      %& 299.0  & \textbf{166.1 } \\
      &
      %& (11.0+285.7) & \textbf{(8.9+135.3)} & (11.4+296.0) & \textbf{(10.3+153.4)}
      & \multicolumn{1}{p{5.625em}}{(10.3+258.7)} & \textbf{(9.2+137.9)} & (10.4+255.2) & \textbf{(8.8+128.7)}
      %& (10.8+273.9) & \multicolumn{1}{p{5.625em}}{\textbf{(9.3+138.8)}}
      \bigstrut[b]\\
\hline
\multirow{6}[2]{*}{0.1} & PSNR
%& 31.26  & \textbf{33.98 } & 33.37  & \textbf{36.63 }
& 35.31  & \textbf{37.73 } & 39.25  & \textbf{41.27 }
%& 34.80  & \textbf{37.40 }
\bigstrut[t]\\
      & SSIM
      %& 0.922  & \textbf{0.952 } & 0.955  & \textbf{0.978 } \\
      & 0.957  & \textbf{0.974 } & 0.980  & \textbf{0.989 } \\
      %& 0.95  & \textbf{0.97 } \\
      & SAM
      %& 0.13  & \textbf{0.09 } & 0.19  & \textbf{0.14 } \\
      & 0.18  & \textbf{0.13 } & 0.13  & \textbf{0.09 } \\
      %& 0.16  & \textbf{0.11 } \\
      & ERGAS
      %& 5.96  & \textbf{4.52 } & 5.53  & \textbf{3.84 } \\
      & 6.14  & \textbf{4.91 } & 3.86  & \textbf{3.18 } \\
      %& 5.37  & \textbf{4.11 } \\
      & \multirow{2}[1]{*}{time}
      %& 271.7  & \textbf{171.1 } & 320.2  & \textbf{164.5 } \\
      & 291.4  & \textbf{167.7 } & 278.3  & \textbf{146.8 } \\
      %& 290.4  & \textbf{162.5 } \\
      &
      %& (9.6+251.5) & \textbf{(9.6+143.9)} & (11.2+295.8) & \textbf{(9.2+138.1)}
      & (10.7+267.9) & \textbf{(9.4+140.2)} & (10.0+256.6) & \textbf{(8.6+124.9)}
      %& (10.4+268.0) & \textbf{(9.2+136.8)}
      \bigstrut[b]\\
\hline
\multirow{6}[2]{*}{0.2} & PSNR
%& 37.13  & \textbf{39.55 } & 39.14  & \textbf{41.94 }
 & 43.13  & \textbf{40.30 } & 44.30  & \textbf{46.22 }
 %& 40.93  & \textbf{42.00 }
 \bigstrut[t]\\
      & SSIM
      %& 0.976  & \textbf{0.986 } & 0.986  & \textbf{0.994 } \\
      & 0.993  & \textbf{0.986 } & 0.995  & \textbf{0.997 } \\
      %& 0.99  & \textbf{0.99 } \\
      & SAM
      %& 0.07  & \textbf{0.05 } & 0.11  & \textbf{0.09 } \\
      & 0.11  & \textbf{0.08 } & 0.07  & \textbf{0.05 } \\
      %& 0.09  & \textbf{0.07 } \\
      & ERGAS
      %& 3.04  & \textbf{2.39 } & 2.82  & \textbf{2.06 } \\
      & 3.49  & \textbf{2.68 } & 2.19  & \textbf{1.82 } \\
      %& 2.89  & \textbf{2.24 } \\
      & \multirow{2}[1]{*}{time}
      %& 308.1  & \textbf{184.0 } & 278.9  & \textbf{165.8 } \\
      & 289.7  & \textbf{164.0 } & 286.2  & \textbf{153.6 } \\
      %& 290.7  & \textbf{166.8 } \\
      &
      %& (10.9+284.4) & \textbf{(10.2+154.2)} & (10.2+256.4) & \textbf{(9.2+138.7)}
      & (10.6+266.7) & \textbf{(9.3+137.4)} & (10.4+264.2) & \textbf{(9.0+138.5)}
      %& (10.5+267.9) & \textbf{(9.4+142.2)}
      \bigstrut[b]\\
\hline
\end{tabular}%
%}
  \label{tabTestMsi2}%
\end{table}%

\begin{figure}[!htp]
 \begin{center}
 \includegraphics[width=0.4\textwidth]{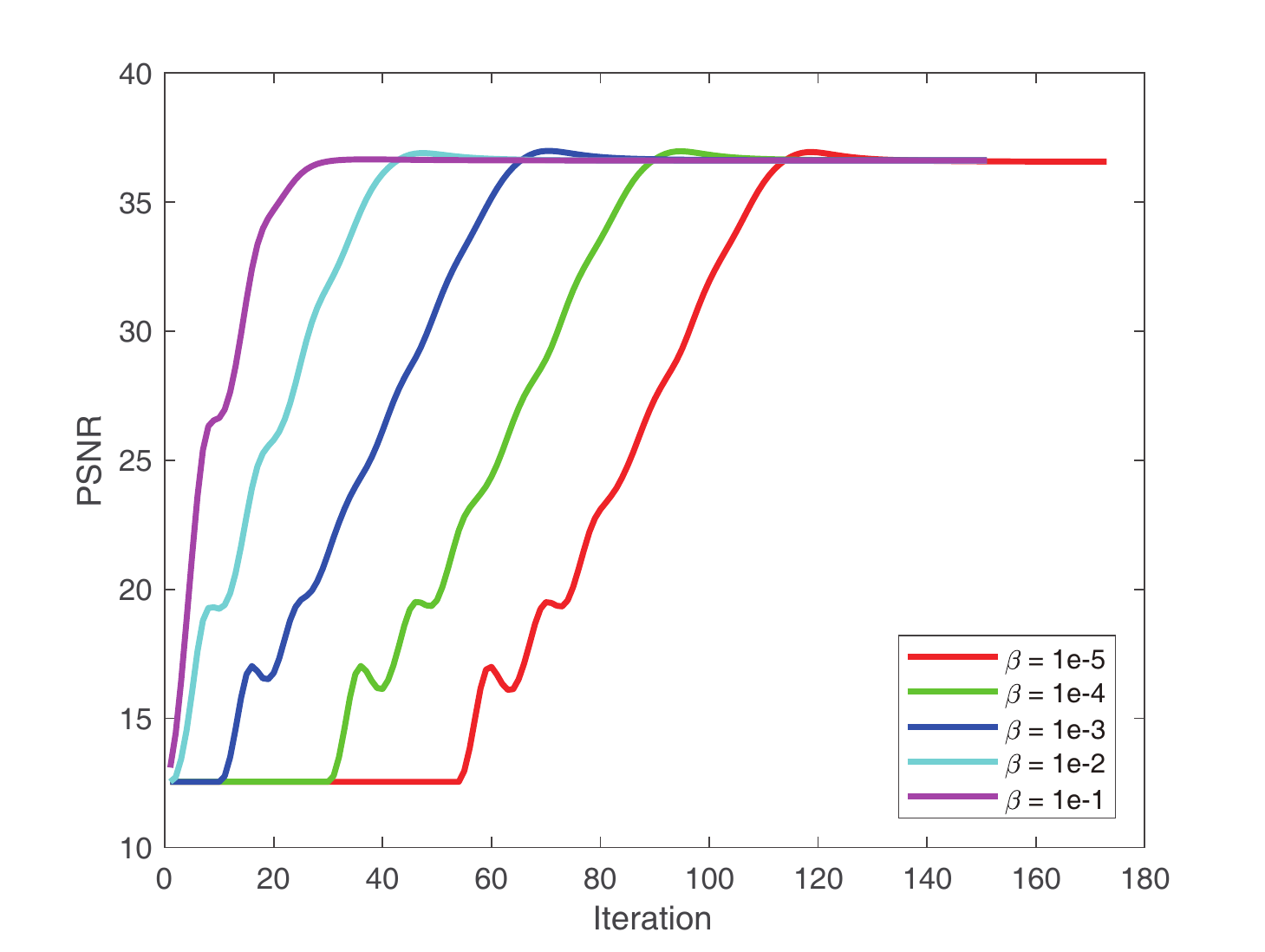}

 \end{center}
 \caption{The PSNR values with respect to the iteration for different values of parameter $\beta$.}
 \label{figParaAna}
\end{figure}

\textbf{Parameter analysis.} We analyze the robustness of TNN-C for different parameters using MSI data \textit{Stuffed toys} with $SR = 0.1$. TNN-C only requires one parameter $\beta$. As shown in Fig.\thinspace(\ref{figParaAna}), different $\beta$ lead to nearly the same PSNR value, but it affects the convergence speed. After testing, we choose $\beta = 1 \times 10^{-2}$ for all experiments.

\section{Concluding Remarks}

We have introduced the DCT as an alternative of DFT into the framework of t-SVD. Based on the resulting t-SVD, the DCT based tensor nuclear norm (TNN-C) is suggested for low-rank tensor completion problem. We have developed an efficient alternating direction method of multipliers (ADMM) to tackle the corresponding model. Numerical experiments are reported to demonstrate the superiority of the DCT-based t-SVD. In the future
research work, other transforms based tensor singular value decomposition can be considered and studied. We expect other transforms based
tensor singular value decomposition can deal with data tensors from specific applications.

\section*{Acknowledgment}
The research is supported by NSFC (61772003) and the Fundamental Research Funds for the Central Universities (ZYGX2016J132),
the HKRGC GRF 1202715, 12306616,
12200317 and HKBU RC-ICRS/16-17/03.

\section*{References}
\bibliographystyle{elsarticle-num}
\bibliography{referrence}

\begin{thebibliography}{10}
\expandafter\ifx\csname url\endcsname\relax
  \def\url#1{\texttt{#1}}\fi
\expandafter\ifx\csname urlprefix\endcsname\relax\def\urlprefix{URL }\fi
\expandafter\ifx\csname href\endcsname\relax
  \def\href#1#2{#2} \def\path#1{#1}\fi

\bibitem{bertalmio2000image}
M.~Bertalmio, G.~Sapiro, V.~Caselles, C.~Ballester, Image inpainting,
  Proceedings of International Conference on Computer Graphics and Interactive
  Techniques (2000) 417--424 (2000).

\bibitem{komodakis2006image}
N.~Komodakis, Image completion using global optimization, Proceedings of
  Computer Vision and Pattern Recognition (2006) 442--452 (2006).

\bibitem{liu2013tensor}
J.~Liu, P.~Musialski, P.~Wonka, J.-P. Ye, Tensor completion for estimating
  missing values in visual data, IEEE Transactions on Pattern Analysis and
  Machine Intelligence 35~(1) (2013) 208--220 (2013).

\bibitem{korah2007spatiotemporal}
T.~Korah, C.~Rasmussen, Spatiotemporal inpainting for recovering texture maps
  of occluded building facades, IEEE Transactions on Image Processing 16~(9)
  (2007) 2262--2271 (2007).

\bibitem{chan2011an}
S.~H. Chan, R.~Khoshabeh, K.~B. Gibson, P.~E. Gill, T.~Q. Nguyen, An augmented
  lagrangian method for total variation video restoration, IEEE Transactions on
  Image Processing 20~(11) (2011) 3097--3111 (2011).

\bibitem{jiang2017a}
T.-X. Jiang, T.-Z. Huang, X.-L. Zhao, L.-J. Deng, Y.~Wang, A novel tensor-based
  video rain streaks removal approach via utilizing discriminatively intrinsic
  priors, Proceedings of Computer Vision and Pattern Recognition (2017)
  2818--2827 (07 2017).

\bibitem{li2012coupled}
F.~Li, M.~K. Ng, R.~J. Plemmons, Coupled segmentation and denoising/deblurring
  models for hyperspectral material identification, Numerical Linear Algebra
  With Applications 19~(1) (2012) 153--173 (2012).

\bibitem{zhao2013deblurring}
X.-L. Zhao, F.~Wang, T.-Z. Huang, M.~K. Ng, R.~J. Plemmons, Deblurring and
  sparse unmixing for hyperspectral images, IEEE Transactions on Geoscience and
  Remote Sensing 51~(7) (2013) 4045--4058 (2013).

\bibitem{li2010tensor}
N.~Li, B.-X. Li, Tensor completion for on-board compression of hyperspectral
  images, Proceedings of IEEE International Conference on Image Processing
  (2010) 517--520 (2010).

\bibitem{xing2012dictionary}
Z.-M. Xing, M.-Y. Zhou, A.~Castrodad, G.~Sapiro, L.~Carin, Dictionary learning
  for noisy and incomplete hyperspectral images, SIAM Journal on Imaging
  Sciences 5~(1) (2012) 33--56 (2012).

\bibitem{sun2005cubesvd:}
J.-T. Sun, H.-J. Zeng, H.~Liu, Y.-C. Lu, Z.~Chen, Cubesvd: a novel approach to
  personalized web search, Proceedings of International World Wide Web
  Conferences (2005) 382--390 (2005).

\bibitem{kolda2005higher-order}
T.~G. Kolda, B.~W. Bader, J.~P. Kenny, Higher-order web link analysis using
  multilinear algebra, Proceedings of IEEE International Conference on Data
  Mining (2005) 242--249 (2005).

\bibitem{varghees2012adaptive}
N.~Varghees, M.~Manikandan, R.~G. John, Adaptive mri image denoising using
  total-variation and local noise estimation, Proceedings of IEEE International
  Conference on Advances in Engineering, Science and Management (2012) 506--511
  (01 2012).

\bibitem{kreimer2012a}
N.~Kreimer, M.~D. Sacchi, A tensor higher-order singular value decomposition
  for prestack seismic data noise reduction and interpolation, Geophysics
  77~(3) (2012) 113--122 (2012).

\bibitem{CPdecomposition}
R.~A. Harshman, Foundations of the parafac procedure: Models and conditions for
  an ``explanatory'' multi-modal factor analysis, UCLA Working Papers in
  Phonetics (1970).

\bibitem{tucker1966some}
L.~R. Tucker, Some mathematical notes on three-mode factor analysis,
  Psychometrika 31~(3) (1966) 279--311 (1966).

\bibitem{kilmer2011factorization}
M.~E. Kilmer, C.~D.~M. Martin, Factorization strategies for third-order
  tensors, Linear Algebra and its Applications 435~(3) (2011) 641--658 (2011).

\bibitem{martin2013an}
C.~D. Martin, R.~Shafer, B.~Larue, An order-$p$ tensor factorization with
  applications in imaging, SIAM Journal on Scientific Computing 35 (2013)
  474--490 (2013).

\bibitem{kilmer2013third-order}
M.~E. Kilmer, K.~S. Braman, N.~Hao, R.~C. Hoover, Third-order tensors as
  operators on matrices: A theoretical and computational framework with
  applications in imaging, SIAM Journal on Matrix Analysis and Applications
  34~(1) (2013) 148--172 (2013).

\bibitem{ng1999a}
M.~K. Ng, R.~H. Chan, W.~Tang, A fast algorithm for deblurring models with
  neumann boundary conditions, SIAM Journal on Scientific Computing 21~(3)
  (1999) 851--866 (1999).

\bibitem{zhang2014novel}
Z.-M. Zhang, G.~Ely, S.~Aeron, H.~Ning, M.~E. Kilmer, Novel methods for
  multilinear data completion and de-noising based on tensor-svd, Proceedings
  of Computer Vision and Pattern Recognition (2014) 3842--3849 (2014).

\bibitem{lu2016tensor}
C.-Y. Lu, J.-S. Feng, Y.-D. Chen, W.~Liu, Z.-C. Lin, S.-C. Yan, Tensor robust
  principal component analysis: Exact recovery of corrupted low-rank tensors
  via convex optimization, Proceedings of Computer Vision and Pattern
  Recognition (2016) 5249--5257 (2016).

\bibitem{semerci2014tensor-based}
O.~Semerci, H.~Ning, M.~E. Kilmer, E.~L. Miller, Tensor-based formulation and
  nuclear norm regularization for multienergy computed tomography, IEEE
  Transactions on Image Processing 23~(4) (2014) 1678--1693 (2014).

\bibitem{boyd2011distributed}
S.~Boyd, N.~Parikh, E.~Chu, B.~Peleato, J.~Eckstein, Distributed optimization
  and statistical learning via the alternating direction method of multipliers,
  Found. Trends Mach. Learn. 3~(1) (2011) 1--122 (Jan. 2011).

\bibitem{lin2010augmented}
Z.-C. Lin, M.-M. Chen, Y.~Ma, L.-Q. Wu, {The Augmented Lagrange Multiplier
  Method for Exact Recovery of Corrupted Low-Rank Matrices}, ArXiv e-prints
  (Sep. 2010).
\newblock \href {http://arxiv.org/abs/1009.5055} {\path{arXiv:1009.5055}}.

\bibitem{he2012alternating}
B.-S. He, M.~Tao, X.-M. Yuan, Alternating direction method with gaussian back
  substitution for separable convex programming, SIAM Journal on Optimization
  22~(2) (2012) 313--340 (2012).

\bibitem{afonso2011an}
M.~V. Afonso, J.~M. Bioucasdias, M.~A.~T. Figueiredo, An augmented lagrangian
  approach to the constrained optimization formulation of imaging inverse
  problems, IEEE Transactions on Image Processing 20~(3) (2011) 681--695
  (2011).

\bibitem{han2012a}
D.-R. Han, X.-M. Yuan, A note on the alternating direction method of
  multipliers, Journal of Optimization Theory and Applications 155~(1) (2012)
  227--238 (2012).

\end{thebibliography}

\end{document}